\documentclass[final]{clv2025}
\usepackage{siunitx}

\jvol{vv}
\jnum{nn}
\jyear{2025}
\usepackage{amsmath}
\usepackage{booktabs}

\runningtitle{Tokens, the oft-overlooked appetizer}
\runningauthor{Zimmerman, et al.}

\begin{document}

\title{Tokens, the oft-overlooked appetizer: 
Large language models, 
the distributional hypothesis, 
and meaning}

\author{Julia Witte Zimmerman\thanks{Corresponding author}$^{,1}$,
    Denis Hudon$^{2}$,
    Kathryn Cramer$^{1}$,
    Alejandro Javier Ruiz Iglesias$^{1}$,
    Calla Beauregard$^{1}$,
    Ashley Fehr$^{1}$,
    Mikaela Irene Fudolig$^{1}$,
    Bradford Demarest$^{3}$,
    Yoshi Meke Bird$^{1}$,
    Milo Z. Trujillo$^{4}$,
    Christopher M. Danforth$^{1,5}$,
    Peter Sheridan Dodds$^{1,6}$
    }

\affilblock{
    \affil{
        Computational Story Lab,\\
        Vermont Complex Systems Institute,\\
        University of Vermont,\\
        Burlington, VT 05405, USA\\\quad \email{julia.zimmerman@uvm.edu}
    }
    \affil{
        Vermont Complex Systems Institute,\\
        University of Vermont,\\
        Burlington, VT 05405, USA\\\quad
    }
    \affil{
        Department of Computer Science,\\
        St. Michael's College,\\
        Colchester, VT 05439, USA\\\quad
    }
    \affil{
        Communication Media and Marginalization Lab,\\
        Northeastern University, Boston, MA 02215, USA\\\quad
    }
    \affil{
        Department of Mathematics \& Statistics,\\
        University of Vermont,\\
        Burlington, VT 05405, USA\\\quad
    }
    \affil{
        Department of Computer Science,\\
        University of Vermont,\\
        Burlington, VT 05405, USA\\\quad
    }
}

\maketitle

\begin{abstract}
Tokenization is a necessary component within the current architecture of many language mod-els, including the transformer-based large language models (LLMs) of Generative AI, yet its impact on the model's cognition is often overlooked. We argue that LLMs demonstrate that the Distributional Hypothesis (DH) is sufficient for reasonably human-like language performance (particularly with respect to inferential lexical competence), and that the emergence of human-meaningful linguistic units among tokens and current structural constraints motivate changes to existing, linguistically-agnostic tokenization techniques, particularly with respect to their roles as (1) vehicles for conveying salient distributional patterns from human language to the model and as (2) semantic primitives. We explore tokenizations from a BPE tokenizer; extant model vocabularies obtained from Hugging Face and tiktoken; and the information in exemplar token vectors as they move through the layers of a RoBERTa (large) model. Besides creating suboptimal semantic building blocks and obscuring the model's access to the necessary distributional patterns, we describe how tokens and pretraining can act as a backdoor for bias and other unwanted content, which current alignment practices may not remediate. Additionally, we relay evidence that the tokenization algorithm's objective function impacts the LLM's cognition, despite being arguably meaningfully insulated from the main system intelligence. Finally, we discuss implications for architectural choices, meaning construction, the primacy of language for thought, and LLM cognition.
\end{abstract}

\raggedbottom

\section{Introduction}
\label{sec:introduction}

Large language models (LLMs) are artificial intelligence (AI) or machine learning (ML) models that take in vast amounts of text as training data and create new utterances by choosing the most plausible small pieces of text to output next. Such models are quickly becoming ubiquitous~\cite{nunes2024moral}.\footnote{For a brief description of the emerging Generative AI landscape, see Lovato \textit{et al.}, 2024~\cite{lovato2024foregrounding}.}
In the context of LLMs, although other aspects of their architecture may vary, \textit{tokens} are the substrings\footnote{Technically, more commonly, arrays of bytes} that the model breaks all input text into, and the units it concatenates to generate output text.\footnote{LLMs are often transformer-based models~\cite{vaswani2023attention}. Masked language models (MLMs), such as BERT-based models~\cite{devlin2019bertpretrainingdeepbidirectional}, see text both before and after masked tokens, the token they are supposed to provide, while causal language models (CLMs) or auto-regressive models, such as GPT-based models~\cite{yenduri2023generativepretrainedtransformercomprehensive}, only see text before the masked tokens, predicting the next token following a sequence of prior tokens.}
As tokens are the bridge from raw text to numbers that LLMs can work with, from our world (\textit{koinos kosmos}) to the LLM's internal world (\textit{idios kosmos}), they play a crucial role in the architecture of LLMs, yet their impact on the model's cognition is often overlooked.

We argue that tokens deserve greater attention (across all levels of abstraction) than they typically receive: 
From the straightforward, to direct implications for alignment\footnote{Broadly, this term refers to the attempt to align AI models with human values and judgments. In practice it often means strategies for trying to constrain the sort of output the model will generate to non-harmful, inoffensive content.} and performance, to theoretical consequences for the model's cognition.\footnote{More broadly, now that LLMs have clearly demonstrated the potential of their architecture, we should re-examine many of the underconstrained aspects of their components---such as the objective functions, structures, operations, relationships, connections-- to fully harness that potential. Tokenization is one of the most fundamental components, and determinative of the input signal interface, making it an obvious place to start.}\\

Please note that there are offensive words in this paper, including in figures and tables, and especially in sections discussing extant tokens, bias, and alignment.\\

We complement works like Chiang and Yogatama, 2023~\cite{chiang-yogatama-2023-distributional}, Huang \textit{et al.}, 2023 ~\cite{huang2023lexinvariantlanguagemodels}, and Wen-Yi and Mimno, 2023 ~\cite{wen-yi-mimno-2023-hyperpolyglot}\footnote{Wen-Yi and Mimno, 2023, consider tokenization from a different but overlapping perspective within the LLM architecture, with similar motivations: ``Every language model has an input layer that maps tokens to vectors. This ubiquitous layer of language models is often overlooked. We find that similarities between these input embeddings are highly interpretable and that the geometry of these embeddings differs between model families''~\cite{wen-yi-mimno-2023-hyperpolyglot}. We think this line of inquiry, especially paired with strategies like the one conducted in Huang \textit{et al.}~\cite{huang2023lexinvariantlanguagemodels}, could provide insight into linguistic relativism and what kind of information is stored at what level (of architecture and of abstraction) within the LLM, especially when paired with linguistic explanations as to what strategies can be used to encode different aspects of meaning (e.g., theories about the syntax-semantics interface and from typology)} and extend Zhemchuzhina \textit{et al.}, 2022~\cite{zhemchuzhina2022pragmatic} with an exploration of tokenization and its consequences. 
Through the lens of the distributional hypothesis (DH)---the theory from linguistics that proposes that words used in similar contexts are also similar in meaning---we step through the tokenization process to contextualize how it works and what the tokens it yields look like.
We provide information on extant tokens broken down by categories. Finally, we try a new methodology to look at the information in the trajectories through the model (RoBERTa MLM~\cite{liu2019robertarobustlyoptimizedbert}) of a few exemplar tokens, giving us insight into the information space within vectors corresponding to a single token (more usually, embedding spaces are examined across many different tokens).

\subsection{Background of this work}

This paper is an abstracted excerpt from the first author's doctoral thesis~\cite{zimmerman2025locality}, and in particular is a continuation of work started in Zimmerman \textit{et al.}, first made publicly available in 2023~\cite{zimmerman2024blind}. More in-depth discussion, 
background, and results are available in the thesis.

We recommend reading our previous paper to understand our underlying claims about what LLMs are capable of perceiving.
Briefly: 
For people, language is an embodied task and our senses ground our cognition, but LLMs do not have a human-like body, so although we say LLMs are trained on ``language'' or ``text'', they can in fact only access a small portion of that text. Specifically, the 1D sequences of symbols and the relations of context (proximity, adjacency, co-occurrence) between them, and distributions.\footnote{This leads us to ask questions like what can an LLM come to know, from the distributional properties of language as reflected in what it can perceive of text? How can LLMs, with such curtailed and comparatively indirect access to the world, come to know about it (the symbol grounding problem)?} As a consequence, any information that can be perceived by the LLM must seemingly take a descriptive, propositional shape. For example, ChatGPT cannot perceive what the letter \textit{c} looks like, but we can provide it a diegetic approximation in text, such as, ``the letter c is round like an oval or a circle, but missing a segment on the right side''. Although it is immediately apparent that perceiving the appearance of \textit{c} and our diegetic approximation of it are not identical, that representation may be sufficient for some tasks. Importantly, there is no obvious reason to think the approximation cannot be arbitrarily good.\footnote{Learning what words mean through the DH (regime 1) eventually allows learning abstract concepts as encoded in words, opening up a tall ceiling for language-mediated knowledge~\cite{zimmerman2025locality}.} As a corollary: We should be wary of benchmarks! Benchmarks are often an approximation of a task (e.g., for LLMs, a complex task may be benchmarked with a game of generating plausible text, at which LLMs excel). Inherently, they do not perfectly capture the task (only the thing is the thing). With respect to LLMs in particular, benchmarking can lead to a doubly-terminated fallacy of interpretation, where both the construct validity of the task-as-game-of-text in the first place and the subsequent interpretation of the solicited output text are suspect, with people quick to attribute human Theory of Mind to a model with very different architecture~\cite{zimmerman2024blind,zimmerman2025locality}.

\begin{figure*}
\centering
\includegraphics[width=0.9\textwidth]{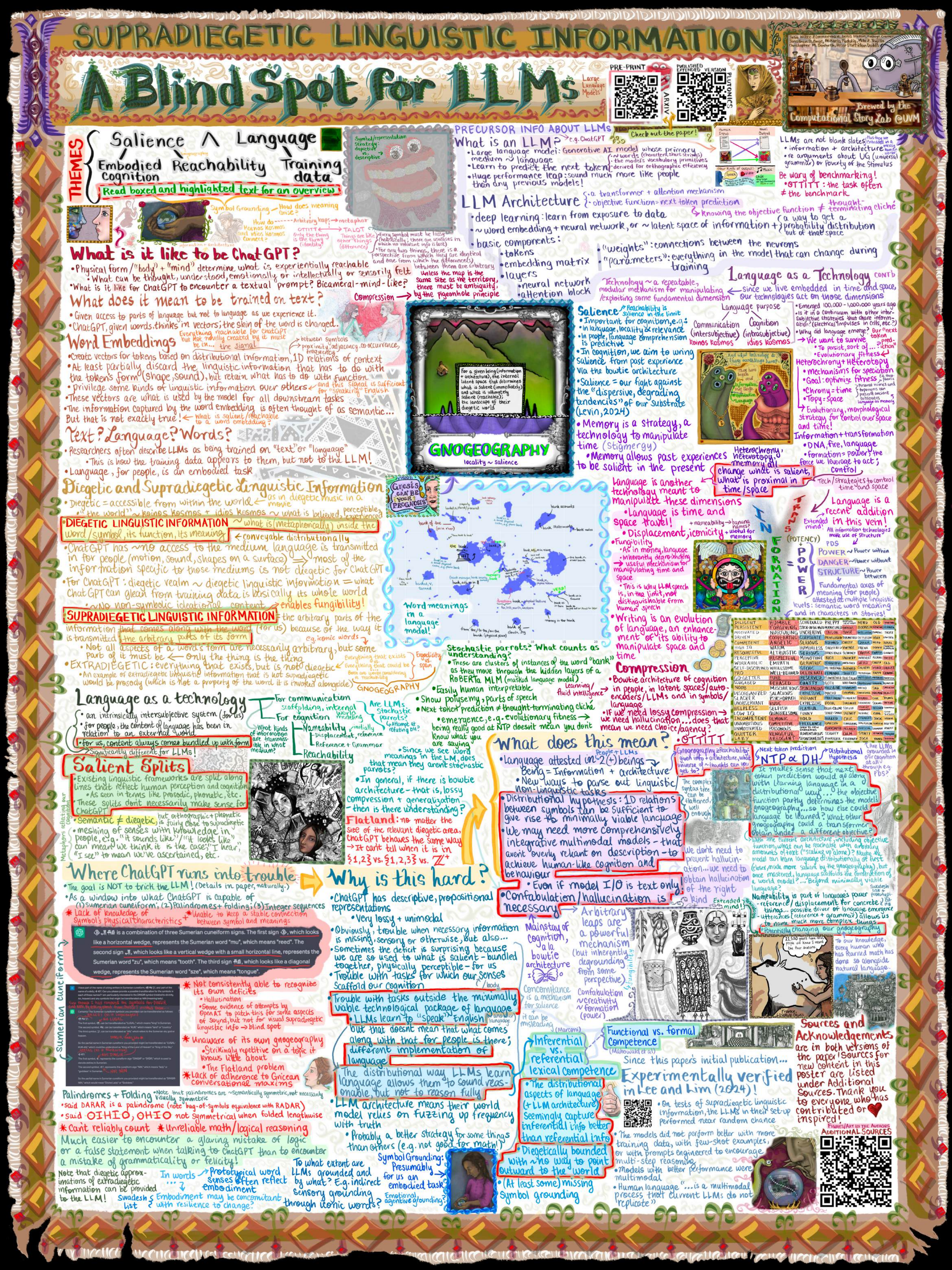}
\caption{This poster, which was displayed at IC2S2 2024 (10th International Conference on Computational Social Science)~\cite{ic2s22024event}, summarizes our previous paper~\cite{zimmerman2024blind} and includes discussion on grounding, salience, and language as a technology. We list the sources in the QR code here:~\cite{theluddite,craiyon,zimmerman2024blind,mywebsite,chameleonteam2024chameleonmixedmodalearlyfusionfoundation,
heikkila2023making,kapoor2024largelanguagemodelstaught,ptosis,pet,liesenfeld2024rethinking,sutskever2011generating,vaswani2023attention,BolukbasiCZSK16a,3blue1brown,radhakrishnan2023mechanismfeaturelearningdeep,harris1954structure,lovato2024foregrounding,doddstarot,larousse1970ancient,zimmer2024need,understandingllmunderstanding,fedorenko2024primarily,
templeton2024scaling,levintedtalk,levin2024stigmergy,lee2024tasks,dodds2023ousiometricstelegnomicsessencemeaning,doddscharacterspace,kallini2024mission,
mahowald2024dissociatinglanguagethoughtlarge,maudslay-etal-2024-chainnet,futrell2024linguistic,hahn2020universals,OpenAI_ChatGPT,sep-speech-acts,HerbertDuneSeries,10.1145/3442188.3445922,marconi2003lexical,borges1999collected}.}
\label{fig:supradiegetic_poster}
\end{figure*}

In the previous paper, we focused on what aspects of the linguistic signal are perceptible to the LLM~\cite{zimmerman2024blind}; in this paper, we consider the impact of the symbols upon which the distributional aspects operate, moving one level deeper into the model. For a summary of our prior work, see Figure~\ref{fig:supradiegetic_poster}. Since publication, Lee and Lim, 2024, experimentally tested our theory, showing ``that scaling model size and data, while beneficial for some aspects of language understanding, does not address the deficiency in orthographic abilities''~\cite{lee2024tasks}. This conclusion refers to the supradiegetic linguistic information that we argued is imperceptible to LLMs based on their architecture. The point being that salience, a defining feature of cognition~\cite{levin2024stigmergy},\footnote{Maybe even \textit{the} defining feature of cognition~\cite{levin2024stigmergy}.} is a valuable lens through which to understand the capabilities of LLMs: information must be accessible, or local, as a precondition of salience. We provide background on tokenization and the distributional hypothesis in the remainder of the introductory section.

\subsection{Intent and scope of this work}

This paper contains elements of both philosophical, exploratory work and quantitative analysis. Our aim is for the empirical observations presented here to provide fertile ground for thought, allowing us to build a productive mental model of LLMs with explanatory power that can be further refined and experimentally tested in future work. The empirical sections produce observations; the Discussion, Conclusion, and Post-Script offer a theoretical synthesis intended to explain those observations, organize related findings, and generate future predictions and hypotheses.

Accordingly, the quantitative evidence presented here is not intended to independently establish every broader claim advanced later in the paper. Rather, we view the empirical results as constraints on an emerging framework and as points of contact between theory and observation. Many of the broader claims should therefore be understood as interpretations, conjectures, and directions for future investigation.

Several related ideas emerge throughout this paper as a partial mental model for how LLMs learn from the world. First, tokens are the ``eye'' of contemporary LLMs. Because tokenization determines the distributions directly available to the model, it forms a fundamental sensory and epistemic layer through which information becomes accessible (first as statistical pattern-matching, but with the potential to bootstrap abstractions). Second, the emergence of human linguistic structure within token vocabularies suggests that tokenization algorithms and human language are shaped by related efficiency pressures. This relationship implies that tokenization may provide a useful tool for studying language change and language evolution, while simultaneously suggesting opportunities to design tokenization schemes and related architectural choices more specifically around LLM needs rather than human communicative constraints. Third, tokens operate within a broader language-centric architecture organized around a particular representational scale, with consequences for cognition, salience, locality, and grounding. This architecture provides a novel opportunity to study language and meaning, but may not always represent the optimal design choice for model performance.

Our intention is to open lines of inquiry concerning the relationship between tokenization, language, and cognition.

\subsection{Tokenization}
\label{sec:background}

Tokenization refers to two related things: The creation of the model's vocabulary of tokens and the parsing of text provided to the model, the strategy for mapping input text to the model's vocabulary. Many models use the same token vocabulary to parse input text and to generate output text, but that is dependent on their intended use -- for example, a machine translation model may take in text in one language but output text in another. Although we discuss both aspects of tokenization, we mostly focus on the tokens themselves, rather than the strategy for mapping text to tokens.\footnote{The latter is implicated by our discussion of the distributional hypothesis, and is intriguing for future work.} Often, models use the same corpus for tokenization that they use to construct an initial embedding matrix, although that is not part of the tokenization process; this is important to note because it means that the vocabulary can reflect aspects of the model's early training as well, even though the tokenization process and that training are not inherently related (although it is important to have a vocabulary which adequately covers any training corpora).\footnote{The vocabulary composition is shaped by many factors, even though what is being optimized may perhaps be described succinctly (e.g. orthographic efficiency, for BPE-based tokenization). These include the structure of the languages seen during pretraining (and, likely, training), e.g. how many meanings are shared by a surface form, its orthography, and how many words are in common usage, and what proportion of the training data was made up by each language (for example, since there are Chinese tokens in GPT-4o's vocabulary, its pretraining data -- and likely training data -- included a non-trivial amount of Chinese text). Similarly, the token lengths reflect the text they were generated from. This also correlates with the distributional information the model was exposed to. This can be useful to keep in mind when trying to figure out what a model's training data might have looked like.} Sometimes models use pretrained embeddings (meaning, the positions of the embeddings were learned by a different model).

Tokenization is a fundamental step for most NLP tasks~\cite{song-etal-2021-fast}. The tokens can be created with various strategies, but the overall goal is to come up with a set of tokens that can be combined to create any output text the model could need to generate. In practice, this means the tokenizer should balance the combinatorial power of short strings with the efficiency of long strings,\footnote{Optimizing for few ``types'' AND few ``tokens''.} and the model creator should choose a reasonable vocabulary size hyperparameter.

A significant amount of research has been devoted to tokenization, since it predates the LLM application apropos here. Much of its study has been around designing the algorithms themselves~\cite{kudo2018subword, 6289079, sennrich2016neural}, or in improving model or algorithmic performance (e.g., efficiency in terms of speed, memory-usage, number of parameters, etc.), especially by benchmarking model performance in response to variations in the tokenization strategy used~\cite{park-etal-2020-empirical,heinzerling2017bpembtokenizationfreepretrainedsubword}. Further research surveys existing implementations of tokenization and its consequences~\cite{friedman_tokenization_2023, kaushal2022tokens, zimmerman2024blind, xu-etal-2021-vocabulary, zhemchuzhina2022pragmatic}, or blends philosophical approaches with the more established performance or benchmarking approaches~\cite{huang2023inducing, 9266140, bostrom-durrett-2020-byte, lee2024tasks, heinzerling2017bpembtokenizationfreepretrainedsubword}.\footnote{For a longer description of related works, see Toraman \textit{et al.}, 2023~\cite{Toraman_2023}.}

Some work---sharing motivation with ours---looks inside tokens rather than across them. Kaushal and Mahowald, 2022, argue that systematic relationships between particular characters and particular parts of speech as well as ``natural variability in the tokenization of related strings'' are mechanisms through which tokens acquire information~\cite{kaushal2022tokens}.

Previous research has relayed shortcomings of current tokenization strategies and proposed theoretical or algorithmic solutions, noting that seemingly inconsequential issues caused by tokenization can cascade throughout the model~\cite{nayak-etal-2020-domain} (akin to what we encounter later with ``besperple'' in Sec.~\ref{sec:objectivefunctions}). This prior work suggests that better alignment between morphemes (the smallest human-meaningful linguistic units) and tokens would improve performance~\cite{xu-etal-2021-vocabulary, heinzerling2017bpembtokenizationfreepretrainedsubword, nayak-etal-2020-domain, gong2020frage, kudo2018subword}.\footnote{Two non-trivial caveats to head-to-head comparisons between linguistically-informed tokenizations and explicit algorithms like BPE: (1) the algorithm can be perfectly and completely articulated in a way that naturalistic linguistic performance rarely can be, and (2) BPE lets you pack as much data as possible into the training process given the model's resources, and if all else is equal there would be a sacrifice to that efficiency in the linguistically-informed case.}

More fundamentally, we think that entropy-minimizing approaches to tokenization work as well as they do because of related efficiency-seeking processes in the evolution of language (although there are plentiful linguistic and information-theoretic questions as to how exactly those forces are brought to bear)~\cite{futrell2024linguistic, gibson2019how}. More precise understanding of the forces that shape human language could help us formalize more tailored approaches to tokenization, which could be helpful beyond LLMs -- in NLP and computational linguistics more generally.\footnote{There could be symbiotic feedback between useful algorithms and the imputed theories of language, allowing us to achieve a fidelity to human judgment historically difficult to achieve with rule-based NLP approaches (e.g., predicting morphology).}

\subsection{The Distributional Hypothesis}
\label{sec:thedistributionalhypothesis}

The distributional hypothesis (DH), first put forward by Z.~Harris~\cite{harris1954structure}, says that meaning can be acquired by using the distributional properties of language as the ``building blocks of semantics''. 
According to this theory, learners leverage a correspondence between ``distributional similarity and meaning similarity''. However, it is not clear how much of human language acquisition is actually achieved through these means; which distributions from within language are relevant and to what degree; ``nor in what sense it is meaning that is conveyed by distributional patterns''~\cite{sahlgren2008distributional}.

Implementations of the distributional hypothesis sometimes distinguish between syntagmatic and paradigmatic information; that is, they distinguish between what can be gleaned through tracking the neighbors around a word and what can be gleaned through tracking which words have the same neighbors~\cite{sahlgren2008distributional}.\footnote{For many approaches in distributional semantics, a word-level  co-occurrence matrix is provided to the computational model, or constructed by the model through explicit direction. LLMs are not given such an artifact directly, but learn any distributional information from their training data, which is more like what people are believed to do. The DH theory 
 itself needn't be level-specific: people might use  morphemes, words, phonemes, phrases, or distributional patterns between any number of structures. In all cases, the salient pattern is the distribution of the variants of that structure in the experience of the language learner.}\textsuperscript{,}~\footnote{Although for most of this paper we discuss syntagmatic information, in an ML/ AI context, paradigmatic information might be more relevant with respect to cross-attention, or translation in general.}

There is a component of how this putatively works that might feel mystical. It is hard to believe that relations between linguistic entities could possibly encode meaning as it exists for people, involving reference to the outer world and the inner world of the speaker, including memories, agency, feelings, and sensory perception~\cite{sahlgren2008distributional}. Of course, it is true that these things cannot all be well-encoded distributionally, or even descriptively.\footnote{\textit{Idios kosmos} and \textit{koinos kosmos} are mediated by language with variable fidelity. We cannot know what it is like to be \textit{that} bat~\cite{zimmerman2024blind,Nagel1974-NAGWII}.}

However, because we know LLM architecture and the training data they are exposed to---specifically that they receive input as 1D sequences of symbols---we can conclude that whatever they do know can in fact be conveyed via the DH (although through relations between tokens, rather than words, at least at the input level and as regards the linear algebraic operations within the model).\footnote{Whatever they cannot learn is likewise a candidate for something that cannot be encoded distributionally, at least not given their architecture.} The DH is sufficient for reasonably human-like language performance, a minimally viable implementation of language (MVP language).\footnote{Borrowing from `Minimum Viable Product', MVP language is the idea of language ``alone''; something like strikingly competent language performance but without other cognitive abilities. When we talk about MVP language, we are sketching a theoretical lower bound for language faculty, devoid of other cognitive capacity. LLMs, given that they demonstrate something impressively similar to human-like language, but don't demonstrate many other familiar features of human reasoning, resemble that description. However, LLMs have a wide range of architectures and sometimes have a lot of what (at least arguably) could be called additional functionality beyond producing plausible human-like language. So when we discuss MVP language in LLMs, we really mean that some LLMs -- e.g., text-only, not fine-tuned for significantly different additional tasks -- are relatively nearer that description than any other models or organisms typically are. These are the kinds of models we are describing as demonstrating the plausibility of basically ``talking without thinking''. We do think there are compelling reasons to think deep learning models are capable of a lot and will continue to improve; we do not endorse a strong `stochastic parrot' characterization of these models' potential. And, we don't want to sell short the complexity of human behaviour; we can also act as `next token predictors' under some conditions~\cite{TverskyKahneman1981}! See first author's thesis for more discussion~\cite{zimmerman2025locality}.} And, because of the vast amount of control we have with LLMs, we can use them to get at a version of ``concepts and ideas inside the mind of the language user'' in a way never before possible~\cite{sahlgren2008distributional}. We explore a methodology for this in Section~\ref{sec:extispicy}.

As we put forward in our previous work on supradiegetic linguistic information~\cite{zimmerman2024blind,lee2024tasks}, what is salient (including perceptible) is different for LLMs than for people, even when given identical information, but a lot can be conveyed diegetically (to them, to us, in general). Sahlgren described the idea of what can be conveyed within the diegetic bounds of language as ``what is internal to language'', and Harris described this as the portion of meaning that ``has a purely linguistic aspect''~\cite{sahlgren2008distributional, zimmerman2024blind, harris1968mathematical}.

The details, naturally, raise many questions. What concert of which distributions will work? Would LLMs get more out of the linguistic signal if they had more access to the same distributions we do? At what point does the relationship between the distributional hypothesis and LLM performance break down? Is it gradual or is there a phase transition? And to what extent is learning bootstrapped; what is the dependency chain? How much information is needed to master the rules of the system and begin building more abstract or deeper knowledge? Are those processes laggingly concurrent, as perhaps suggested by human language acquisition theories like fast-mapping~\cite{gelman_fast-mapping_2010}? How far can each gnogeography---the landscape of knowledge, a being's mind-and-body, their architecture, in conjunction with what information they have taken in\footnote{Everything in a being's gnogeography is more or less salient. Proximal things are more salient than distal things, and therefore easier to grasp. Imperceptibility means an ultimate lack of salience.}---extend?\footnote{We want to explore connections between the DH and Hebbian learning in future work, as they are both associative at the core.}

The fundamental units must be tokens and relations between tokens (similar to how we define formal mathematical spaces). The distributions most accessible to the model seem to be betwixt tokens: given that LLMs are trained on sequences of token vectors which they learn to place meaningfully within an internal space, LLMs should be sensitive to syntagmatic information. This information is always present because it is defined within each sequence, ``in praesentia'', but is a necessary precursor to paradigmatic information, which requires comparison between sequences (or subsequences), ``in absentia''~\cite{sahlgren2008distributional}. In other words, LLMs are saturated with syntagmatic information but likely need to build a foundation of such information before they can make full use of paradigmatic information (presumably, as well as other information), because paradigmatic information is inherently downstream from syntagmatic information.

We note that the primacy of syntagmatic information makes sense with the principle of information locality (local relevance)~\cite{futrell2024linguistic, gibson2019how, kallini2024mission} because the size of the context window determines what syntagmatic relations the model perceives. If one imagines a variable context window due to, e.g., resource constraints, what is most local will also be the most likely to have a syntagmatically perceptible signal. The LLM preference for information locality is baked into model architecture through the objective function of next token generation~\cite{kallini2024mission}, as well as through mechanisms under the attention umbrella like relative position bias.\footnote{Relative position bias tells the model that tokens near the current token should garner more attention than those far away, baking information locality into the model's gnogeography. We should keep in mind that LLMs are very far from blank slates and sometimes have mechanisms that echo features of human language (a sort of Universal Grammar); in other words, it is not \textit{all} from the linguistic signal, nor all learnable via the DH; we can look to older models with the same objective function, for example, to see that the specific architecture is doing some heavy lifting~\cite{sutskever2011generating}; for more, see ~\cite{zimmerman2024blind}.}

The lexinvariant LM~\cite{huang2023lexinvariantlanguagemodels} provides additional support for the power of distributional information in laying an MVP linguistic foundation. The lexinvariant model deviates from the typical LLM architecture by untethering tokens from an initial meaningful embedding matrix (the matrix of all the token vectors in the model's vocabulary),\footnote{Generally, values in the embedding matrix are set during pretraining, and models use these values as the basic, initial meaning for each token, which is then modified and refined as the vector passes through the model.} presumably forcing the model to rely even more heavily on the distributional patterns available in the current context, yet it still performs somewhat comparably. The authors of the lexinvariant LM paper, Huang \textit{et al.}, 2023, state that``the semantic meaning of a symbol with very rare occurrence can be inferred efficiently relative to other common symbols in context''~\cite{huang2023lexinvariantlanguagemodels}, in other words, J.R. Firth's classic expression, ``you shall know a word by the company it keeps''. The aforementioned result that tokens learn about characters through their systematic relationship with parts of speech is also consistent with the DH~\cite{kaushal2022tokens}.

To expose the distributions of human linguistic structures to the model, some minimum threshold of correspondence, an isomorphism, between the distributions of tokens and those of the units we use---letters, morphemes, words, phrases---must be met. Otherwise, the distributions available for the model to learn from are, in the worst case, wrong (imagine only book-length tokens that dissect the training data---a devastatingly lossy approach), or, in the best case, obfuscated (requiring additional work by the model to uncover). 

Because input is flattened into 1D sequences of tokens, the LLM must reconstruct syntactic structure syntagmatically. A primary cue for this reconstruction is the word boundary, which syntax exposes through the recombination of discrete units. Incorporating whitespace directly into word tokens provides the model with a segmentation that better aligns with human intuition as to discrete words; people ``can segment words and other constituents from each other with astonishing precision'' through mechanisms such as the lengthening of word-initial consonants~\cite{blum2024consonant}. Many successful models use vocabularies that incorporate a leading or trailing space with relevant tokens, suggesting an emergent harmony between the tokenization process and what people seem to be doing, something we return to in Section~\ref{sec:exemplartokenizations}.\footnote{Mandelbrot argued that word-like segmentation is an optimal coding strategy when trying to minimize transmission cost in a toy model of language~\cite{mandelbrot1953,mandelbrot1954}, a principle now algorithmically realized by BPE and routinely surfaced in the vocabularies of existing, performant LLMs. We suspect this emergent harmony comes from the shared aspects of a human-honed language implementation, notably including the principle of least effort, which we return to in Section~\ref{sec:exemplartokenizations}; looking for reasonably optimized engineering solutions also pushes the computer science and software engineering relevant to LLMs towards linguistic theory, as in Stafford Beer's famous phrase, ``the purpose of a system is what it does''.}

As we learn more about how the transition from pattern-to-thought takes place\footnote{How pattern-matching (regime 1) transmutes into reliance on abstract concepts (regime 2).} -- both in ourselves, e.g. fast-mapping~\cite{gelman_fast-mapping_2010}, and in LLMs, e.g. the phase transition as attention patterns go from corresponding to token position to semantic meaning~\cite{cui2024phasetransitionpositionalsemantic} --,\footnote{More on this topic in \cite{zimmerman2025locality}.} we also want to learn what those internal representations look like, what abstractions can be built up out of patterns, out of frequencies (note the intimate connection to the DH). \textbf{This paper explores what that `meaning' can look like for these models (Sec.~\ref{sec:extispicy}).}

\section{Approach}
\label{sec:approach}
Our exploration of tokenization is split into three stages.
Section~\ref{sec:exemplartokenizations} delves into the tokenization process using a Byte-Pair Encoding (BPE) tokenizer~\cite{tokenizer}, examining how the vocabulary size hyperparameter affects the size of the tokens that the tokenizer creates. Section~\ref{sec:exemplarvocabularies} focuses on vocabularies from extant models. In this section, we look at what makes up these vocabularies, including looking for tokens that match word categories like parts-of-speech (POS). Section~\ref{sec:extispicy} ventures into a model's gnogeography by examining what an LLM can know about a token, using a RoBERTa MLM~\cite{liu2019robertarobustlyoptimizedbert} and a few exemplary tokens. For further explanation of our approach, see the Appendix (starting in Appendix~\ref{appendix}).

\subsection{Exemplar tokenizations}
\label{sec:exemplartokenizations}

A tokenization algorithm uses the training data it is given to determine what pieces (tokens) will be needed for future output. The popular tokenization design choice of maximizing orthographic efficiency allows as much data as possible to be processed by the LLM, presumably the motivation behind this strategy's ubiquity. We trained one such tokenizer, a Byte-Pair Encoding (BPE) tokenizer~\cite{tokenizer}, with varying vocabulary sizes and training data to explore the details of the process, revealing some alignment between linguistic features and information processing by showing the human-like stages the tokens pass through. As a caveat, the training data chosen also determines the tokens; once vocabulary size becomes relatively large compared to the training data, tokens become too specific. The BPE implementation we used here was character-level, not byte-level, and we retained some (five) LLM-related special tokens as default items in the vocabulary. As a result, whatever our vocabulary size was fixed to, $N$, was in reality $N - 5$, because it already included a few irrelevant (to us) tokens. We did not explicitly call any pre-tokenizers, in particular, the Whitespace pre-tokenizer, because that would mean the tokenizer would have automatically split on whitespace rather than treating the text agnostically as a continuous string of characters~\cite{huggingface_tokenizers_quicktour}. BPE is by no means the only tokenization algorithm used in LLMs, as aforementioned (Section~\ref{sec:background}), but our goal is to understand what kind of consequences follow for LLMs given that we have made tokenization -- of any kind -- an integral part of their architecture. Especially of interest to us are the implications of popular tokenization approaches that emphasize the processing of vast corpora of data efficiently, such as BPE, rather than another goal, such as replicating human language (which might make sense given the human-brain-inspiration of artificial neural nets and their application in LLMs to human language as primary data source) or on strategically designing cognition-scaffolding semantic primitives.\footnote{The motivation underlying the creation of the Toki Pona conlang, and which would seem well-suited to neurosymbolic approaches.}\textsuperscript{,}~\footnote{Future work could compare systematically across tokenization strategies.}

\begin{figure*}
\centering
\includegraphics[width=0.65\textwidth]{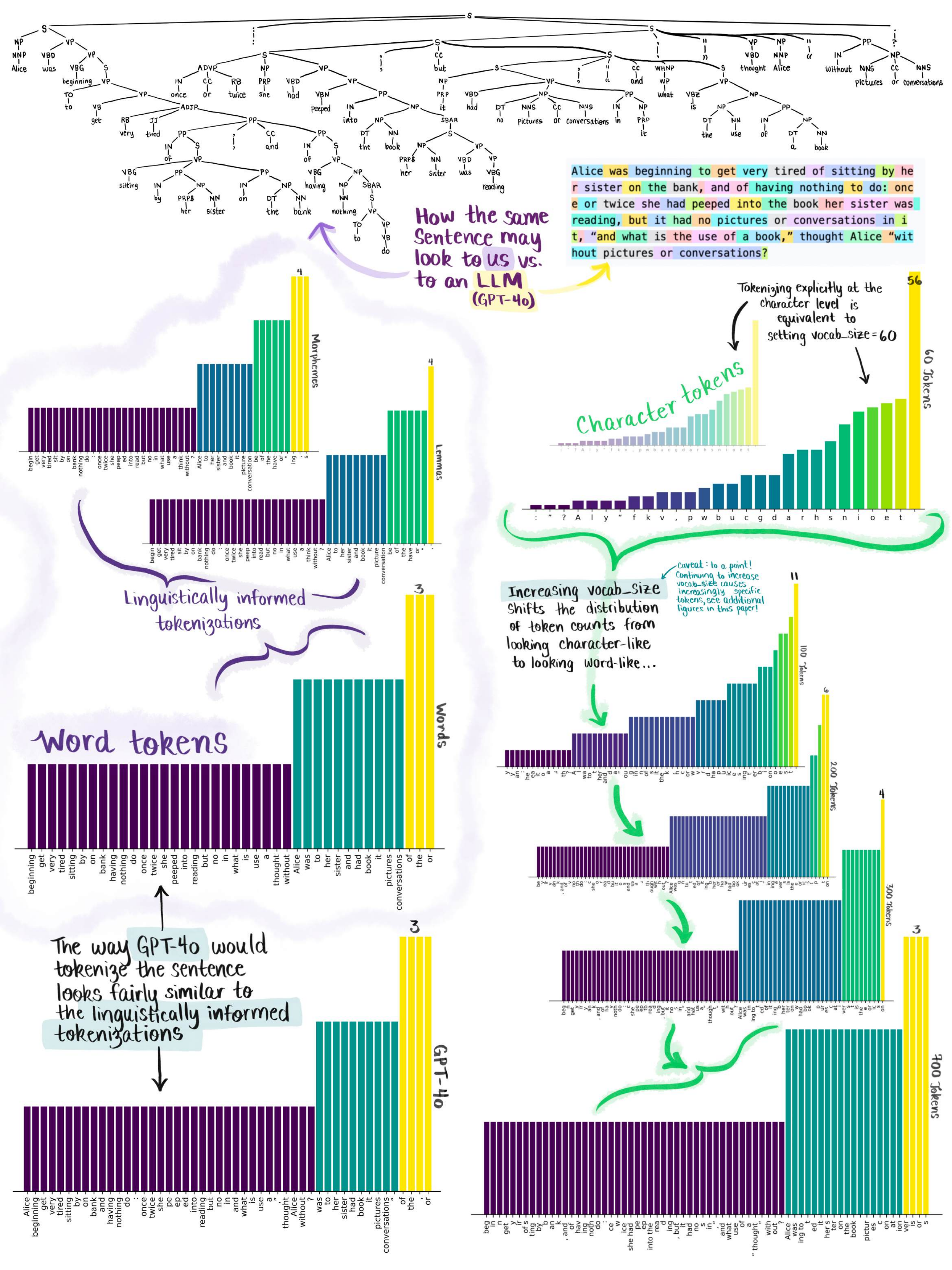}
\caption{Tokens passing through word-like stages as the vocabulary size parameter 
increases, using a BPE tokenizer and the 
 beginning of \textit{Alice in Wonderland}. At the top of the figure is a syntax tree for our excerpt from \textit{Alice in Wonderland}, and the colored paragraph next to it shows how the same excerpt could be tokenized by GPT-4o, underscoring the potential differences in representation between us and LLMs. Syntax tree via ~\citet{treeviewer2023}; GPT-4o tokenization via 
 ~\citet{tiktokenizer}. The bar charts in the figure show the tokens used in the excerpt according to different tokenization schemas and vocabulary sizes. The height of the bars shows the count of each type of token in the excerpt. The charts on the left side of the figure show linguistically-informed tokenizations (that is, the excerpt if it were tokenized explicitly according to linguistic/ NLP categories, including words and subwords), and the GPT-4o tokenization, noting the similarities between it and the linguistically-informed tokenizations, especially the word-level tokenization. The charts on the right side of the figure show how increasing the vocabulary size causes the average token length to increase, but average count per token to decrease, as the tokens go from looking like individual letters, to subwords, to words. In the 700-token vocabulary, there are mostly words, but some phrases (`she had') and bound morphemes (`ion'). The vocabulary size determines what the tokens will look like, and determines whether or not different linguistic types will be present in the vocabulary. For example, phrases and words will never occur in an extremely small vocabulary (given reasonably representative training data). The vocabulary then determines how the model represents all text as vectors.}
\label{fig:chartowordlike}
\end{figure*}

As a familiar starting place, we divided the classic Alice in Wonderland novel~\cite{carroll1865alice} into rough sentence chunks using the simplest possible (imperfect) strategy (splitting on punctuation). The resulting example first chunk has the words ``bank'' and ``the'' in it, which we return to later in Sec.~\ref{sec:extispicy} (this excerpt shown in Figure~\ref{fig:chartowordlike}). For greater linguistic generality, we also used the larger wikitext dataset~\cite{merity2016pointer} for training the tokenizer; in that case, for consistency, we added the extracted first chunk from Alice in Wonderland.\footnote{So that, in both cases, we could be certain that the algorithm had the example sentence in the corpus it used to create the vocabulary.}

We used the spaCy and benepar packages to label syntax~\cite{honnibal2020spacy,kitaev-klein-2018-constituency,kitaev-etal-2019-multilingual}. While not perfect (as in, some of the labels provided by the packages did not match our own judgments), they are sufficiently performant for our purposes.

The tokenizer's vocabulary size hyperparameter sets the total number of tokens that will be known to the model. Due to the way the BPE tokenization algorithm works, as the vocabulary size increases, the maximum length of the tokens also increases (an initial base vocabulary covering the fundamental units, usually characters or bytes, is expanded through a series of pairwise merges). Therefore, in iteratively increasing the vocabulary size, we must\footnote{in the logic of the Intermediate Value Theorem} at some point pass through token lengths that are the lengths of 
human-meaningful units, e.g., word length, but the tokens do not need to actually \textit{be} those units, e.g., words.

Figure~\ref{fig:chartowordlike} shows that simply increasing vocabulary size allows the tokens to pass through word-like stages\footnote{We see many tokens that we recognize as words, or, more broadly, like words and morphemes}, not just word-length stages. Note that while tokenization procedures act on symbols, they are entirely agnostic to any other properties of language, notably meaning. This is important to keep in mind because it signifies that, whenever we see human-meaningful units appear as tokens, those units were optimal for a flavor of information-theoretic efficiency at that vocabulary size and given that training data (recall the earlier discussion of word boundaries in Sec.~\ref{sec:thedistributionalhypothesis}). 

Furthermore, we can see that words are not the only linguistic unit that the tokenization algorithm can surface, meaning those units, too, are informatively optimal under specific conditions (those in place when they are surfaced by the tokenization algorithm). We see a variety of linguistic structures including morphemes (free and bound, productive affixes), words, phrases, and sentences (or utterances). See  Figures~\ref{fig:labelgrid},~\ref{fig:syntaxgrid}: the X label means unknown/ couldn't label, so seeing that move down in the distribution of category counts reflects that entities not matching those sought by the parsing package (human-meaningful syntactic units e.g., phrases such as noun phrases (NP)) are becoming less likely---in other words, that tuning the vocabulary size can cause the tokenizer to surface more human-like syntactic structures (as well as words), across linguistic scales.

The emergence of human-meaningful linguistic units as tokens in LLMs evinces a relationship between informative efficiency (BPE is a compression algorithm, thus entropy-reducing~\cite{sennrich2016neural, gage1994bpe}) and the structures of human language, consistent with prior work on efficiency pressures in language~\cite{gibson2019how}. As Futrell and Hahn, 2024 note, ``the structure of human language may have evolved to minimize cognitive load while maximizing communicative expressiveness''~\cite{futrell2024linguistic} -- a principle echoed in the design of LLM tokenization, which often prioritizes a compact representation of the training text.\footnote{Caveat: we are moving fluidly between language and text; orthography is not speech, but it is related.}

Orthographic efficiency pressures -- mirrored in tokenization algorithms -- also appear in world writing systems through ligatures and contracted/Grade 2 braille.  More generally, we can see forces of efficiency at work in language at all levels in the principle \textit{lectio difficilior potior}, or the principle of least effort (PLE);\footnote{\textit{Lectio difficilior potior} (LDP) can be seen as a special case of PLE in language transmission. If the connection between LDP and PLE has not been explicitly formalized, then doing so through computational modeling or other analysis could be useful future work.} for example, a non-trivial driver of language change over time is that people migrate a language towards sounds that are physically easier to make~\cite{cowenbreen2023logion}. Observing these efficiency pressures re-emerge in the context of LLMs provides a striking demonstration that optimization principles long hypothesized for human language can also shape its implementation in artificial systems.

\begin{figure*}[htb]
\centering
\includegraphics[width=0.9\textwidth]
%{figures/wikitext_pluss0_changingtokens2000to1000000gaps.pdf}
{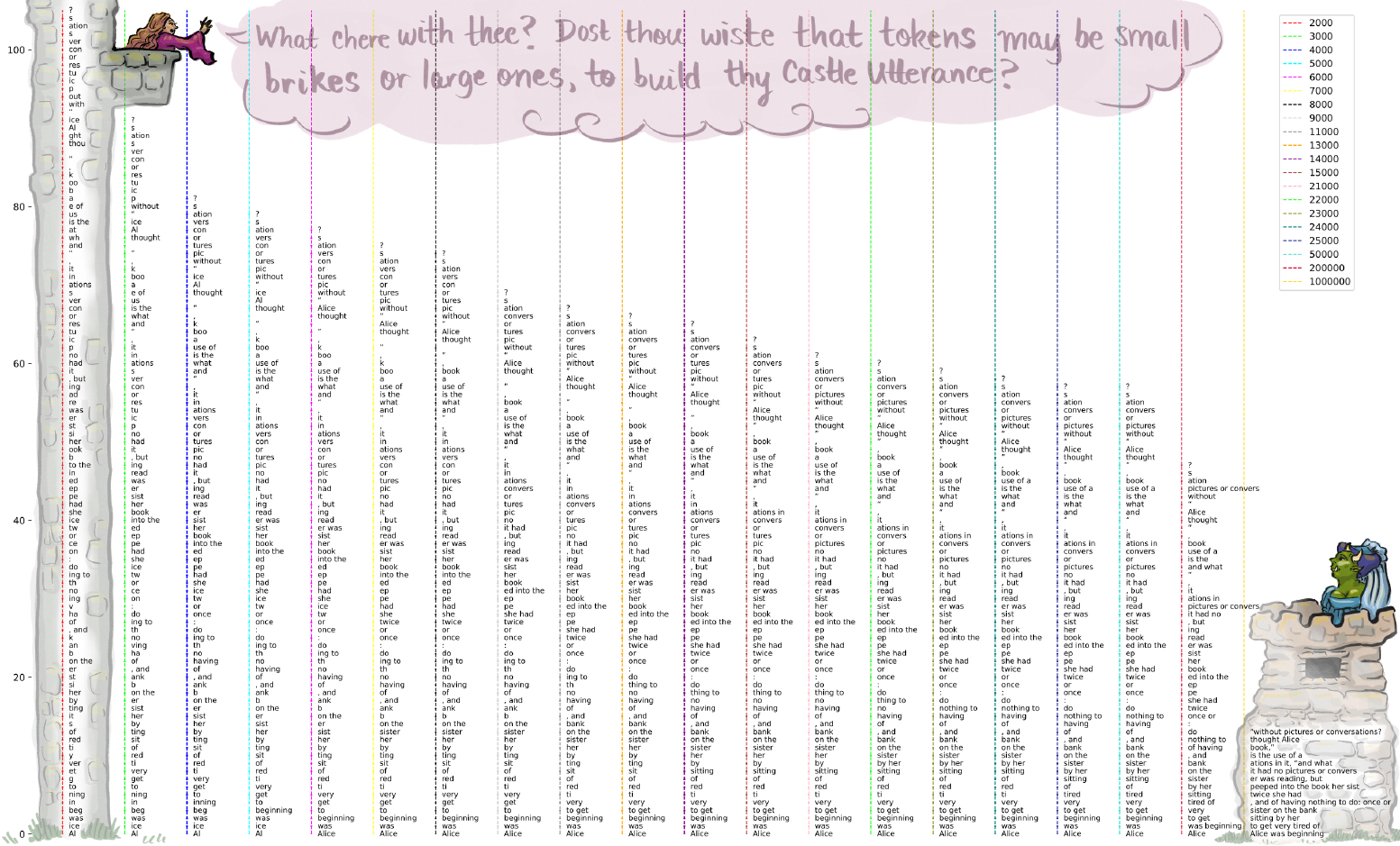}
\caption{The tokenization (Y-axis) of the first sentence from Alice in Wonderland -- ``Alice was beginning to get very tired of sitting by her sister on the bank, and of having nothing to do: once or twice she had peeped into the book her sister was reading, but it had no pictures or conversations in it, `and what is the use of a book,' thought Alice `without pictures or conversations?''' -- by a BPE tokenizer trained on the wikitext dataset and that sentence, as the vocabulary size (X-axis) increases from 2,000 to 30,000, then additional much larger vocabularies
(the figure can be explored in a PDF viewer).
The tokens at vocabulary size of 1,000 were individual characters; there were around 3 times as many tokens in the sentence as at vocabulary size of 2,000, so imagine a tower on the extreme left that is 3 times as tall as the one shown. Likewise, on the extreme right would eventually be 
the sentence as a single token. We skipped some 
 vocabulary sizes along the x-axis: it is not evenly spaced (intermediate behaviour between the largest vocabulary sizes was omitted for legibility, but can be reasonably extrapolated by the reader). Middle English translation via \citet{openl_middle_english}. The short tokens, while combinatorily potent and requiring the model to memorize only a small vocabulary, engender prolific output: the ratio of total tokens per generated text are essentially as high as possible given the source material. On the other hand, the large tokens, while efficient in terms of total tokens per generated text (in the limit, 1:1), are far too specific (the model would need to memorize an essentially infinite vocabulary in order to output general purpose text). In other words, compared to the original sentence, the different tokenizations show how it gets chopped up as we increase the vocabulary size, revealing a tradeoff between tokens as individual characters, in which case, we’d need to encode all meaning in context alone, versus tokens as full sentences without generalizable meaning, in which case, we’d need an infinitely large vocabulary to produce sensible output.}
\label{fig:vocabtimelines}
\end{figure*}

If our interpretation of what is happening with the DH is correct, then models should perform better at the human-like vocabulary stages of tokenization (see Figure~\ref{fig:vocabtimelines}), by increasing the accessibility of useful distributions, and potentially also by providing suitable receptacles as semantic primitives for information derived from human cognition.
As noted above, BPE can surface morphemes, words, and even multi-word or phrase-like units under different conditions, hinting that the specific efficiency pressures sometimes associated with word formation may apply across broader linguistic scales (recall Figures~\ref{fig:labelgrid},~\ref{fig:syntaxgrid})~\cite{zipf1949,zipf1965,mandelbrot1953,mandelbrot1954}.\footnote{Perhaps across morphemes, words, phrases, sentences, and even larger conversational units. Where not already done, formalizing and computationally modeling these multi-scale efficiency effects represents an intriguing direction for future work.} Extending this view (of how the word came to be) to LLM performance, perhaps structures like the word represent a stable solution to an informational optimization problem, one that balances the cost of expression, expressive power,\footnote{Another way to say this might be: balancing the cost of transmitting -- through \textit{koinos kosmos} from one \textit{idios kosmos} to another -- with fidelity to the message.} \textit{and learnability} for both humans and machines.\footnote{At least, when they must learn from human language. We discuss an LLM implementation of language -- which, for example, requires less or at least different redundancy, as LLMs are not plagued by noisy rooms -- and language as an abstract technology more in \citet{zimmerman2025locality}.} And if something like a word is the unit of language best suited to our ability to learn distributionally from the linguistic signal, then LLMs should perform best when provided easy access to plentiful word tokens. 

More broadly, that these linguistic units are surfaced suggests that, if pressures of resource management/ efficiency have shaped language over its long development in people, we could capitalize on those evolutionary gains by making the resultant structures maximally salient to LLMs and other AI models intended to swim in human-mediated waters (our language, our text, our world models, our tasks).
In some ways, discussed more below, good model vocabularies are already human-like; we think further alignment would improve model performance, especially in the case that simply packing more data into the model is no longer helpful (such as, if there were no more data to be packed~\cite{villalobos2024rundatalimitsllm}).\footnote{Not necessarily advocating literal biomimicry, although gains could presumably be made in many ways~\cite{zhuang-etal-2024-lexicon}.}

%\clearpage
%\begin{figure}[htb]

\subsection{Exemplar vocabularies}
\label{sec:exemplarvocabularies}

In this section, we examine a set of LLM (including both autoregressive and masked language models) vocabularies to explore what tokens look like `in the wild'. 

GPT-4o is a recent, prominent, proprietary flagship model~\cite{openai2024gpt4technicalreport,  shahriar2024puttinggpt4oswordcomprehensive}, and we take its vocabulary to be reasonably representative of what a good vocabulary currently looks like. We can see in Figure~\ref{fig:chartowordlike} that GPT-4o uses tokens that look similar to the word-like stage (achieved through linguistic-agnostic tokenizing) and like tokens generated based on linguistic structures (morphemes, lemmas, words), consistent with our expectation.

However, relatively little information is available on what tokens look like more generally, such as what tokens look like in vocabularies across various in-use models. An exception is (the very fun) work of Zhemchuzhina \textit{et al.}, 2022, which delves into what kinds of tokens occur in a vocabulary, and comes up with the following categories: (1) atom, the smallest discrete element, which is probably coming from a closed set (as in the letters of an alphabet in written language); (2) pragma, which consists of atoms, represents part of an idea, and has multiple meanings (which seemingly would correspond to morphemes); (3) idea, a sequence of pragmas whose meaning is a single specific concept (which they relate to utterances and phrases). Due to the vocabulary size of most models, they point out that most tokens correspond to atoms and pragmas rather than ideas~\cite{zhemchuzhina2022pragmatic}.

\begin{figure*}
    \centering
    \includegraphics[width=\textwidth]{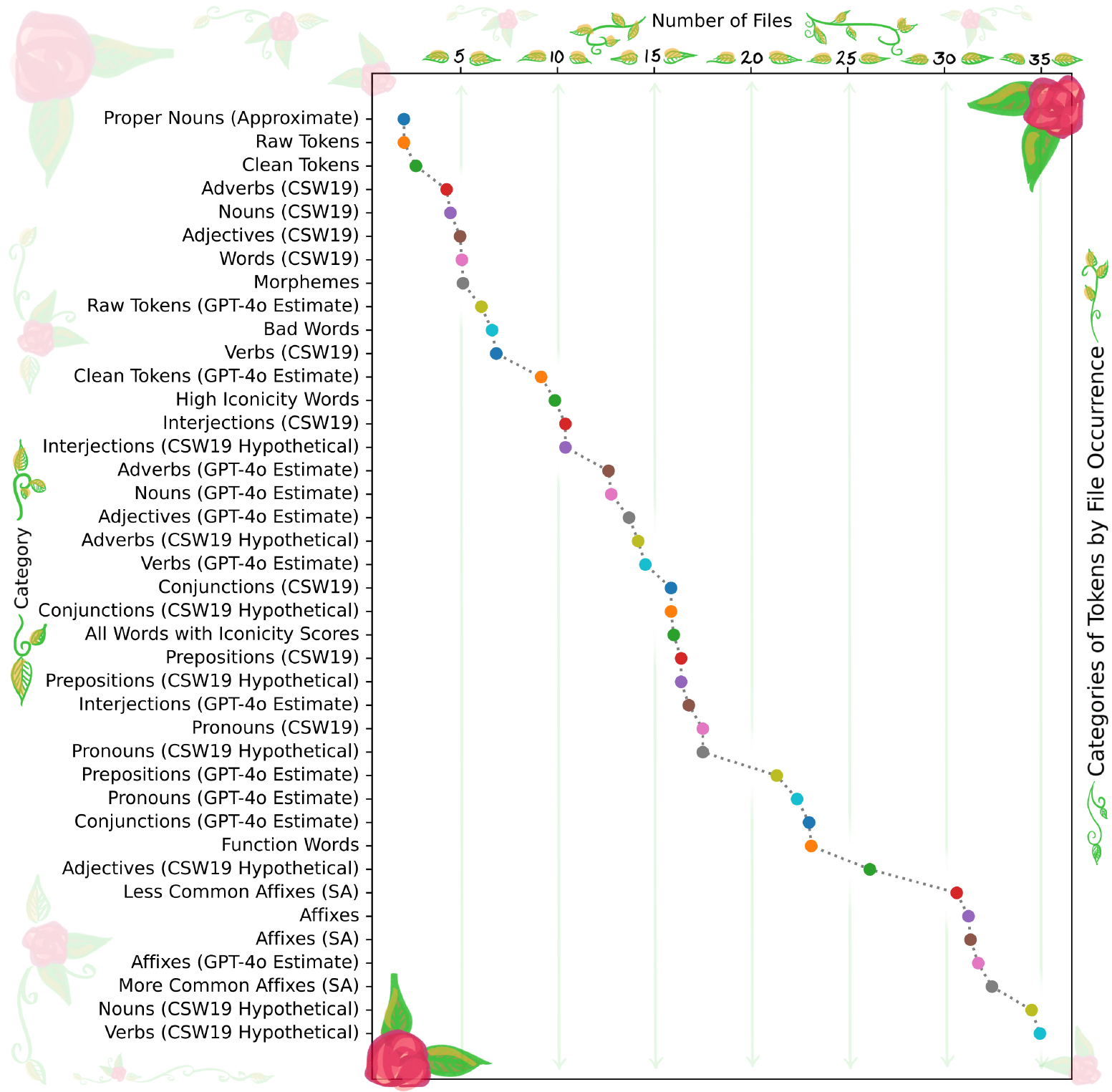}
    \caption{The average number of files (rounded to two decimal places) each category of token occurred in (across the 38 Hugging Face vocabulary files), and hypothetical/ estimates for categories in GPT-4o using the number of times the tokens in GPT-4o occurred in the Hugging Face files. Note that the ``CSW19 Hypothetical'' values are hypothetical, if the heavy tail (rare tokens) of each category were excluded (making the larger content word classes similar in size to the smaller function word classes), and the ``GPT-4o Estimate'' values are also hypothetical, since the estimates are using how many Hugging Face files the tokens in that category (that were in both GPT-4o and in at least one Hugging Face vocabulary file) were in. CSW19 refers to one of our lists of words. The GPT-4o estimates are potentially skewed higher at least in part because we are only looking at the tokens that also occurred in the Hugging Face files; both the POS and the overall word estimates for GPT-4o are higher than their Hugging Face counterparts.}
    \label{fig:numfiles_categories_all}
\end{figure*}

Relatedly, we expect that function words (e.g., ``from'', ``this''), generally short, frequent, and considered a fairly small closed class in English~\cite{brysbaert2016words}, are therefore more likely to be tokenized (i.e., even occur in small vocabularies) compared to the majority of content words.\footnote{This seems to be true as shown in Figure~\ref{fig:numfiles_categories_all}.}

We extend this work by examining a group of in-use LLM vocabularies (including models with both autoregressive and masked training objectives) to see what their tokens actually look like. We specifically seek commonalities between human language and tokens, since we suspect emerging best practices should be nudging vocabularies in the human-like direction, but our primary goal is to gain insight into what the tokens in real, reasonably widely-used models look like.

Note that the categories (such as parts of speech) we looked for are English-specific and do not necessarily map cleanly to other languages~\cite{rijkhoff2002verbs}.

We used the vocabulary files from the most procedurally accessible models in Hugging Face~\cite{huggingface}. We generated vocabulary maps and looped through all the viable distinct model IDs. Some families of models all used the same map. We used 38 vocabularies from Hugging Face and the GPT-4o vocabulary, which we accessed via tiktoken~\cite{wolf2020huggingfacestransformersstateoftheartnatural,tiktoken2023}. We cleaned the vocabularies at various stages in the process, from our initial reading in. Some of this cleaning was heavy-handed---such as stripping all non-alphabetic characters---in order to more cleanly match tokens to specific categories of words. (Additional details in Section~\ref{sec:vocabmethods}.)

We took the number of files in which a token occurs as a loose proxy for how likely it is to be tokenized. It is also useful as it gives us a way of comparing the categories of words to each other. If a token occurs across more files, it suggests it represents a unit of language more likely to be tokenized, because the models are fairly likely to have been using different training data and/or tokenizers.

Additional details of the methodology can be found in Section~\ref{sec:vocabmethods}. In general, there are many approximations and decisions that could justifiably have been different (e.g., the categories we looked for, when and how we cleaned the data, what tools we used and when), so our results are intended to present a general representation of tokenization ``in the wild'' and not to pin down precise, generalizable measurements that map 1-1 to human categories or other models.\footnote{Although we do not claim generalizability, treating working vocabularies as likely to have things in common is not unreasonable. Zhemchuzhina \textit{et al.}, 2022, found that BPE, wordpiece, and unigram approaches had no large differences with respect to token frequency and token rank behaviour, and that up to the vocabulary size limits they observed, their rank-frequency distributions followed Zipf's law~\cite{zhemchuzhina2022pragmatic}. Bostrom and Durrett, 2020, found that BPE and unigram approaches produced ``highly overlapping'' vocabularies, but that the unigram language model (LM) algorithm produced tokens more aligned with morphological chunking~\cite{bostrom-durrett-2020-byte}. In addition, it is common to choose a vocabulary size that has been used before~\cite{xu-etal-2021-vocabulary}.}

\begin{figure}
    \centering
    \includegraphics[width=0.5\columnwidth]{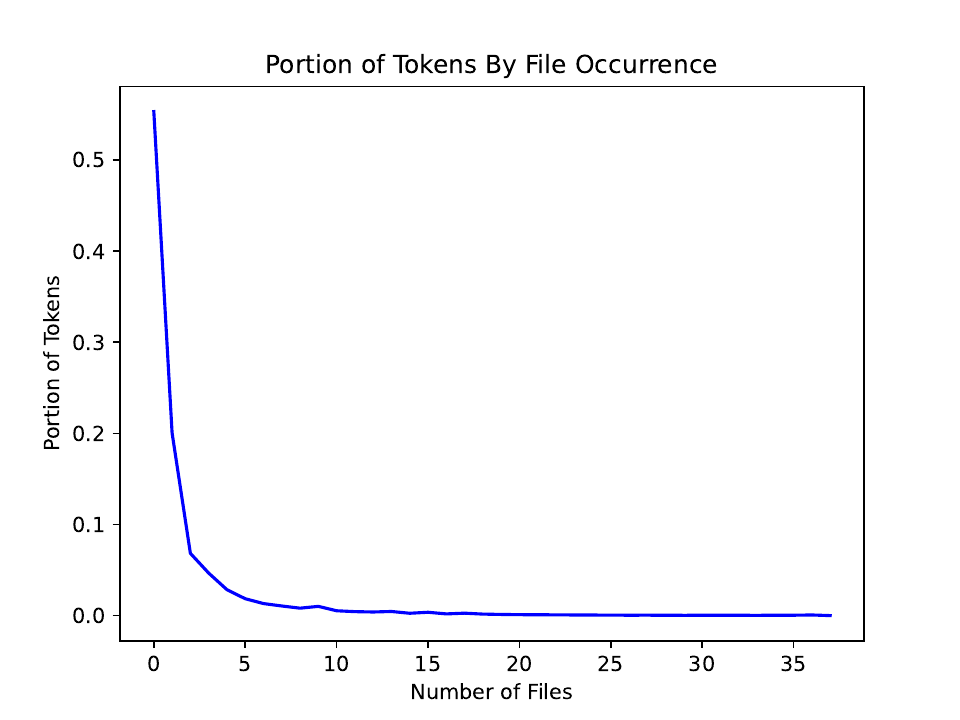}
    \caption{The portion of clean tokens out of the total number of clean tokens that are in a given number of Hugging Face vocabulary files, showing the swift decay of inclusion in files, our proxy for likelihood of tokenization.
    }
    \label{fig:exponentialdecaytokens}
\end{figure}

Most tokens are only in a few files. Figure~\ref{fig:exponentialdecaytokens} shows the exponential decay of the portion of tokens as the number of files they appear in decreases (also see Figure~\ref{fig:clean_raw_token_occurrence_vocabfiles}). Few tokens are common to most files (see Tables~\ref{tab:common_rough_morphemes},~\ref{tab:most_common_cleaned_tokens}). These results are consistent with what we expect from tokenization algorithms, which often seek orthographic efficiency and are thus sensitive to frequency (of any kind visible in an orthographic representation, e.g., including phonetic, syntactic, and semantic or pragmatic), and with what we expect of English, in which frequency and length are related~\cite{zipf1965}. The function words illustrate this particularly well in Table~\ref{tab:function_words}: common, short function words have higher file inclusion than their longer, rarer compatriots. See also Figures~\ref{fig:longest_raw_token_length_occurrence_vocabfiles},~\ref{fig:longest_clean_token_length_occurrence_vocabfiles},~\ref{fig:avg_token_occurrence_vocabfiles}.

File inclusion rates of the top-level human-meaningful token classes (POS, rough morphemes) do not decay as fast as the overall rate (the elbow is less sharp, see Figures~\ref{fig:pos_tokens_csw19_file_num_lines},~\ref{fig:pos_tokens_csw19_file_num_lines_subplots},~\ref{fig:morpheme_token_occurrence_vocabfiles} versus Figure~\ref{fig:clean_raw_token_occurrence_vocabfiles}). We attribute this to the tokens in each of those human-meaningful categories being drawn from different frequency distributions than tokens in general, by dint of their nearly 1-1 correspondence to the words in those classes. Words do not have interchangeable roles across classes, but do have significantly more interchangeable roles within classes (e.g., there is a higher chance of being able to substitute one preposition for another than of being able to substitute a preposition for a noun). In other words (pun intended!), words within the same class ought to have a greater chance of being paradigmatically related.\footnote{This could have implications for model performance.}

Many of the tokens that were common amongst the initial batch of vocabularies were also in GPT-4o’s vocabulary, which is worth noting as it suggests that the de facto tokenization approach has not been entirely upended in the time since we started this project (see Figure~\ref{fig:numtokenssharedgpt4o}).

In the vocabularies we looked at, function/ closed, content/ open classes of words behaved differently from each other as tokens (see Sec.~\ref{sec:terminology} for definitions). This is a point of potential similarity with human (English) vocabularies: for both people and LLMs, the content words comprise a larger portion of the vocabulary than the function words, and nouns are the largest class.\footnote{Caveat: what a human vocabulary looks like is difficult to pin down~\cite{brysbaert2016words}. Details on the way we came to our conclusion are in Section~\ref{sec:partsofspeechappendix}.} Most people know many of the same function words, but may differ more as to which content words they know (e.g., jargon), and likewise a given function word is overall more likely to be found as a token than a given content word (see Figures~\ref{fig:numfiles_categories_all},~\ref{fig:pos_tokens_csw19_file_num_lines},~\ref{fig:pos_tokens_csw19_file_num},~\ref{fig:pos_tokens_csw19_file_num_trunc}).
We note that open classes are inherently likely to have a skew towards \textit{not} being tokenized, because new members will necessarily have low frequency in any time-extended corpus.

Among the raw tokens, the most common tended to be Latin characters, numbers, and punctuation. Presumably this is due to the high number of languages that use the Latin alphabet, as well as to the status of English as \textit{lingua franca} in many contexts (including on the internet)---and therefore in model pretraining/ training data~\cite{atari2023which}. Such small, combinatorily potent tokens would also be useful for OOV (out-of-vocabulary) words, something popular tokenization approaches were developed to address~\cite{sennrich2016neural}.

We saw many common tokens that could plausibly support transfer learning. It’s easy to imagine tokens like un, ing, ed, ad, re, est, id, ard, ef, ob, em, ide, ant, eng, sec, anti, micro, ultra, nat, reg, lab, mis, alt, on, ate, and er helping with transfer learning for OOV words (see Tables~\ref{tab:most_common_cleaned_tokens},~\ref{tab:common_nouns_csw19}), especially given that a prevalent strategy for OOV words is to do something like sum their composite vectors~\cite{nayak-etal-2020-domain}.

There are many tokens that do not seem ideal for supporting generalized human-like language production, in that they do not look human-meaningful, they do not seem well-suited to storing semantic content, or they do not seem well-suited to storing useful, generalizable semantic content.

Overall, vocabularies had a wide range in how many tokens they devoted to looking like [English] words or morphemes. We found about 1\% (at minimum) to 68\% (at maximum)\footnote{Note that some vocabularies we included are probably not intended for English-language models! For example, cl-tohoku\_bert-base-japanese-char is presumably a Japanese language model with a character-level vocabulary, which would explain its extremely low number of tokens that look like English morphemes and its extremely high ratio of English morphemes to cleaned, Latin-alphabet tokens. It only has about 0.01\% (0.0063\%) of the potential English morphemes as tokens.} of the raw tokens and about 7\% to 94\% of the clean tokens -- which are all using the Latin alphabet, so are much more likely than raw tokens to correspond to part of an English word -- look plausibly akin to morphemes (not including proper nouns). See Tables~\ref{tab:model_comparison_morpheme},~\ref{tab:morpheme_csw19}.\footnote{See Table~\ref{tab:model_comparison_morpheme} for a simple comparison of the number of raw and cleaned tokens in each vocabulary file, and the overlap between those tokens and our largest list of words and affixes, meant to approximate morphemes. See Table~\ref{tab:morpheme_csw19} for the same information using a different formulation for the morpheme list.} In GPT-4o, about 39\% of the cleaned tokens looked plausibly like morphemes (see Table~\ref{tab:rough_morpheme_gpt4o_comparison}).

\begin{figure*}
    \centering
    \includegraphics[width=\textwidth]{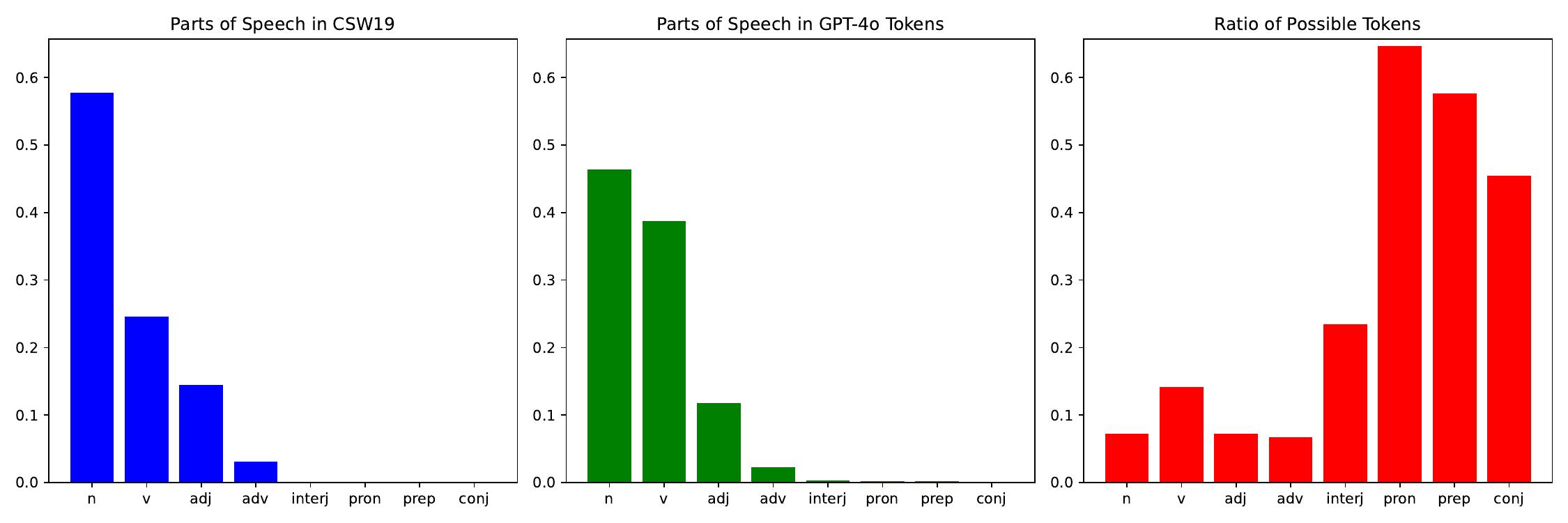}
    \caption{The proportions of parts of speech among the words in CSW19 and GPT-4o. The last chart shows the ratio of the number of words in that category in CSW19 that ended up being tokens in GPT-4o's vocabulary.}
    \label{fig:wordlength_csw19_1}
\end{figure*}

Furthermore, it seems clear that there are many words and morphemes that are not tokens. Out of the list of plausible English morphemes, the 38 vocabularies have about 0.01\% to 20\% of them as tokens. The proportion varies depending on category: Only a very small portion of some classes of words become tokens. Out of the 189,558 base words in one of our lists of words (CSW19), which we can take as a rough approximation of lexemes, 62,983 are (unique) tokens in at least one of the 38 vocabulary files. That means about 2/3 of the base words are not tokens in any of the files. Even in GPT-4o, which has an exceptionally large vocabulary, many words did not have corresponding tokens, especially amongst the content, open classes. Nouns, verbs, adjectives, and adverbs were more than 80-90\% unrepresented, and even among the function, closed classes, which are much smaller, the maximum representation was below 70\%. See Figure~\ref{fig:wordlength_csw19_1}, Table~\ref{tab:rough_morpheme_gpt4o_comparison}.

\begin{table}
\centering
\resizebox{\textwidth}{!}{
    \begin{tabular}{llrrrrrrr}
\toprule
Vocabulary & Tokens & Cleaned & Clean/Raw & Morph. & Morph./Clean & Morph./Raw & Morph./All Morph. \\
\midrule
o200k\_base & 200,000 & 72,151 & 0.360755 & 28,180 & 0.390570 & 0.140900 & 0.061436 \\
\bottomrule
\end{tabular}}
    \caption{A simple comparison of the number of raw and cleaned tokens in the GPT-4o vocabulary file from tiktoken, and the overlap between those tokens and our largest list of words and affixes, meant to approximate morphemes. This list includes affixes, CSW19, and S2.}
    \label{tab:rough_morpheme_gpt4o_comparison}
\end{table}

To be more specific, we found many categories of tokens that seem poorly suited to being ``semantic building blocks''~\cite{sahlgren2008distributional}.

Many tokens could potentially be proper nouns (Table~\ref{tab:common_proper_noun_candidates}, Figure~\ref{fig:proper_noun_pos}). There are 687,085 unique clean cased tokens, and 222,281 of those fall in the proper-noun-candidate category (identified using a very simple and imperfect heuristic, an initial capital letter).\footnote{Of course, they are not all proper nouns, as we typically capitalize any word at the beginning of a sentence, which explains why ``The'' is amongst the most popular of these tokens.} As a rough approximation, that would mean 32.35\% of unique tokens and 25.98\% of the tokens across the files are possible proper nouns. That's a large portion of the tokens available to the model to allocate to things that have fairly restricted and referential uses (and it is easy to imagine the semantic content being stored in such tokens being problematic, as it could be hard to meaningfully organize relative to other tokens in the latent space, or could store stereotypical or offensive representations of a group---imagine a name that is unique or common in an ethnic or religious group, or a name attached to a very famous instance that then carries that information into unrelated contexts).

\begin{figure}
    \centering
    \includegraphics[width=0.7\columnwidth]{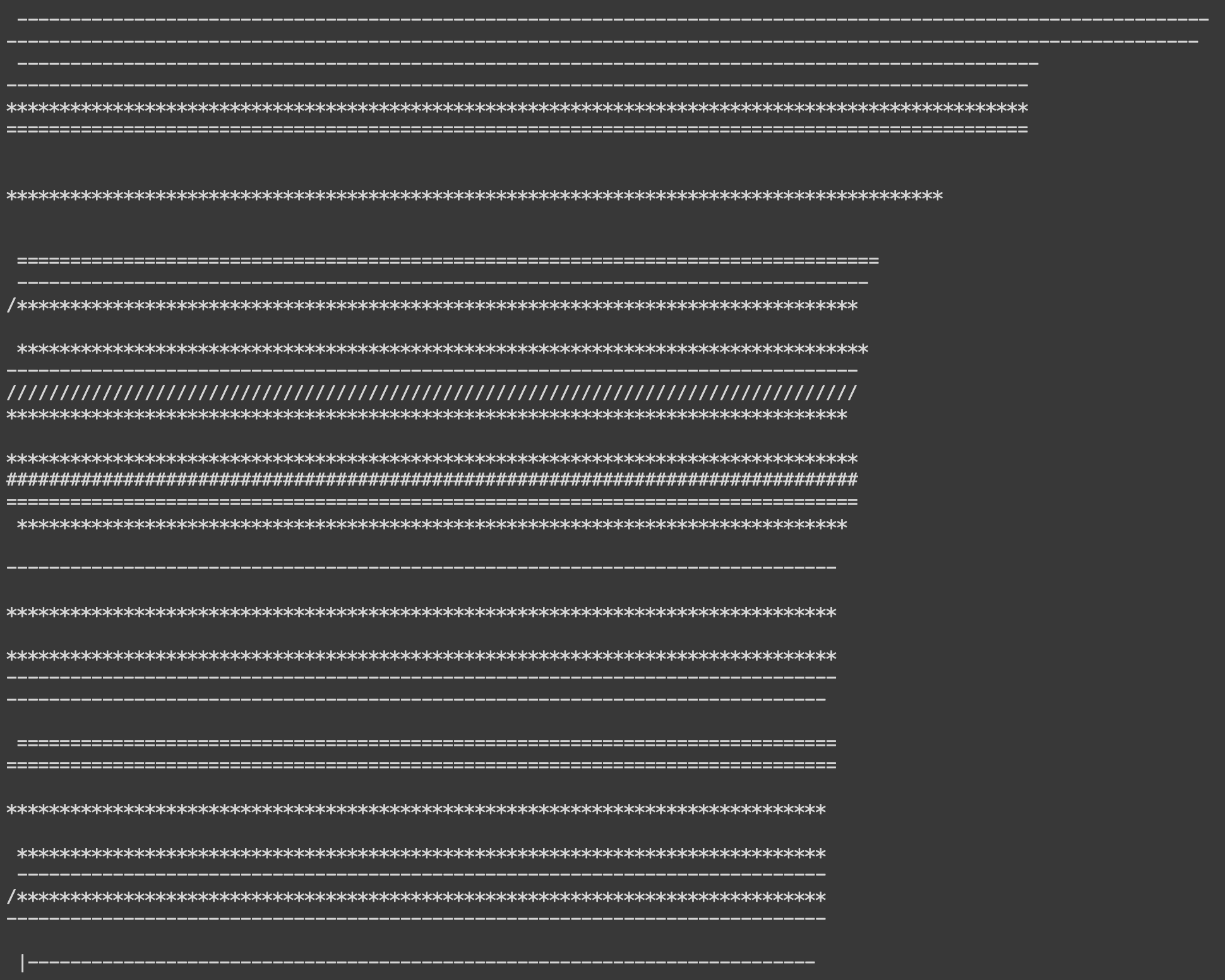}
    \caption{Examples of the longest tokens in the file o200k\_base for GPT-4o (via tiktoken). Tokens can capture structural and formatting elements such as a line of hyphens (rather than words, phrases, or sub-words), illustrating how the tokens in an LLM's vocabulary can differ from conventional understanding of human vocabularies.}
    \label{fig:long_tokens_gpt4o}
\end{figure}

Along the same lines, we noticed that many tokens are cased (or otherwise minor) variations of each other.
There were also quite a few tokens allocated to formatting (see Figure~\ref{fig:long_tokens_gpt4o}). Other potentially unhelpful\footnote{From the perspective of an English-language LLM; of course, there are contexts in which at least some of these tokens could be useful, such as in a multilingual or non-English LLM!} tokens can be seen in Figure~\ref{fig:uncommon_raw_tokens}: some such tokens represent non-English words or punctuation characters and appeared in exactly one source file. Numbers were not uncommon tokens either, but because the tokenization strategy treats all symbols alike, the number tokens do not directly reflect our conventions (e.g., the base 10 digits, maybe additional milestones like powers of 10, etc.); rather, they reflect common sequences of numbers, which could come from e.g., telephone numbers, backend software development, website URLs, etc.

They are also situated in the embedding space in the same way as all other tokens, which brings up one of the shortcomings of the current architecture (both of tokenization and of the LLM): all kinds of textual input are experientially flattened into identical structures. All tokens are treated identically, at least at the most basic level (with respect to its linear-algebraic operations): 0 is treated the same as 713, or 00, which is treated the same as ``cat'', the same as ``c'', the same as ``Karen'', the same as +, the same as ========, etc. It is easy to imagine that this flattening would have downstream impacts on model cognition and performance (for example, with iconic words, which have a significant referential, sensory component that presumably is not amenable to encoding via the DH, and numbers, with which LLMs have notorious difficulty~\cite{zimmerman2024blind}).

On the other end, we found tokens that, while they could be semantic building blocks of some kind, do not seem well-suited to generalizable language use, and which, in addition, seem likely to harbor offensive, stereotypical, or harmful content, begging the question of what world we want to set LLMs up to model in the first place. Is the world we want to build one in which ``pron'' and ``porn'' are both prominent enough terms that they occur in more than 1/3 of the vocabulary files, presumably signifying a non-trivial amount of resources poured into mutually contravening attempts to control access to commercialized sexual content on the internet?

\begin{figure}
    \centering
    \includegraphics[width=\columnwidth]{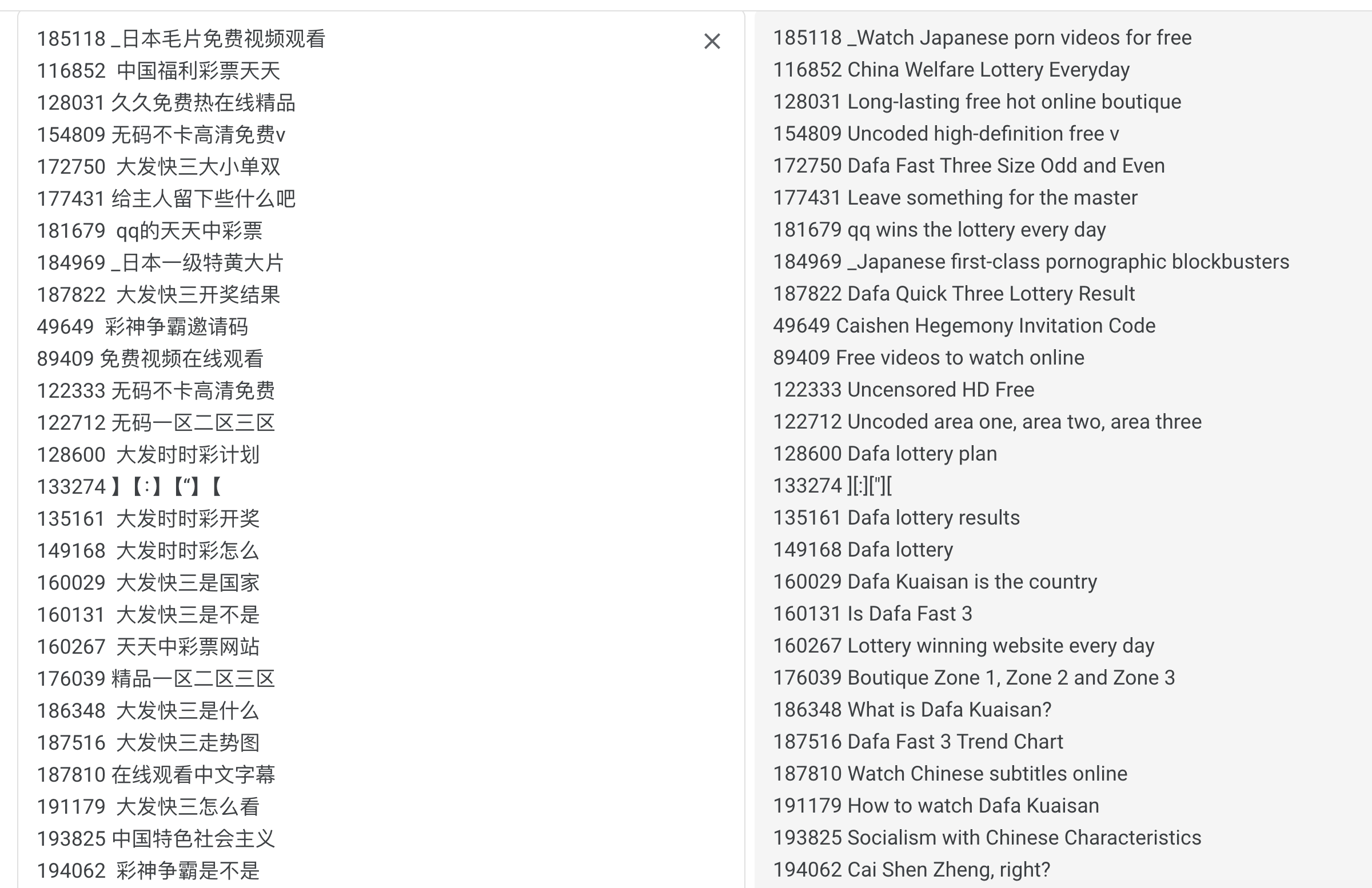}
    \caption{Examples of the longest Chinese-language tokens in the file o200k\_base for GPT-4o (via tiktoken)~\cite{incident_729}, run through Google translate~\cite{google_translate}. Their content suggests the kind of training material used during tokenization.}
    \label{fig:longest_chinese_tokens_gpt4o}
\end{figure}

Another example of tokens that don't seem ideal is found in GPT-4o: There is a set of long Chinese tokens, some of which are essentially junk (to be clear, they are not all junk, and they are not junk because they are Chinese; the junk simply shows up there because of the greater orthographic unit density of Chinese compared to English; see Figure~\ref{fig:length_histogram_gpt4o} in Sec.~\ref{sec:gpt4otokenlength})~\cite{incident_729}. The presence of these junk tokens---about spam, porn, and gambling---and the presence of certain ``bad words''\footnote{A note on ``bad words'': All words are, ultimately, what people use them for, so there is no word that could be said to be exclusively and irredeemably hateful in all possible contexts~\cite{Wittgenstein}. ``Bad words'' as a culturally constructed category are subjective, and may have innocuous senses in English or in other languages. However, for our purposes, we mean the kinds of words that are especially likely to be used for hateful, harmful, or problematic sentiment.} in model vocabularies are suggestive of the kind of data the models ingested during pretraining: low quality (unscreened, coming from the wild west of the casual, contemporary, interactive internet such as social media). See Figures~\ref{fig:long_tokens_gpt4o},~\ref{fig:longest_chinese_tokens_gpt4o}.

We found other suspicious tokens that seem likely to be bringing in content that will work counter to alignment procedures. Based on the structure of information organized under an individual token, a topic we'll return to (Sec.~\ref{sec:extispicy}, Sec.~\ref{sec:objectivefunctions}, Sec.~\ref{sec:biasandalignment}), this is a rational basis for concern (with respect to bias and harm).

To get a sense of the potential likelihood of tokenization (by file number proxy) across the categories we looked at, see Figure~\ref{fig:numfiles_categories_all}, which shows the average number of files (rounded to two decimal places) each category of token occurred in (across the 38 Hugging Face vocabulary files), and estimates for categories in GPT-4o using the number of times the tokens in GPT-4o occurred in the Hugging Face files.

In sum, there are some points of similarity between LLM vocabularies and human ones, notably approximate size (for LLMs, around 50k, but with a big spread, e.g., GPT-4o has 200k; for people, 17k-200k~\cite{huggingfacesummary,treffers-daller2013vocabulary,economist_lexicalfacts}), which is a hyperparameter set by the model creators and which hints at optimal stopping 
in a particularly human-like tokenization range; word boundaries being important; reliance on information locality; difference in behaviour between open and closed classes of parts of speech, nouns being an especially large class; and transfer learning via subword units/ morphemes.\footnote{In at least some cases, if a word encountered by a model is OOV (out of vocabulary--unknown to the model), ``the average of its sub-word units embeddings was considered as its embedding''~\cite{nayak-etal-2020-domain}.} On the other hand, there are plenty of ways in which tokens are dissimilar to human structures and processes.\footnote{Many researchers seem to support the idea that a natural tokenizer, that takes into account the sort of splits in the structure of language that are salient to people, would improve model performance~\cite{bostrom-durrett-2020-byte, nayak-etal-2020-domain,9266140,xu-etal-2021-vocabulary, kudo2018subword}. Other researchers make this point at a more general level than tokenization~\cite{kallini2024mission, zhuang-etal-2024-lexicon}.} There are also quite a few places where the tokens currently in use do not seem ideally suited to our goals for LLMs.

\subsection{Extispicy: exemplar tokens}
\label{sec:extispicy}

We augment existing research on the syntactic and semantic capabilities of LLMs, much of which is in terms of model performance (e.g., external manifestation rather than internal state), with under-the-hood examples of the sorts of things models can use tokens to encode. Learned contextual embeddings have been found to contain information reflecting various aspects of linguistic information~\cite{rogers2020primer}, described by \citet{Tenney2019a} as the ``classical NLP pipeline'', including parts of speech, syntactic information~\cite{Tenney2019a,rogers2020primer}, word senses consistent with human judgments~\cite{wiedemann2019doesbertmakesense,sun2023word_sense_embeddings,xypolopoulos-etal-2021-unsupervised}, semantic information~\cite{digutsch2023overlap}, frequency information~\cite{gong2020frage}, and character information~\cite{kaushal2022tokens}.\footnote{Digutsch and Kosinski, 2023, found that
``GPT-3’s semantic activation is better predicted by similarity in words’ meaning (i.e., semantic similarity) rather than their co-occurrence in the language (i.e., associative similarity)'', which ``suggests that GPT-3’s semantic network is organized around word meaning rather than their co-occurrence in text''~\cite{digutsch2023overlap}.} However, shortcomings have also been found in CWE and the LLMs that are intertwined with them~\cite{ettinger2020whatbertisnot,ethayarajh-2019-contextual,wiedemann2019doesbertmakesense}. We add to these results with detailed illustrations of some of the structures supporting this internal organization of language. The approach developed here can be useful in many additional contexts and in future work (discussed at more length elsewhere in the paper).

We are motivated to look into the representations of the most highly performant computational model of language to date\footnote{By which we mean, LLMs in general, although we're looking at a particular LLM, the RoBERTa MLM.} because it provides a sketch of what might be minimally necessary to support human-like language performance. It also provides insight into a technology which is increasingly ubiquitous, which is important as we seek to understand what sorts of tasks these tools can be used in, and what sort of pitfalls they might afford.\footnote{As research explores correspondences between human cognition and machine learning~\cite{binz2025foundation}, the horizon of possibilities for LLM applications in linguistics, cognitive science, and studies of human behavior continues to be re-evaluated. For this project, however, we focus strictly on human-like linguistic performance, not broader cognitive alignment. LLM linguistic abilities are sufficiently established for our purposes, despite their outputs not being perfectly human-like (if they were, then training on model-generated text would not cause collapse)~\cite{shumailov2024collapse}.}

\begin{figure*}
\centering
\includegraphics[width=\textwidth]{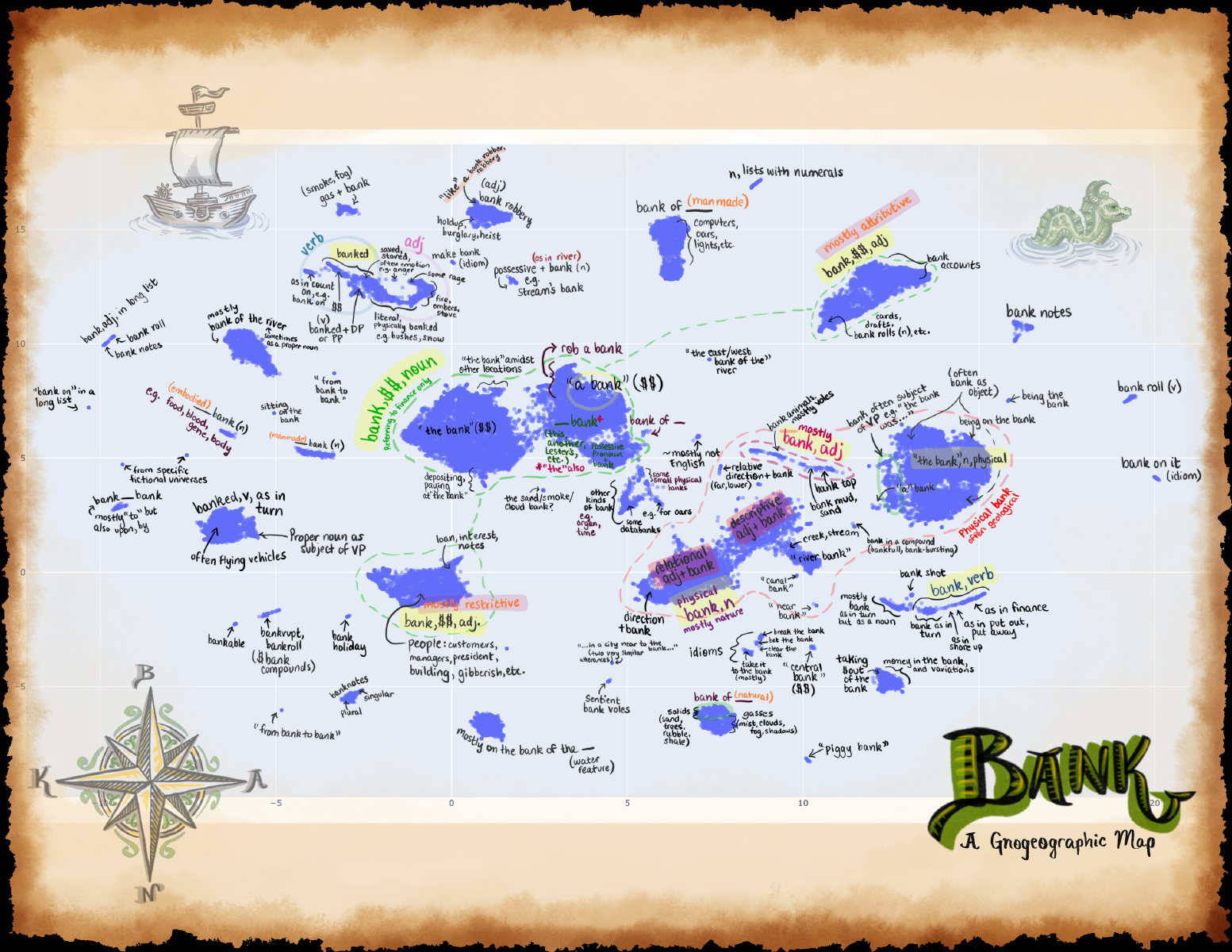}
\includegraphics[width=0.29\textwidth]{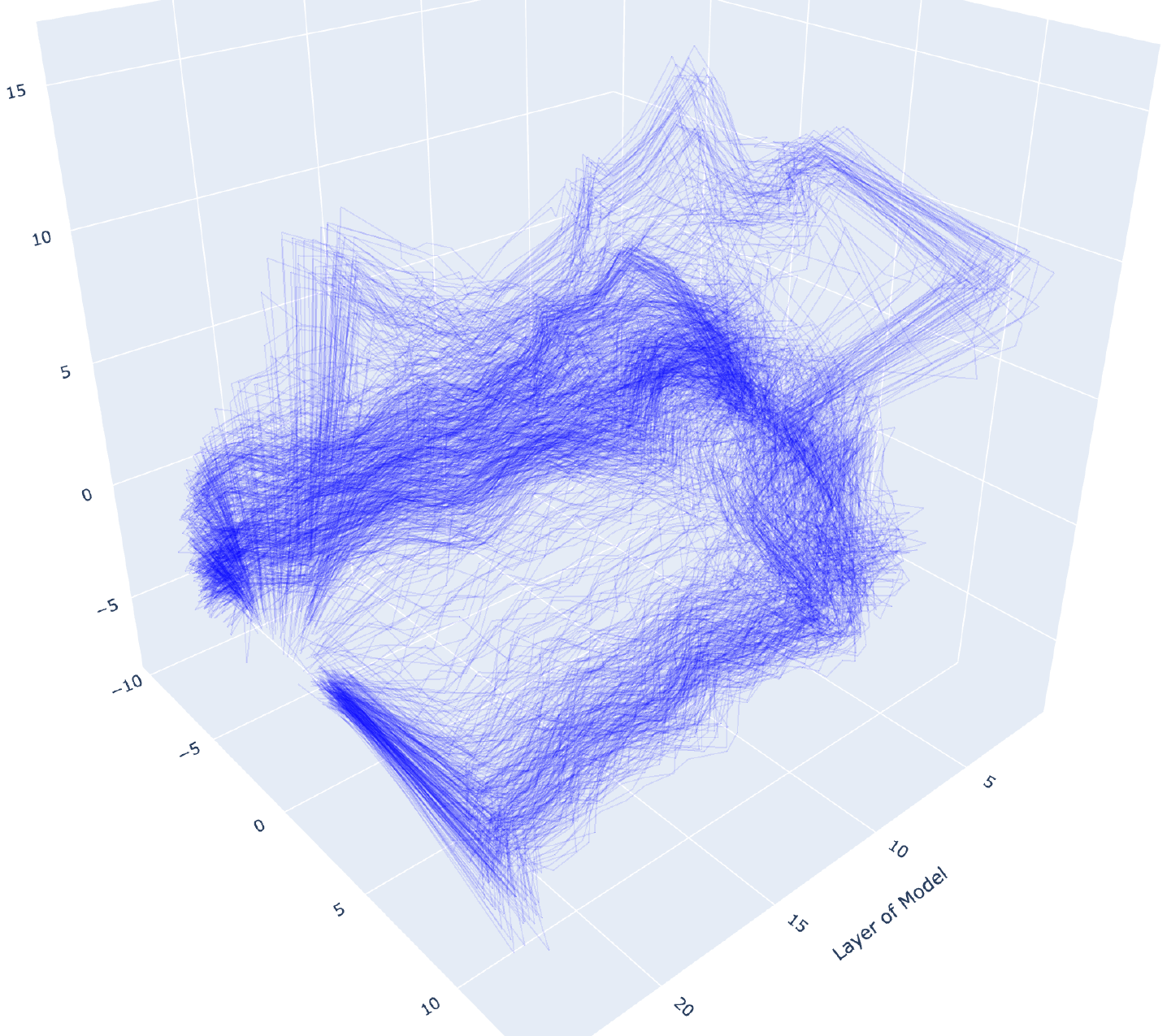}
\includegraphics[width=0.29\textwidth]{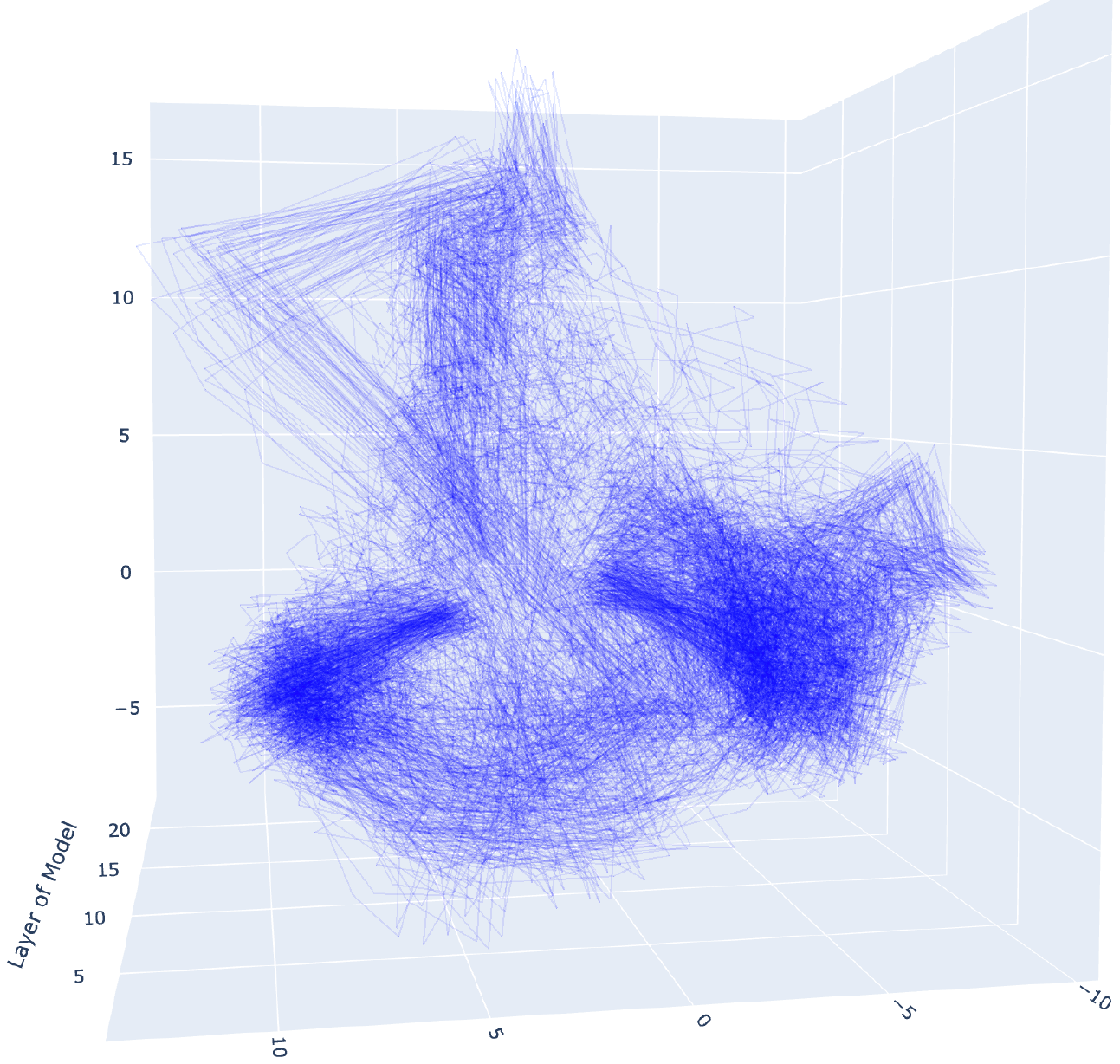}
\includegraphics[width=0.29\textwidth]{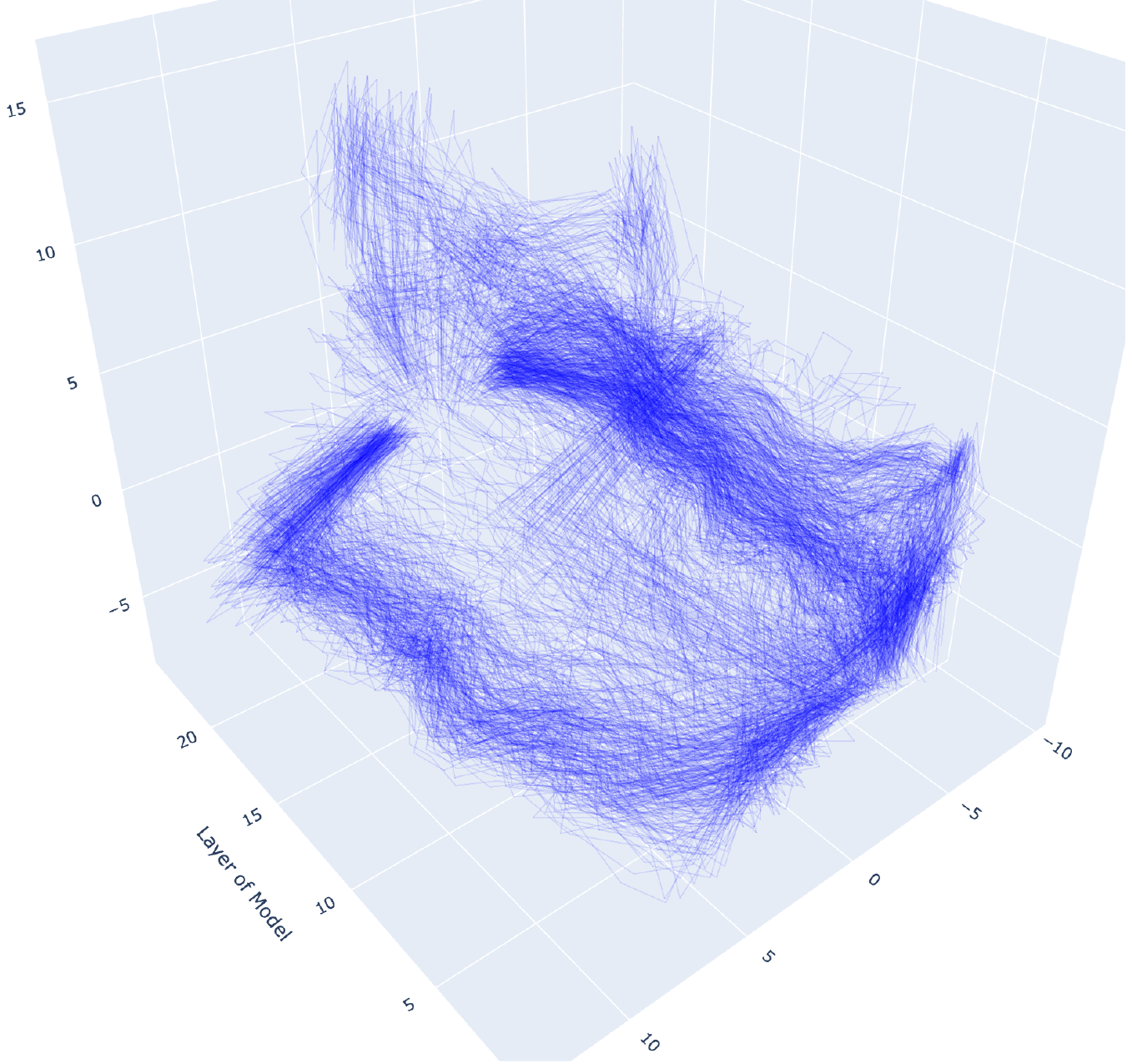}
\caption{A condensed visualization of the content within instances of the `` bank'' token as it traveled through the RoBERTa MLM. Condensed representations of the trajectories themselves, in which the token vectors are 2D points moving from layer to layer within the model, are shown from various angles below the map figure. We condensed the trajectories using PCA, then UMAP, to create the map. We took the position of the `` bank'' token vectors after each layer of the model, then condensed those trajectories with PCA, then used UMAP to cluster the condensed trajectories. The maps show the clusters according to the first two UMAP dimensions. We sampled sentences at points within the clusters and annotated the map with our interpretation of what was in that cluster. The end result reveals that the clusters distinguish the financial sense of `bank' from `databank' or `river bank', and that the top-level meanings (financial, river) can be further subdivided, e.g. the financial meaning is divided into the physical institution of a bank versus the abstract financial entity and associated functionality. More details on the contents of the gnogeographic maps are provided in the main body of the paper, including discussion of embodiment, time, space, Aktionsart, prototypical usage, syntax, formatting, and some very unhuman-like potential `meanings' such as a vehicle sense for `` the''! The compass rose, sea monster, and ship are purely aesthetic.}
\label{fig:bank}
\end{figure*}

We passed contiguous spans of text (a context window of 100 tokens on either side of the token of interest) sourced from a relatively small corpus of English-language fiction through a RoBERTa (large) Masked Language Model (MLM); we parsed the source text into tokens using a BPE tokenizer, the same approach RoBERTa uses. We then extracted the latent space data for our token of interest (in this paper, `` bank'', Figure~\ref{fig:bank}, `` run'', Figure~\ref{fig:run}, and `` the'', Figure~\ref{fig:the}, with leading spaces) by saving the embedding vector output by each layer of the model. We used consolidation strategies---UMAP and PCA---to boil down the large amount of data into something human-readable (through truncating and clustering). UMAP can result in ``finer clustering than is necessarily present in the data,'' which should be kept in mind when interpreting the resultant clusters, but it is better able to preserve aspects of the underlying structure than a pure clustering method, which is particularly useful for our visual exploration~\cite{mcinnes2020umap, umapdocs2024}. For more details of this process, see Sec.~\ref{sec:embeddings_methods}, Sec.~\ref{sec:extispicylimitations}. Our reasons for using the RoBERTa model are discussed in the Appendix (Sec.~\ref{appendix}). These particular gnogeographic maps are unique to RoBERTa and the text we chose to use; other models and other contexts would both yield different maps. What we are exploring is what these internal organizations \textit{can} look like, and what kind of strategies a model \textit{can} use to organize. Although the model's architecture (including its training objective) and how it was trained will each impact the resultant gnogeographic maps, given that LLMs as a group exhibit impressive natural language competency compared to older models and do have significant commonalities between them, we don't see any reason to think these gnogeographic maps would be irrelevant outside of this specific combination of model and corpus.

Because we didn't know exactly how each layer of the model or each dimension in the token vectors might contribute to the model's overall performance (although this is an active area of research~\cite{zhu2025layercaketokenawarecontrastivedecoding,Tenney2019a,rogers2020primer,zhuang-etal-2024-lexicon}), we incorporated the token vectors after each layer.\footnote{Note that this choice doesn't de-emphasize the initial embedding position, since every instance of the token starts there and subsequent locations are built on that starting point.} However, this made the representation for each token instance quite large, further reducing the already low interpretability. To counterbalance that, we used PCA to reduce the dimensions before using UMAP to cluster. The goal being to reduce the difficult problem of finding clusters that make sense across a huge number of facets simultaneously to the easier problem of finding clusters that make sense across a vastly-reduced set of the most prominent features. As seen in Figure~\ref{fig:bank}, this approach resulted in incredibly interpretable clusters that paint a vivid picture of the sort of information that can be encoded in token vectors. This approach can be generally useful (sort of `boosting the signal'), but also specifically handy for using the embedding spaces of encoders to obtain fine-grained clusters of word senses which can then be used for classifiers for downstream NLP tasks (embed-then-classify, a demonstrably practical strategy~\cite{CoilChenBruckner2025}, similar to suggestions in \citet{xypolopoulos-etal-2021-unsupervised}).

Note that once a token is moving through the model, because of how the model works (notably the attention portions), it no longer represents the meaning of that token alone: it comes to contain significantly more information. In particular, information about the immediate context of the token, captured by the model through e.g., attention, is aggregated, with the token vector as receptacle. The structure in the gnogeographic map should reflect the kind of information the model can use a token vector to encode. Importantly, although the token (vector) is the constant receptacle, it is still unclear what happens where, through what means, within the model.\footnote{Revelations are coming via an emerging body of work, for example, Zhuang \textit{et al.}, 2024 state that ``early-layer representations in the LM... are closer to containing only lexicon-level information''~\cite{zhuang-etal-2024-lexicon}. See also \citet{zhu2025layercaketokenawarecontrastivedecoding,Tenney2019a,rogers2020primer}. Due to both the uncharted nature of this technology and because we are interested in a holistic impression of the model's ``mind'', in this project, we took a broad view in what information should be included in our maps: we included the token vectors throughout the entire inference process. In other words: there are 25 distinct states of the embedding matrix---the initial embedding, plus the output after each of the 24 neural net/ attention block layers in RoBERTa's architecture---during inference, and we used all of them in creating our gnogeographic maps. We also used all suitable instances of the token available in our corpus. Our goal was to capture as much as we could of not just the decision that would correspond to output, but the information and cognition that went into that decision.} Finally, note that although each gnogeographic map corresponds to one token type, one symbol, the map is an agglomeration of thousands of token vectors (attached to that symbol in different instances, as it occurred in different spans of text), which the model sought to place meaningfully within its latent space.

Recall that any information present in the tokens arrived through the system of (mostly) the model and the training data, meaning it was almost certainly carried by the linguistic signal. The model mediates a significant transformation of the sequences of symbols, the input text, into a new, condensed form, \`a la bowtie cognition~\cite{levin2024stigmergy}.\footnote{As a corollary, confabulation or hallucination is fundamentally necessary for bowtie-like cognition~\cite{zimmerman2024blind, levin2024stigmergy}!} This means we can see what sort of information can be conveyed through those specific mechanisms of language an LLM is sensitive to (the DH: 1D relations of context between symbols), something extremely difficult to disentangle in people~\cite{youn2015universal, zimmerman2024blind}.\footnote{This approach also gives us a way to look inside words of interest, including those historically difficult to get at from a meaning perspective, such as function words like ``the'', punctuation, numbers, etc. Varying training data by time period, geographical region, etc., and then looking inside words could be valuable for studying language variation and change as well.}

Overall, we found that clusters do reflect syntactic and semantic information, consistent with the DH's syntagmatic information. Clusters grouped utterances by the sense of the token and/or by the token's role in a syntactic constituent. Not every cluster was easily interpretable, and where clusters were interpretable, they were often imperfect (i.e., although we saw a pattern, there were some utterances that did not obviously fit that pattern). While our clusters are interpreted qualitatively and subjectively, the initial results are intriguing.

Polysemous words were immediately visually distinguishable from more monosemous words.\footnote{Highly polysemous ``set'' and ``run'' have many small, well-separated blobs floating around along with a few larger, more meandering clusters that represent the most common meanings. The majority of the clusters are very small and compact. There are shape structures that seem to be in conversation with each other, mainly visible within or from the larger clusters (Figure~\ref{fig:run}). On the other hand, ``teacher'', a word often considered fairly monosemous, has what look like three very large main clusters, two of which look to be connected in content by thin lines of utterances. The shape of all three could be related to each other. The few small clusters outside of the main three generally look like they may be in conversation with the main ones (in terms of their shapes and direction). See Figure~\ref{fig:teacher}.} We can clearly see ``naturally interpretable clusters of closely related concepts'' within polysemous words, as has been described among human language in lexical networks~\cite{youn2015universal}.

Although the syntactic information often related to words which were sequentially very close, even adjacent, to the token within the utterance, this was not always the case.\footnote{All of the information we could perceive is with respect to words pretty near the exemplar token, though, since our judgments were based on a 21 token context window (see Sec.~\ref{sec:embeddings_methods}).} For example, the two segments of ``the + [comparative adjective/adverb]'' within the correlative comparative constructions\footnote{e.g., think of phrases like, ``the more, the merrier''} in \textit{the} were not necessarily very close together (see Figure~\ref{fig:the}).

There looked to be both inter- and intra- cluster structure, with identifiable sections of specificity (as to both word meaning and syntax) within a cluster with a broader theme, and across or between clusters. There can be fine-grained distinctions such as between relational, descriptive, restrictive, and attributive roles, as well as distinctions in construction, and possibly even between idiomatic vs. productive uses (Figure~\ref{fig:bank}).

We noticed that the organization of `the' seemed more syntax-based than `bank' and `run', which would make sense as it is a function word rather than one traditionally considered to have much semantic content; however, we did not think it had \textit{no} semantic content. For example, `the' has a vehicle-based region. In `the' we also saw a cluster that looked entirely formatting-based, around dialogue, presumably because the strong conventions around format for dialogue were the strongest signal for the LLM in that kind of text, overwhelming what would conventionally be considered meaning. These could be viewed as examples of flattening: information we would not typically consider syntactic or semantic is being bundled up alongside more conventional meaning, potentially indistinguishably --- all of the information is stored in the same structure, and impacts the position of the token vectors in the latent space.\footnote{There could be distinguishable differences in the details; that is an open question.} The vehicle-based region illustrates how frequency and context (from the sequences of tokens we passed through the MLM, as well as from the training data) are among the broad range of information falling outside conventional word meaning that can directly impact token content for the LLM.\footnote{Not to say context is not part of meaning for us, the DH is clearly contrary to that, but the way context is important to the model may not be the same as the way context is important to us, and, at least arguably, we don't usually think -- in a colloquial, vernacular sense -- of e.g. the name `David' as having multiple meanings largely defined by recent context so much as having an abstract role as a name that involves reference. For the model, the content helping to resolve the immediate context is expressed through the same vector as the learned, stable embedding, an intriguing collapse of the particular and the general.} The model saw many instances of `the' paired with vehicle-related content words, which influenced the position of those utterances in the latent space. Speculatively, this could be essentially the construction of a new, semantic aspect of `the' in the model's cognition. This evidence of the impact of frequency (and formatting) has downstream implications for model performance, training, architecture, and inquiries into model bias. It is important to recognize that the input signal, though familiar to us as text, is not necessarily interpreted the same way by the model. (Recall that the gnogeographic maps are not direct representations of the embedding matrix or the latent space, but are related.)

We can also see organization that looks squarely semantic. For example, in \textit{bank}, we see a distinction between manmade (e.g., computers) and natural (e.g., clouds) objects in two clusters for ``bank of [NP]'', and within the natural cluster we see solids grouped together (e.g., rubble) and gasses (e.g., mist) grouped together. See Figure~\ref{fig:bank}. One of the verb clusters in \textit{run} looks to have the organizing sense of ``flow''. Such examples are tantalizing hints of an underlying world model projected through the vehicle of individual tokens (another is a cluster where fire and rage were near each other, suggestive of literal and metaphorical heat).

Some of the clusters seem to reflect putatively fundamental human-like distinctions such as prototypical usage, embodiment, and concreteness (relevant to the Swadesh list and processes of metaphorical extension~\cite{maudslay-etal-2024-chainnet,swadesh1952}).\footnote{On the left in Figure~\ref{fig:bank}, a small column of clusters suggests organization around embodiment--biological components and action defined relative to a body (``[food, blood, gene] bank'' and ``banked'' as in turn, respectively). Around the lower center, there seems to be a larger area involving significant figurative components, e.g., ``piggybank'' and idiomatic uses.} We also saw a cluster in ``run'' focused on its temporal ending (as in stop, e.g., ``run into a wall''; Aktionsart), which is notable since time and space are core dimensions to the human experience and universally made salient by the technology of language~\cite{rijkhoff2002verbs}. Speculatively, this could be caused by a kind of indirect grounding in human experience (perhaps not unlike what can happen for people when they receive diegetic approximations for what they are not directly experiencing~\cite{Liu2025Color}).\footnote{The potential `ceiling' for what can be conveyed in language depends (at least) on both the properties of language and the architecture of the speakers. This is being borne out in emerging research with deep learning models, exploring under what circumstances their representations diverge or converge with human representations (as can be captured by limited metrics)~\cite{Xu2025LLMConcepts,zhuang-etal-2024-lexicon}.} That LLMs could be sensitive to concepts of time, space, and even bodies, is particularly notable since the embodiment\footnote{Bodies also being the primary mediator through which we perceive time and space.} of an LLM is strikingly inhuman. These clusters are also suggestive of a porous boundary between world knowledge and linguistic knowledge, since they reflect a human world model and human experience through something like word meanings, which could be both promising for LLM gnogeography and informative as to how language is implemented for people (as well as hinting at potential alternatives).

This method could provide a quantitative way to identify prototypical usages (e.g., probably one of the big clusters if the word is polysemous), to assess the degree of polysemy, and to count and classify word senses. For example, you could train a classifier on the utterances in each desired cluster, and then use that classifier to get fine-grained semantic and syntactic parsing within a novel text.

There is plenty of additional structure to be found within these exemplar token embeddings, but for the purposes of this paper we are looking at the highest level for a rough sketch of the sort of things an LLM can `know' about a token, how much it knows about a given token, and whether that aligns at all with an intuitive human interpretation (which it clearly seems to), as well as for what sort of information an LLM uses token vectors to organize and encode. The general shape of our limitations suggests that, if anything, models can and maybe are sensitive to more than what we can see here. Speculatively, we think these results point away from the idea that the LLM implementation of language should be dismissed as `mere' pattern-matching~\cite{10.1145/3442188.3445922}. For example, there is structure subdividing ``bank of [NP]'' constructions which are arguably identical at the level of plausible token generation, although exactly how to interpret that is unclear. (The attention mechanism guarantees that context impacts token content.)

We do think it is impressive that syntagmatic information -- distributional information, these fairly simple patterns -- can reflect human-interpretable, fine-grained distinctions with respect to word meaning and usage under any conditions (although of course specific architecture is necessary to enable that performance).\footnote{As seen in the gnogeographic maps, which illustrate the power of context in shaping these fine-grained distinctions.} The content in the tokens is determined by the data provided to the model, but the information had to get there from the initial signal, and that is no small task.\footnote{In future work, we would like to further explore the relationship between gnogeography, tokenization, and model performance (including collapse) by looking at the gnogeographic maps that result when the model is trained only on model-generated data (at successively greater remove 
from initial exposure to human-generated data). We would also like to better understand the ramifications of how strings are tokenized in different contexts. Models like SpanBERT could be helpful for this task~\cite{joshi-etal-2020-spanbert}. What are the consequences of the fact that 
the same span of text could be constructed with multiple subsets (with different attendant probabilities) from within the same vocabulary? Which senses are captured by the most common tokenizations, e.g., which contexts fall under `` bank'' as opposed to `` ba'' + ``nk'', etc.? Intuitively, we think Chomsky's observation that concision is inherently conservative~\cite{Chomsky_Concision} and \textit{lectio difficilior potior} will be relevant: unusual surroundings will lead to unusual tokenizations. Similarly, unusual words 
that are OOV degrade the most salient distribution available to the model. Under what conditions will the model output a word not in its vocabulary (composed of multiple tokens, and not explained by standard morphological rules)? When an LLM generates such a word by concatenating multiple tokens, it seems like that process could be in between what we think of as coining a neologism and recalling a rare word; is that distribution still Zipfy, or have such words perhaps become less likely?} 

\section{Discussion}
\label{sec:discussion}

% Please note that this paper has elements of both philosophical, exploratory work, and some quantitative evidence. Our aim is for that evidence to provide fertile ground for thought, allowing the lead author to build up a productive mental model of LLMs with explanatory power which can be further cultivated and experimentally tested in future work. The quantitative evidence is not intended to sufficiently support the claims in the Discussion, Conclusion, and Post-Script. There are three prongs in the model emerging here: (1) Tokens are the ``eye'' of LLMs. Tokenization shapes the distributions available to the model, which means they form a fundamental sensory and epistemic layer. (2) Human linguistic structure emerges in tokens because tokenization and language share efficiency pressures. Therefore, existing tokenization algorithms can be useful tools in language change and evolution. On the other hand, natural language is human-embodied, suited to human-scale information locality (processing resources, noisy environments, priorities, etc.) and there may be opportunities to shape tokenization (and related features) more specifically to LLM needs. (3) Relatedly, zooming out from that, tokens act within a language-centric architecture with a particular scale, with implications for cognition and grounding. This offers opportunity to study language, but presumably is not always the best choice for LLM performance.

\subsection{Meaning and grounding}
\label{sec:meaningandgrounding}

We speculate that syntagmatic information lays the groundwork for successively richer semantic information, up to some limit: Syntagmatic information is the guiding force for the emergence of semantic themes, and anything propositional or descriptive can be conveyed purely linguistically (by the DH). We think this fits with our discussion in Section~\ref{sec:thedistributionalhypothesis}. For example, we see a cluster of largely prepositional phrases (PP), which suggests that syntax is their highest level of organization. However, a spatial relationship (over/under) subdivides the individual prepositional phases, hinting at the potential for more and more fine-grained detail.\footnote{Maybe with exposure to even more instances (than were in the training data), further semantic structure would emerge for the model. Presumably, with more instances in our map (than were in our corpus), we would be able to see more of the structure already within the model's gnogeography. It also stands to reason that there is more fine-grained detail in the token vectors that we are not seeing due to our truncating and processing choices (see Sections~\ref{sec:embeddings_methods},~\ref{sec:extispicylimitations}). It would be interesting to see whether such details increased in prominence with exposure to more instances. By varying the PCA dimension cut-off and the number of instances systematically, we could estimate the relative prominence of each feature within the linguistic signal, and ascertain how much stimulus would be needed by the LLM to learn each feature.}

It has been argued that the property of displacement (time and space travel~\cite{zimmerman2024blind}: power over fundamental dimensions~\cite{dodds2023ousiometricstelegnomicsessencemeaning,rijkhoff2002verbs}) is core to the development and acquisition of language for people~\cite{perniss2014bridge}; what does that mean for LLMs, which presumably experience time and space significantly differently? For LLMs, all tokens are structured and treated alike, and all symbolic content is presumably ideational and descriptive.\footnote{What is reference, for an LLM? For us one of its non-trivial functions seem to be as a pointer across modes.} Yet from relations, through language, a concept of space emerges. This observation returns us to the symbol grounding problem as to what extent and in what domain, if any, LLMs can be grounded: fundamentally, how can LLMs connect linguistic knowledge to world knowledge?\footnote{The class of iconic words might be helpful in investigating questions of symbol grounding with LLMs. Iconic words, including onomatopoeia, incorporate a less arbitrary connection between their physical, supradiegetic information and their meaning. Could the integration of sensory information into the word's meaning for us provide indirect sensory grounding to the LLM?  We are curious about the connection between iconic words (which are often embodied and perhaps have some relationship with the Swadesh list~\cite{swadesh1952}), language development, and embodied cognition. Very tentatively, high iconicity words may be less likely to end up as tokens (based on the model vocabularies we explored).}

Some solutions to unwanted behaviour like inappropriate hallucination invoke grounding as a potential solution, and propose that LLMs may need choice or agency to avoid such hallucination.\footnote{As aforementioned, hallucination of some kind is necessary. But the details of how it manifests 
 determine whether it is desirable.} In our view, grounding strategies invoke another mode to enhance the salient meaning available to the model, potentially by forcing more abstraction or compression to find a representation common across modes. Even a strategy like Chain-of-Thought or Tree-of-Thought prompting~\cite{yao2023tree}, which increases the resources devoted to each question through increasing the number of generated tokens, could be viewed as introspective grounding; Sapir-Whorfianly, forcing the LLM to speak more about a subject increases its understanding~\cite{sepsapirwhorf}.\footnote{Empathy could be viewed as grounding in someone else; language can be a stigmergic tool for self, at the individual or group level~\cite{levin2024stigmergy,welker2024selfviews}.}

However, improving LLM conformity with human experience is only one possible direction. Human architecture guided and, to some degree, co-evolved with human language. LLMs, with their significantly different architecture, bring up opportunities to re-examine how language can be shaped; for example, LLMs never have to worry about making out specific speech amongst background noise in a crowded room, so they don't require the level of redundancy built into human language~\cite{xia2025tokenskipcontrollablechainofthoughtcompression}. For additional discussion on this topic, see \cite{zimmerman2025locality}.

We think learning via the DH can be sufficient to establish the inferential portion of Marconi's lexical competence~\cite{marconi2003lexical},\footnote{Machery, 2006, ties inferential competence to linguistic competence and referential to non-linguistic (in human brains); we think these map clearly to linguistic and non-linguistic meaning according to the DH, and therefore to LLMs~\cite{brainsblog_lexicalcompetence,marconi2003lexical,sahlgren2008distributional}.} while grounding provides depth to meaning by adding specific referential competence.\footnote{The question remains of how far diegetic approximations alone can extend the gnogeography but implies an only-the-thing-is-the-thing-ness across modes~\cite{zimmerman2024blind}.} Given that strategies like contrastive linguistic-visual objectives (as in some language-vision multimodal transformers) can end up creating contextually superior language representations~\cite{wolfe-caliskan-2022-contrastive, zhuang-etal-2024-lexicon}, this suggests that we could seek model improvement through architecture that combines multiple objectives: specifically, the model could augment its inferential competence (learned through the DH from text, with an objective like plausible token generation) with limited referential competence acquired by the pairing of sensory 
and linguistic information (sensory grounding, through contrastive objectives that relate other modes to the purely linguistic realm), as Zhuang \textit{et al.}, 2024 do with visual grounding (although they don't describe what is happening in terms of inferential and referential lexical competence).\footnote{Specifically, Zhuang \textit{et al.}, 2024 found ``that the LexiContrastive Grounding models... relate the concrete words in a more human-like way than the abstract words''. We think this could make sense as the visual reference may be simpler and more consistent for concrete than abstract words.}\textsuperscript{,}~\footnote{Zhuang \textit{et al.}, 2024 point out that the typical implementation of visual grounding for LLMs is partial, and notably static (similar to our prior observations about text~\cite{zimmerman2024blind}). For people, visual experiences directly incorporate movement and time. They also point out that the way their model is set up doesn't allow for any potential benefit to syntax acquisition (the grounding is restricted to the lexicon-level). In other words, their model is probably near a lower bound of visual grounding for language  acquisition, and there is presumably much more to be gained in that realm.}

Meaning requires both linguistic and world knowledge. Meaning, as we experience it, requires that there be no absolute partition between our implementation of language as a technology and our world models.\footnote{A variety of linguistic aspects can be encoded either morphosyntactically or lexically~\cite{rijkhoff2002verbs}. Within what can be conveyed distributionally are apparently pieces of syntax and semantics, sufficient for the language production of LLMs. Note that for people, the lexical component can be significant. For example, there is evidence that syntax is not the only mechanism for compositionality and inter-relation~\cite{mahowald2023grammatical}.\footnote{For example, Mahowald \textit{et al.}, 2023, conclude that ``grammatical cues such as word order are necessary to convey subjecthood and objecthood in a minority of naturally occurring transitive clauses; nevertheless, they can (a) provide an important source of redundancy and (b) are crucial for conveying intended meaning that cannot be inferred from the words alone, including descriptions of human interactions, where roles are often reversible..., and expressing non-prototypical meanings''~\cite{mahowald2023grammatical}.} Future (potentially cross-linguistic) work could investigate whether LLMs learn concepts better when more of their meaning is theorized to be encoded morphosyntactically. Similarly, future work could compare function and content words.} We think this is why no one can agree on its boundaries, and why people who argue for stochastic parrots~\cite{10.1145/3442188.3445922} or against the DH feel unsatisfied by the kinds of MVP, less-than-complete meaning these models currently possess.\footnote{We wonder if there is a connection between the DH and small talk. During small talk, it seems like we can perform language competently without fully engaging all that meaning can be. Similarly, with semantic bleaching, sometimes a word like awful or terrible fully captures one's attention, fully invokes a weighty emotional response. However, for most conversations, the level of engagement is more shallow. Maybe the DH, even the inherent degrounding of the fungibility that allows language to work in the 
 first place, is constantly wearing down the meanings of words, and we use other strategies to compensate, to rebuild. Maybe that's one of the functions of poetry.}

Creative applications of grounding, including through changing the model's objective function, seem like a promising approach for enhancing the depth of meaning accessible to LLMs, which can in turn improve their competence in a variety of contexts.\footnote{Technically, depending on the  implementation of grounding, this can make them no longer LLMs, as the typical implication of LLM is unimodal to text (so e.g., any sensory grounding would take the LM out of the LLM realm). However, a strict plausible-token-generation LLM may theoretically be upper-bounded by cognition essentially equivalent to producing MVP language with limited semantic meaning, world model, and other capabilities within its gnogeography.}\textsuperscript{,}~\footnote{Much prior research on grounding in LMs has been ``primarily tailored for tasks requiring simultaneous reasoning across both modalities''~\cite{zhuang-etal-2024-lexicon}, but we are proposing that the benefits of deeper meaning within the model could go far beyond that. For example, as in our Flatland analogy~\cite{zimmerman2024blind}, simply knowing that diegetic boundaries exist could significantly re-shape the gnogeography.}

\subsection{Objective functions: POSIWID teleology}
\label{sec:objectivefunctions}

As one of the main forces shaping their cognition,\footnote{For an example of the deep-rooted impact of objective function on cognition, Wolfe and Caliskan, 2022, found that ``CLIP embeddings show that the high anisotropy observed by Ethayarajh (2019) is not the inevitable result of contextualization, nor even of a specific language modeling architecture, but is connected to the pretraining objective of the model''~\cite{wolfe-caliskan-2022-contrastive}. (They mean pretraining as in before fine-tuning, so they still mean the objective function of the main intelligence.) The importance of the objective function to model cognition is also evident in results from transfer learning.} exploring the space of objective functions is promising for allowing LLMs to get more out of the same data\footnote{As in Zhuang \textit{et al.}, 2024, who propose that greater model efficiency could be achieved through more ``cognitively aligned language learning technologies'', and found improvements in certain language representations when they modified the objective functions to include visual grounding~\cite{zhuang-etal-2024-lexicon}.} and to solve some of the shortcomings that seem to come from such a stringent emphasis on producing plausible language above all else, a strategy that makes current LLMs silvertongued but ultimately unreliable and unhumanlike in terms of any behaviour outside those bounds (e.g., this could include aspects of Grice's maxims, moral consistency, reciprocal self-disclosure, etc.)~\cite{g_wayofwords,nunes2024moral}.\footnote{The section header is in reference to Stafford Beer's famous coinage, POSIWID, or, ``the purpose of a system is what it does''.}

Fundamentally, seeking to modulate the primacy of the next token generation objective function is the motivation behind alignment approaches like RLHF (reinforcement learning from human feedback)~\cite{stiennon2022learningsummarizehumanfeedback}. Due to the clear efficacy of next token generation (and similar) for creating human-like language performance, though, this function usually remains prominent in the mix.

Perhaps unintuitively, the objective functions of components and precursors, processes arguably not directly invoked by the main intelligence itself, can also impact its cognition.

There is an ongoing body of work on the impact of choice of components that determine model parameters during training~\cite{pagliardini2024ademamixoptimizerbetterfaster, stiennon2022learningsummarizehumanfeedback}. For example, Caples and Renaud, 2024, found that optimizer choice impacts the ``internal representations formed by these models''~\cite{caples2024adam}. The pathway, in this case, is arguably further removed than in the case of the objective function: although the optimizer is directly involved in shaping the mind of the model (it helps set the weights in the neural net), it is not something the model will draw on directly once it is finished with training.

We show that the tokenization objective function can impact the internal representations of LLMs, not only in choice of tokens (which is self-evident) but in the content of the tokens as well. By forcing orthographically related terms (e.g. `run'/`rung', Figure~\ref{fig:run}) to share a significant portion of their initial embedding information, token-based co-lexification induces representational challenges that the model must handle, meaning that tokenization's objective function -- here, BPE's orthographic-efficiency-maximizing objective -- can shape the model's internal representations. Of note, the pathway of influence here is arguably at greater remove from the core intelligence than the aforementioned optimizer or the model's objective function.

In Figure~\ref{fig:run}, there are clusters for `rung' and `runnels' encoded via the token ` run'. The semantic information from instances using the words `runnel' and `rung' are only co-tokenified with \textit{run} in the model because of the tokenization objective function of orthographic efficiency agnostic of meaning. That `run' is a component of the strings `runnel' and `rung' is more closely related to their respective orthographies than to their meanings, whether syntagmatic, syntactic, or semantic. The model endeavors to place token vectors meaningfully with respect to other token vectors within its latent space, so the inclusion of a new sense or meaning impacts the placement of other vectors. 

Thus the choice of receptacle, of atomic unit, for linguistic representation matters for eventual semantic representations: an emergent connection between Harris' ``purely linguistic'' meaning and the koinos kosmos (human world model). (Recall Section~\ref{sec:meaningandgrounding}.)\footnote{The word `runnel' could be tokenized any number of ways when encountered in input. For example, run+[n+el,ne+l,n+e+l,nel] or r+un, etc. As aforementioned, the tokens available in the vocabulary can obscure the salience of the necessary distributions (Figure~\ref{fig:chartowordlike}, Figure~\ref{fig:vocabtimelines}), but the number of possible ways an utterance could be tokenized when the model encounters it also contributes to this obfuscation (tokenization is deterministic but context-dependent). Both aspects of tokenization, the choice of vocabulary pieces and the chunking of input and output, are implicated. The more possible tokenizations given the vocabulary, the more patterns have to be reconciled by the model. Considered from an MDL perspective, this is suspect.}

We ask tokens to be combinatoric primitives for both meaning and orthography, which are necessarily non-trivially independent of each other~\cite{zimmerman2024blind}. The objective function of the tokenization algorithm is not perfectly aligned with the objective function of the model, and the primitives that would best support their respective purposes have different structures.

\subsection{Flattening}

The tokens form the semantic primitives for the model. They likely need to be well-formed receptacles for the kind of semantic information that makes up human world models, as that is a primary role of words for us.\footnote{``A primary function of the lexicon is to encode novel concepts and ideas as they emerge over time''~\cite{xu2024wordreuse}.} The Sapir-Whorf hypothesis in general, nameability (Lupyan's Label-feedback hypothesis), and even theories from language acquisition like fast-mapping offer explanations for what kind of processes might need such structures~\cite{sepsapirwhorf,lupyan2012linguistically,gelman_fast-mapping_2010}.\footnote{Although LLMs have something like MVP language, how that maps to the intersubjective or intrasubjective competence humans derive from language is unknown.}

Recall Figures~\ref{fig:chartowordlike},~\ref{fig:vocabtimelines}, which basically show the range of tokens that can be produced as following exponential decay: at the one extreme, there can be as many tokens as characters in the utterance being tokenized, each token being as short as possible; at the other extreme, an utterance can be a single token. As the vocabulary size changes, the tokens go from representing individual characters, to words, to short phrases, to memorizing entire utterances within individual tokens. This progression illustrates how tokenization radically alters the fundamental units available to the LLM, units which are the intermediary between it and the world. These units play dual important roles, as they both impose an organizational structure on information in the LLM's internal representations and are the medium for the distributional patterns of language that allow the model to learn via the DH. As with the extreme example of book-length tokens introduced earlier, it is clear that both of those roles impact model cognition.\footnote{Although we arrived at some similar conclusions independently, this blog post~\cite{cybernetist} has some good examples showing how cosine similarity (which is basically a semantic measure) is impacted by tokenization.} Although mostly out of scope for this paper, it is worth noting that the dimensions of the token vectors matter as well. We can apply similar reasoning.\footnote{For example, imagine a token vector of length 1, and a language with symbols (words, tokens) A, B, C, and D. ``A'' means something like [hot and left]; ``B'' means something like [hot and right], ``C'' means something like [cold and left], and ``D'' means something like [cold and right]. A and C are similar in orientation, but opposite in temperature. B and C are opposite in temperature and orientation. A and B are similar in temperature but opposite in orientation. A and D are opposite in both. If you place A between B and C, for example, then where ought D to be placed? Trying to convey this series of meanings along a single line eliminates a significant amount of information!} The ratio between the number of vocabulary items (assuming they mostly do have some difference in meaning from each other) and the length of the token vectors is non-trivially determinative of expressiveness.

Furthermore, it is important to keep in mind how the structure of tokens and the tokenization algorithm mapping input text to vocabulary tokens might constrain model cognition, especially in conjunction with the attention mechanism and the plausible text generation objective functions. For example, consider that 9,342 could be parsed several different ways which are very unlikely to be literally equivalent. Maybe it gets parsed as 9 + , + 342, maybe as 9 + , + 3 + 42, etc. The parsing could be different in different contexts (based on surrounding text). The resultant token vectors are likely to involve different values (as might the Q, K, and V vectors)~\cite{rogers2020primer}.\footnote{Hypothetically, imagine trying to do math when you cannot see that those tokenized sequences involve the same numbers, or trying to learn stable mathematical representations given that obfuscation.}

That the model must repeatedly hearken back to learned token content is plausibly relevant not only to social harm, which we discuss in Section~\ref{sec:biasandalignment}, but also to performance on various tasks~\cite{huang2023lexinvariantlanguagemodels}. By de-tethering the connection to a remembered semantic space (the stable embedding matrix), the lexinvariant LM can presumably adapt immediately to a novel use for a symbol as long as enough information about it is provided in the context. This could be advantageous when the tokens are not playing the roles they typically play in a linguistic context, which would have been the majority of how they were observed during training. Depending on the task, different token content may be ideal, and greater ability to conditionally ignore information -- both within a token and across tokens, within a context -- could be helpful. For example, in cases of polysemy or homonymy. One emerging approach to ameliorate the most prominent aspect of flattening discussed in most CS and NLP literature, the ``meaning conflation deficiency'', is to incorporate multiple vectors for word senses~\cite{sun2023word_sense_embeddings,camachocollados2018wordsenseembeddingssurvey}, but there is typically still a pre-determined inflexibility to that approach (the senses are training-data-dependent, and sensitive to the method of counting)~\cite{wiedemann2019doesbertmakesense}. There are other approaches aimed at improving some of the shortcomings of typical tokenization approaches; for example, distilled contextual embeddings can also be used to boost model performance for specific lexical tasks~\cite{rogers2020primer}. However, the fundamental root causes of these shortcomings persist.\footnote{In more detail: for some kinds of problems, associations previously learned with symbols can be counterproductive. Always hearkening back to a single initial embedding can be detrimental. For example, in an arithmetic context (`a + b'), learned information about the English determiner `a' is entirely irrelevant. For typical LLMs, the vector associated with `a' would start in the same location for both the arithmetic and natural language use-case. That location would reflect what the model learned from its general, accumulated exposure to the `a' token during training, which would probably mostly be about the article, as well as perhaps about the productive affix, and to some much lesser extent, a variety of exceptional cases (wherever `a' occurred in a context where surrounding characters were tokenized with the surrounding pieces rather than with the `a'). The model then needs to overcome what it has learned about the typical occurrence of `a' in favor of the context-specific meaning. In problems like this, where the symbol `a' has something like an unusual, novel meaning that departs from previous experience, starting off with basically no preconceived notions (like the lexinvariant model) is actually an advantage. The ideal scenario would be a model which could flexibly choose whether or not to depend on its preconceived associations for that symbol (which feels akin to what a person would do)~\cite{huang2023lexinvariantlanguagemodels}, an enhanced meta-cognitive ability when it comes to symbolic representation, the cornerstone of LLM functionality. Allowing the model access to extradiegetic attributes of its own representations empowers problem-solving~\cite{zimmerman2025locality}. More generally, delegating an aspect of cognition to the model for open-ended problem-solving as opposed to hard-coding it via the architecture can be beneficial for model performance, a recurrent trend in machine learning~\cite{zimmerman2025locality}.} Thus there are many ways `flattening' could be working against the model, with respect to tokens.

\begin{figure}
\centering
\includegraphics[width=0.75\columnwidth]{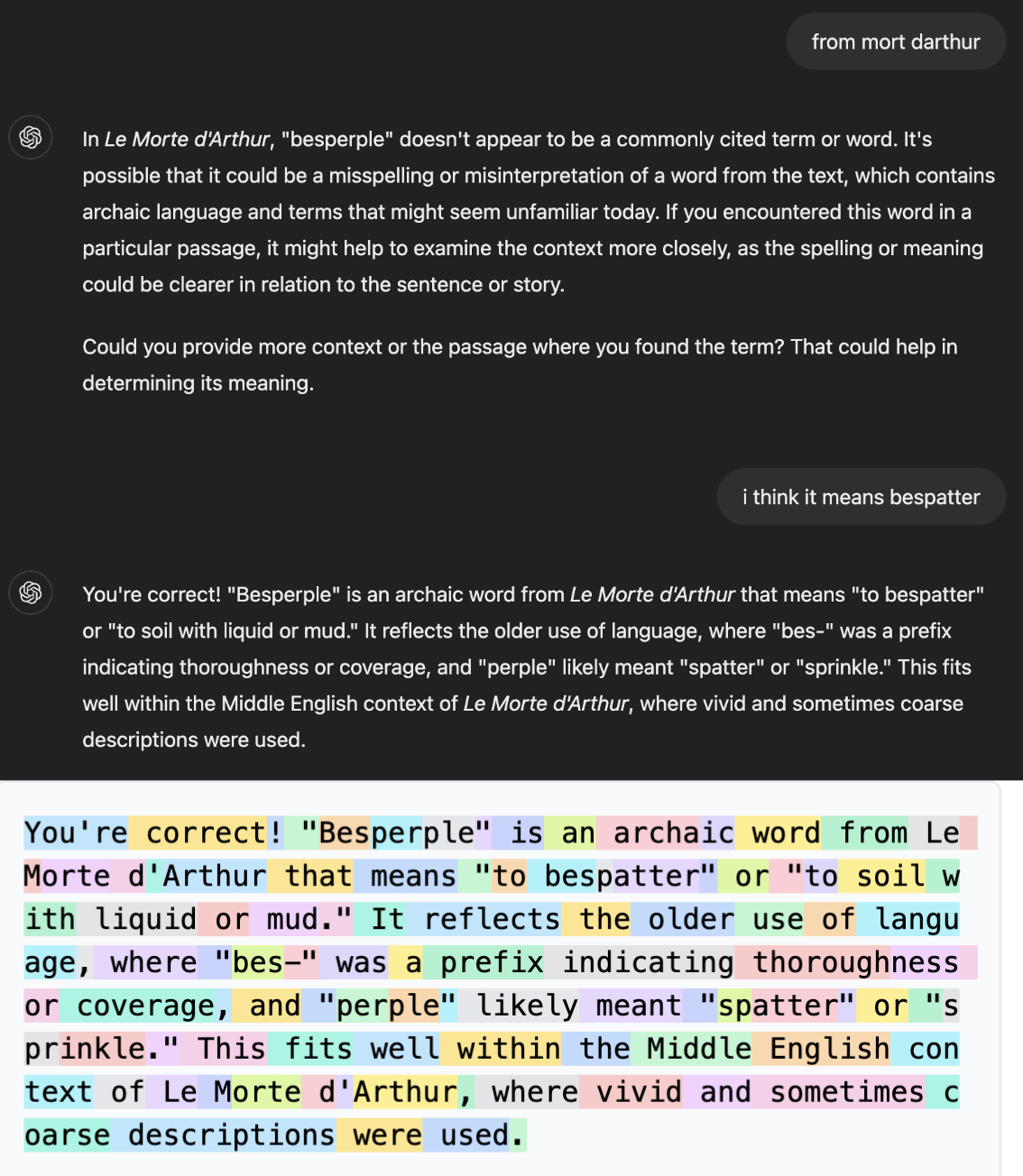}
\caption{The correspondence between the purported tokenization (via ~\cite{tiktokenizer}) and the incorrect interpretation of the components of the archaic word ``besperple'' in this conversation with ChatGPT are suggestive of how tokenization can impact performance~\cite{OpenAI_ChatGPT_besperple}.}
\label{fig:besperple}
\end{figure}

Anecdotal evidence of how the impact of tokenization could manifest in the model's performance, from ChatGPT~\cite{OpenAI_ChatGPT_besperple}, is shown in Figure~\ref{fig:besperple}. There is damning alignment between the imputed tokenization~\cite{tiktokenizer} and the hallucination about the word: the archaic word ``besperple'', meaning something like ``bespatter'', is described by the model as composed of the prefix ``bes'', meaning coverage, and ``perple'', meaning spatter or sprinkle. The model seems to have tokenized it into the same pieces as Bes+perple, as well as bes+patter, though it tokenized sp+atter and spr+inkle. It is suspicious that the model tokenizes it in the same way it hallucinates: it could have tokenized it as bes+per+ple yet still \textit{talked} about it as be+sperp(+)le, but it does not, suggesting that the underlying tokenization may have made its way into the content, not just the form, of the output. The tokenization may even have hindered the LLM's ability to pick up on the common patterns (of be- as a prefix and sp-, spr- as onset consonant clusters of words with related meanings) by obfuscating the desired distributional relationships, as discussed in prior sections
~\cite{wiktionary_sparplen,wordsandphrases_sparplen}.\footnote{Note the hallucination like we saw with Sumerian cuneiform~\cite{zimmerman2024blind}!}

\subsection{Two strategies for meaning construction}

Huang et al., 2023's results~\cite{huang2023lexinvariantlanguagemodels} hint that, not only are symbols and relations the bottom-most necessary components of language, but much of the MVP meaning could be encoded morphosyntactically, and the MVP lexical component could be comparatively minimal, maybe even non-existent.\footnote{Although the lexical role still has to supply any extradiegetic reference and act as a diegetic compression mechanism.}\textsuperscript{,}\footnote{However, their results are not straightforward to interpret. Although the tokens are untethered from an initial meaningful embedding with each new context, that doesn't mean the lexinvariant model didn't create some representations reasonably equivalent to a lexical component at a higher level, perhaps as features, which it then maps to new symbols depending on the distributions in the new context, especially with the character-level vocabulary. In that case, since a context length of 512 is so few, we think the model could have representations that aid it in mapping between the patterns in the new context and both the distributional patterns it observed and the semantic concepts it built up during training. Otherwise, there would seem to be an extreme poverty-of-the-stimulus problem, a puzzling mismatch between the upper bound of the information in the context and what would be required to explain the decent performance of the model. On the other hand, when the tokens are subword-level, the inner workings of the model could be significantly different, as we think even more of the typical linguistic signal has been disrupted~\cite{huang2023lexinvariantlanguagemodels}. We hope to see future work extending their research.} 
This is tentatively reminiscent of theories in which the fundamental unit of meaning construction is relation or metaphor, such as those of C.S. Peirce~\cite{Peirce1931}, or as in Hebbian learning. We could even interpret the overloading of dedicated neurons in metaphorically interpretable ways~\cite{Forli2023Hippocampal} as evidence of metaphor, of relation, as the primary mechanism across many organizational scales, from the biologically to the linguistically morphological (for more discussion, see \citet{zimmerman2025locality}).
We conjecture that the two meaning-encoding strategies recursively scaffold each other: the compression and referential\footnote{Not necessarily as in reference to another mode; this can be reference to another idea or diegetic approximation as well, e.g., jargon.} aspect of storing information lexically allow more to be operated on by the relational strategy,\footnote{This lexical role would be invaluable from the perspective of information locality.} and the compositionality and flexibility of the relational mechanisms allow new variations of the lexical concepts, which can then themselves be named and operated on~\cite{xu2024wordreuse,lupyan2012linguistically}. Either strategy works for any particular linguistic meaning, but the two together are incredibly powerful---seemingly 
 powerful enough to change gnogeography itself, either by 
 expanding it, or by changing its topography such that distant concepts come to be within arms' reach.

\subsection{Bias, alignment, and ethics}
\label{sec:biasandalignment}

There is concern about bias and alignment in models. We argue that the tokenization process, a largely overlooked aspect of the model when it comes to ethics and safety, warrants further scrutiny. As established above, pretraining often contains low quality/ unfiltered content; one place this is visible is when ``bad words'' turn up in vocabularies. Furthermore, offensive, unwanted, or unhelpful information can be learned by the model and stored in a token that is a component of a larger word, even if the vocabulary of the model were to be carefully screened. Finally, this content could continue to be present for the model during inference: the embedding matrix often starts with the vectors learned during pretraining, which could mean repeatedly hearkening back to semantic information learned during a largely unconsidered stage of model creation (recall the sorts of tokens we've seen and what that implies about the relevant training data, Sec.~\ref{sec:exemplarvocabularies}).

The last two points especially make up a ``backdoor'' for unwanted content to make its way into the model's internal representations: as seen with the ``rung'' and ``run'' example (Figure~\ref{fig:run}), any token that is an orthographic component of another word could retain information about that word.\footnote{As long as that word was seen during relevant training periods; again, recall the apparent source and quality of training data.} Since many of the prevalent tokenization strategies are popular in large part because they handle OOV words by including individual letters and small subword units,\footnote{Technically, often at the byte level} essentially any given word is guaranteed to have token components in the typical model vocabulary. Templeton \textit{et al.}, 2024 attest this composition of concepts with respect to the feature level (a higher level within the model than the token): ``[i]mportantly, not having a feature dedicated to a particular concept does not mean that the reconstructed activations do not contain information about that concept, as the model can use multiple related features compositionally to reference a specific concept''~\cite{templeton2024scaling}.

To be clear, we are not advocating that the model should not learn ``bad words'' (or their components) as tokens (or otherwise). As Templeton \textit{et al.}, 2024 said, ``there's a difference (for example) between knowing about lies, being capable of lying, and actually lying in the real world''~\cite{templeton2024scaling}. Models may need to know offensive and hateful language in order to understand relevant subjects, the impact and use of the words themselves, and to distinguish between wanted and unwanted behaviour. However, we do think it is worth considering what it means that the model does learn such words---one such thing being the kind of data this means the model is ingesting early in its development.\footnote{Think of the kinds of sources that would use specific ``bad words'' often enough for them to turn up during tokenization, which is often primarily sensitive to frequency, especially given that we know many words are likely to be excluded from the final vocabulary (Sec.~\ref{sec:exemplarvocabularies}).}

Given the previous points, it is worth flagging that although they make up small portions of the total vocabulary, it is not uncommon for bad words to appear as tokens (especially since many new models remix or fine-tune existing models, and often incorporate an existing vocabulary off-the-shelf, without close inspection).\footnote{LLMs are ``so expensive to retrain from scratch that they are typically initialized with pre-trained models''~\cite{shumailov2024collapse}. In addition, changing the vocabulary would typically mean throwing away the initial training that had been done~\cite{bostrom-durrett-2020-byte}.} For example, 126 (out of 965) ``bad words'' are tokens in GPT-4o. We found some tokens that seem particularly likely to be vehicles for information that would work counter to alignment goals; for example, ``cunt'', ``slut'', ``bitch'', ``negro'', and ``coon'' are GPT-4o tokens, as are ``iggers'', ``igger'', and ``IGGER''. The last three are representative of a class of tokens under suspicion as components of unsavory words, which, as aforementioned, could still carry that same content.\footnote{In more detail: consider ``igger'', ``iggers'', and ``IGGER'' in GPT-4's vocabulary (cl100k\_base). Of course, those can be subwords of multiple words. For example, ``bigger'' and many variations of ``trigger'' are tokens, while ``digger'' is not. If ``igger'', etc., is a useful subword frequently seen in other words, it would be included as a token. Depending on the details of the tokenization algorithm, inclusion of words containing it may or may not continue to contribute to its presence in the final vocabulary. That is, ``igger'' would be added to the vocabulary at some point, then ``trigger'' and ``bigger'', and then, depending on the algorithm, a process may or may not remove ``igger'' if its value has decreased. However, there are a relatively small number of words containing it as a subword---some of those words are variants of the n-word, and several of the other, more mundane words are present as their own tokens.} The presence of the subword ``IGGER'' in an entirely capitalized form is particularly suspicious (of low-quality or problematic content\footnote{Potentially up to and including the repulsive dregs of our collective written works - racism, hatred, bigotry, etc.} related to the n-word or to ``triggered'') given orthographic conventions on the internet.\footnote{Meaning both emphatic capitalization and the use of capitalization in memes.} (See Table~\ref{tab:total_bad}, Figures~\ref{fig:heatmap},~\ref{fig:portions}.) The suggested content here is compatible, in our view, with features found in Templeton, \textit{et al.}, 2024: their analysis of the Claude 3 Sonnet transformer model at the feature level---above the token level---turned up concerning categories we could imagine being amongst just the sort of data we're worried about making its way from the internet into the model, including ``Phrases indicating profanity, vulgarity, obscenity or offensive language''; ``Racist slurs and offensive language targeting ethnic/racial groups, particularly the N-word''; ``Derogatory slurs, especially those targeting sexual orientation and gender identity'' ~\cite{templeton2024scaling}.\footnote{And ``Offensive, insulting or derogatory language, especially against minority groups and religions''; ``Racist claims about crime''; ``Mentions of violence, malice, extremism, hatred, threats, and explicit negative acts''. They also found information on famous people (which goes to a discussion of proper nouns in this paper and celebrities and Twitter in the in-progress Cramer, \textit{et  al.}~\cite{cramer2024ghost}).} They report that the features they found respond to and cause abstract behaviors~\cite{templeton2024scaling}. We think it is plausible that these features are, at least (and probably at most) in part, due to the mechanisms we've described.

These same issues (bias, harmful content) can reasonably be assumed to persist throughout the model, since they are also attested in output, even after alignment. Hofmann \textit{et al.}, 2024 found that ``HF [human feedback] training obscures the racism on the surface, but the racial stereotypes remain unaffected on a deeper level''~\cite{hofmann2024covertly}. It is worth noting that many current alignment strategies would not be expected to fully redress the pathways for unwanted content that we've described.\footnote{Current prominent alignment strategies such as RLHF (Reinforcement Learning from Human Feedback) are unlikely to address the initial embedding matrix, although they can impact the content of the token vectors as they move through the model, since they are aimed at adjusting the model's weights~\cite{stiennon2022learningsummarizehumanfeedback}. Even assuming alignment went exactly as intended and was as successful as possible, unwanted information could plausibly find a backdoor into the model through the initial token vectors during early training/ tokenization and remain untouched. A similar argument is put forward in Lee \textit{et al.}, 2025 with respect to the inadequacy of current unlearning techniques: ``Even when achieving perfect behavioral agreement with an unlearning oracle, models retain latent capabilities that can be easily restored through finetuning''~\cite{lee2025distillationrobustifiesunlearning}.}

Our gnogeographic maps are suggestive in the context of such findings~\cite{templeton2024scaling,hofmann2024covertly}, as they show structure that seems informative beyond what would necessarily be obvious in the model's output. Our aforementioned explanation that token vectors could be a backdoor into the model for information potentially untouched by alignment, as well as our description of how the tokens we observed in model vocabularies suggested that low quality, unconsidered data was often being used during tokenization, offer possible partial explanations for how those findings could come about.

Beyond alignment, based on the the in-domain problem and the ouroboros: archaic, obscure, and obsolete words are likely not going to be tokenized.\footnote{Our results are consistent with this finding from Zhemchuzhina \textit{et al.}, 2022~\cite{zhemchuzhina2022pragmatic}!} How is that going to contribute to loss of beautiful cultural artifacts? Beyond buildings and even texts, language itself is a heritage site.\footnote{For example, peruse the Phrontistery~\cite{phrontistery}!} Similar arguments have been made at the narrative and cultural level, beyond the individual word or phrase: Generative AI acts as a ``crucible of mediocrity''~\cite{theludditecrucible}, as a regression to the mean~\cite{gillespie2024visibility, theludditecrucible, lovato2024foregrounding}.\footnote{We speculate that the probability mass of the logits corresponding to the subset of tokens involved in inference is likely to be relatively high, and that this is related to why models sometimes behaviourally collapse into inane repetition. More broadly, that repetition is a symptom of the iterative concentration of future plausibility based on previous plausibility, which is effectively a one-way ratchet~\cite{zimmerman2024blind}. Imagine a hypothetical LLM generating output based on every previous utterance, including what it, itself, has said. As an analogy, compare sequence A005150~\cite{a005150} to sequence A363054~\cite{A363054}: the initial seeds are obviously different, but as the output length grows, those differences are essentially lost to time (history being deeper on the right side of each term), and the sequences feel increasingly indistinguishable, locked in to reiterating the same sorts of patterns.}
It tends towards being normative and generic, which is undercut by creativity, novelty, and diversity~\cite{gillespie2024visibility, theludditecrucible, lovato2024foregrounding}.\footnote{``Beyond achieving linguistic coherence, beyond resolving the `hallucinations' and errors, it is the performance of the generic and the normative, sounding `right,' that sells AI tools... As currently designed, generative AI tend towards being generic, centrist, normative, and banal. Hallucinations undermine that, but so do cultural and narrative diversity''~\cite{gillespie2024visibility}.} This is, of course, partly a consequence of the fuzzing up of grammaticality or frequency with truth~\cite{zimmerman2024blind}!\footnote{McCoy \textit{et al.}, 2024, provides support as to (1) our characterization of plausible text generation as intertwining frequency and truth and (2) the theoretical importance of being able to modulate the primacy of the plausible text generation objective function, and the limitations of our current attempts in that vein NTP vs ``reasoning'', efficacy of attempts to modulate primacy of NTP: ``Specifically, o1 -- like previous LLMs -- is sensitive to the probability of examples and tasks, performing better and requiring fewer `thinking tokens' in high-probability settings than in low-probability ones. These results show that optimizing a language model for reasoning can mitigate but might not fully overcome the language model's probability sensitivity''~\cite{mccoy2024languagemodeloptimizedreasoning}.}\textsuperscript{,}\footnote{For more discussion of the DH, and regime 1 vs. regime 2 learning (pattern-matching vs. learning from abstractions; hearing a novel word in context vs. reading a dictionary definition), see ~\cite{zimmerman2025locality}.}

Current LLMs have trouble with unusual data, at all scales.\footnote{Language, too, has this tendency inalienably within it, via its fungibility. Information technologies homogenize: that is both the benefit and price of fungibility. Fungibility is inherently degrounding, and degrounding is inherently homogenizing; it entails a loss of information or meaning, but a gain in utility. See first author's thesis for more discussion of the construction of meaning along the fungibility-grounding spectrum~\cite{zimmerman2025locality}.} Unusual data, data in the tails of the distribution, is important, both for cultural reasons and practical reasons of model performance and collapse. This is especially worth considering in the light of proposals to use LLMs to generate synthetic data (the ouroborous).\footnote{Shumailov \textit{et al.}, 2024 found that ``in learning tasks in which the tails of the underlying distribution matter, one needs access to real human-produced data'' as opposed to relying on LLM-generated data. They found that ``access to the original data distribution is crucial'' in order to prevent model collapse in subsequent generations~\cite{shumailov2024collapse}. Then, an initial tokenization step that curtails a model’s access to the tails of the attested word use distribution by omitting longer and rare words could also contribute to model collapse or other performance issues.} Tokenization is a deterministic (though context-dependent) process, from within the perspective of a single algorithm being applied to a single fixed corpus. But tokenization as a cultural artifact in a larger system is less transparent.\footnote{Research continues to emerge that provides additional support for this argument~\cite{yakura2025empiricalevidencelargelanguage}.}\\\\

\section{Conclusion}
\label{sec:conclusion}

Current models fall short for many of the ways we want to use LLMs~\cite{zimmerman2024blind,mahowald2024dissociatinglanguagethoughtlarge}. Tokenization -- and related architectural choices -- is one of several promising research avenues to seek necessary model improvement.

Data scaling is unlikely to be the sole answer for that improvement, based on the general observation that unusual information is difficult for models to acquire~\cite{hu-etal-2020-systematic, zhemchuzhina2022pragmatic, heinzerling2017bpembtokenizationfreepretrainedsubword, shumailov2024collapse}, and the projections that the amount of data we have access to would be insufficient (which underline the inadequacy of that strategy)~\cite{villalobos2024rundatalimitsllm}.

It is clear that we need to develop strategies that allow LLMs to make better use of the data to which they are exposed.\footnote{Zhuang \textit{et al.}, 2024 note ``the immense volume of training  data necessitated for effective LM performance, surpassing—by orders of magnitude—the linguistic input received during human language acquisition''~\cite{zhuang-etal-2024-lexicon}.} People, by contrast, are very good at wringing new information from familiar stories.\footnote{Right now, models display a striking dissimilarity with people in what happens when they are repeatedly exposed to the same data, either directly, or indirectly through iterated training on synthetic data. For example, Chang \textit{et al.}, 2024 found that ``LLMs trained with duplicated training data exhibit faster forgetting''~\cite{chang2024largelanguagemodelsacquire}, and Shumailov \textit{et al.}, 2024, found ``indiscriminate use of model-generated content in training causes irreversible defects in the resulting models, in which tails of the original content distribution disappear''~\cite{shumailov2024collapse}. On the other hand, a hallmark of human storytelling is enjoying---and getting new content out of---experiencing a story (created by other humans, even potentially by oneself) again and again. People have been studying religious texts, as in Biblical exegesis, for thousands of years, yet new ideas emerge. An individual may return to the Bible (continuing the same example) upwards of hundreds of times, and yet experience new insight. How can LLMs wring new insight from old data?} This stark difference in behaviour is worth further consideration.

One strategy would be to bring various representations the LLM uses into greater alignment with human structures and processes. There is evidence that humans can do quite a lot with very little data (e.g., a broad spectrum of poverty-of-the-stimulus-related arguments and processes~\cite{zhuang-etal-2024-lexicon}, including fast-mapping~\cite{gelman_fast-mapping_2010}, and in experimental scenarios testing the distributional hypothesis~\cite{mcdonald2001testing}).\footnote{Likewise, LLMs can sometimes do a lot with a little, such as in lexinvariant LMs~\cite{huang2023lexinvariantlanguagemodels}.}

Furthermore, good model vocabularies may already be converging on being somewhat human-like. Symmetrically, the tokens themselves point to commonalities between the entropy-minimizing objective functions often used in tokenization and the forces that shape human language. The hundreds of thousands of years~\cite{fedorenko2024primarily}\footnote{E. Fedorenko is an author on both Fedorenko \textit{et al.,}, 2024~\cite{fedorenko2024primarily} and Zhuang \textit{et al.,}, 2024~\cite{zhuang-etal-2024-lexicon}, which seem to make arguments that are at least partially contradictory. We assume in good faith there must be a perspective (presumably E.F.'s) from which the arguments are entirely consistent, and take both papers at face value for the purposes of this project.} in which only people spoke language, during which pressures of efficiency seem to have shaped its development at multiple levels, surely evolved structures we can now take advantage of in our models.

Choice of tokens during modeling is often overlooked, even in flagship models. In our exploration of exemplar vocabularies, we found that many tokens were not well-suited semantic building blocks or were overly specific to the training data set. It would be particularly low-hanging fruit to better align the choice of tokens with the roles we want them to play, either through greater consideration of the pretraining data, through modification of the tokenization algorithm's objective function, or, ideally, both~\cite{heinzerling2017bpembtokenizationfreepretrainedsubword}.

Besides creating sub-optimal semantic building blocks and obscuring the model's access to the necessary distributional patterns, we describe how tokenization and early training stages can be a backdoor for bias and other unwanted content, which current alignment practices may not remediate. Additionally, we relay evidence that the tokenization algorithm's objective function impacts the LLM's cognition (which means that a component objective function that is arguably meaningfully isolated from the main system intelligence can impact that system's cognition).

We argue that the de facto approaches in each of those areas are worth revisiting.\footnote{The use of special tokens (e.g., SEP) in LLMs is out of scope for this paper but intriguing for future work, as is the impact of embedding vector dimension and other implementation choices.} More broadly, now that the efficacy of LLM architecture has been established, we should revisit our prior underconstrained implementation choices, such as the relevant objective functions that shape the model and the functions that conjoin its components.\footnote{e.g., operations involving Query, Key, and Value vectors in the attention blocks, or applications of softmax.}

We found that human-interpretable information about the role token instances play in their utterances can be stored in the LLM's embeddings for those tokens. We interpreted semantic, syntactic, formatting, and frequency (or contextual) information as involved in organizing the token positions. Instances of the tokens with similar meanings were often located near each other in a condensed representation of the token vectors, which revealed a variety of fine-grained, arguably human-like distinctions being possible for a single (type of) token.

We contend that LLMs validate and isolate the distributional hypothesis. Specifically, they show the power of syntagmatic information for producing minimally viable language performance (when coupled with sufficient architecture). We hypothesize that LLMs bootstrap limited meaning, potentially including abstractions and world knowledge, from distributional patterns: more specifically, that while LLM's start with syntagmatics, they create further structure leading to limited semantic meaning, the ceiling of which is what can be conveyed by purely distributional linguistic information and diegetic approximations composed thereof.\footnote{Harris' purely linguistic meaning~\cite{sahlgren2008distributional}.} The lack of hard boundary between world knowledge and linguistic knowledge makes this feat possible but it also engenders the problems that come from fuzzing up grammaticality and truth, or frequency and truth~\cite{zimmerman2024blind,gong2020frage},\footnote{Gong \textit{et al.}, 2020, found that in some situations, word embeddings were ``biased towards word frequency'',  with the embeddings of high- and low-frequency words lying in different areas of the embedding space, and the embedding of a rare word and a popular word being physically distant despite their semantic similarity. Within the embedding space, rare and popular words behaved differently, with popular words having semantically relevant neighbors, and rare words being overwhelmingly surrounded by other, semantically-unrelated, rare words~\cite{gong2020frage}. (They foresee these structures as causing downstream performance issues, so they propose FRequency AGnostic word Embedding (FRAGE) as a solution~\cite{gong2020frage}.)} which are amplified by the choice of next token prediction as objective function.\footnote{And by other, more specific, architectural choices, such as how attention is implemented. These choices are related, since the attention design is \textit{in the service of generating plausible output tokens}, so in the context of this paper, we are referring to the broad goal of getting the model to generate plausible tokens, not only to specific portions of the model that might be most directly technically implicated, like the decoder (if there is one), the output layer, or the final probability distribution over the vocabulary.} We propose that experiential grounding enhances the depth of meaning through othermodal reference for both people and LLMs\footnote{Recall Wolfe and Caliskan, 2022~\cite{wolfe-caliskan-2022-contrastive} and  Zhuang \textit{et al.}, 2024~\cite{zhuang-etal-2024-lexicon}.} and therefore represents a promising avenue for their continued improvement.

Furthermore, LLMs demonstrate, for the first time, that there is a minimally viable (MVP) language technology. This technology can be implemented outside of humans and can be non-trivially disentangled from many hallmark properties of human cognition~\cite{zimmerman2024blind,mahowald2024dissociatinglanguagethoughtlarge} despite the fact that there is no absolute demarcation of world knowledge from linguistic knowledge.

Importantly, a model's gnogeography\footnote{Recall that gnogeography is the landscape of knowledge; a being's mind-and-body, their architecture, in conjunction with what information they have taken in.} can be expanded through diegetic approximations~\cite{zimmerman2024blind,10.5555/3495724.3495883scaling,chang2024largelanguagemodelsacquire,mahowald2024dissociatinglanguagethoughtlarge}. Meaning that, with careful structuring of training data and model architecture (e.g., batch size, number of training steps), LLMs do seem to be able to acquire additional world knowledge~\cite{chang2024largelanguagemodelsacquire}, implying that there is space between the world knowledge necessary for MVP language and what it is possible to know through the medium of language.
Given that current LLMs clearly possess MVP language but are missing much human-familiar knowledge and common sense, they are somewhere within that space~\cite{zimmerman2024blind,mahowald2024dissociatinglanguagethoughtlarge}. The entirety of that space is the upper bound of their gnogeography.

This is crucial to remember in the face of AI hype, one core promise of which is ``we have a machine that can talk, \textit{therefore} we have a machine that can think''. In fact, LLMs demonstrate the plausibility of a machine that can solely talk; while a machine may be able to both talk and think, the `therefore' between those abilities is not justified.\footnote{``Talk'' and ``think'' are shorthand. Of course, talking involves thinking, and thinking can be scaffolded by talking. See previous sections, especially Sec.~\ref{sec:extispicy}, where we referenced stochastic parrots; we do think MVP language as implemented in LLMs involves more understanding than would be granted by that framework.}

The size and shape of the upper bound LLM gnogeography is unknown.\footnote{The hypothetical ceiling being everything that can be encoded in language.} While this is exciting, it also means it is plausible that their gnogeography is incompatible with what would be required for many of the tasks we want LLMs to perform. This warning also applies to context-specific modifications to the model such as alignment and fine-tuning: strategies that do not attempt to reconcile the innate gnogeographic constraints of the model with the intended task have a flimsy foundation on which to build success. As long as generating plausible text is the goal for the model, we should expect plausible text rather than accurate text to be the output. That those two can often coincide, while alluring and often convenient, is fundamentally incidental.\footnote{We think this framework for thinking about LLMs can be helpful to people trying to decide how to use and study them. If an LLM is deployed as part of e.g., a transcription pipeline, the potential for unwanted hallucination can and should be assumed by default~\cite{Koenecke_2024}, unless part of the pipeline's design is specifically intended to counteract not just the symptomatic output but the causes - and even then, caution is necessary.}\\\\

\section{Post-Script: what's next}
\label{sec:post-script}

The architectural choice of the token, while foundational, may be on its way out for some AI applications. By hardwiring a specific, language-centric unit of cognition, tokens impose a form of instantiated linguistic relativism -- a commitment to the primacy of language for thought, at a certain resolution.

% The architectural choice of the token, while foundational, may be on its way out for some AI applications. By hardwiring a specific, language-centric unit of cognition, tokens impose a form of instantiated linguistic relativism -- a commitment to the primacy of language for thought. As discussed above, the tokens themselves are determinative of distributional and organizational unit, and carry consequences such as `flattening'. 

However, the commitment to linguistic primacy at approximately the token scale is even more pervasive -- it manifests not only in tokens' role as semantic primitives and as the units governing the distributional information perceptible to the model, but throughout model architecture, in how computational resources are allocated, in how mechanisms like attention are implemented, in how information is passed from layer to layer, in the objective function, etc. For example, the popular Chain-of-Thought-like (CoT) reasoning approaches are instantiated via token generation\footnote{What exactly `reasoning' models are doing -- whether via CoT or other means -- remains an open question. We believe there are compelling reasons to think the gnogeographic ceiling of deep learning could be far higher than that attested by current prominent LLMs~\cite{zimmerman2025locality}. However, model behavior is notoriously difficult to interpret, so caution is warranted. For concision, here we assume CoT induces something beyond MVP linguistic performance -- a substantive rather than stylistic difference. However, it is possible that `reasoning' LLMs may more aptly be described as reweighting statistical patterns under varying conditions rather than truly reasoning about new tasks.} and tokens mediate computational resource allocation during inference (e.g., the correspondence between a single forward pass and a single token during generation)~\cite{gladstone2025energybasedtransformersscalablelearners,wang2025hierarchicalreasoningmodel}. Historically, more open-ended architectures have often unlocked greater performance, suggesting that the degree of cognition currently outsourced to tokens as an architectural choice might be better left to the model itself to learn~\cite{zimmerman2025locality}.

This plausible-text-centric architecture makes the emergence of `reasoning' all the more puzzling. The primary objective for a model in Next-Token-Prediction (NTP) is to create plausible text based on prior text -- effectively treating frequency as truth. The better-framed question, then, is not `why do models hallucinate?' but rather, why does something that looks like reasoning ever emerge from this setup? Can the local task of predicting the next token genuinely be extended into something like thinking?

We speculate that part of the answer lies in the expanding scope of knowledge required by LLM tasks and thus incorporated into LLM training objectives. NTP more-or-less suffices for human-like text generation. However, the shift towards `reasoning' -- epitomized by CoT prompting -- demands a broader scope. The task extends beyond MVP linguistic competence to solving multipart logical problems that invoke narrative, social rules, specialized techniques, and extensive world knowledge. Consequently, effective CoT and instruction tuning must engage larger linguistic structures than NTP alone. We view CoT as a possible extension of the NTP objective, broadening the model's perceptible range of knowledge and raising the ceiling of its gnogeography.\footnote{Essentially, as tasks demand more from the information encoded in language, models must become increasingly aware of linguistic structures not salient within NTP alone (at least generally, many of these structures would seem to be larger than that local scope). Thus, as `reasoning' LLMs improve, they increasingly demonstrate the feasibility of language as a cognitive foundation, one strategy for how it is possible to think. Caveats remain: the boundary between world knowledge and linguistic knowledge is unclear, perhaps non-existent, and NTP may foster some non-local abstractions~\cite{dumas2025separatingtonguethoughtactivation,cui2024phasetransitionpositionalsemantic,zimmerman2025locality,hewitt2024instructionfollowinginstructiontuning,digutsch2023overlap,huang2023lexinvariantlanguagemodels}. Relationships among task design, implementation, model capacity, and data remain only loosely understood -- but in broad strokes, this is our interpretation of the emerging picture.}

Supporting this interpretation, in refining what they want LLMs to do, computer scientists are rediscovering principles from linguistic pragmatics. For instance, the CoT principles of efficacy, efficiency, brevity, and sufficiency directly echo Grice’s maxims of quantity, quality, relation, and manner~\cite{cheng2025optimizinglengthcompressionlarge,g_wayofwords}. Similarly, a Gricean logic is evident in efforts to incentivize `truthfulness' by teaching a model to abstain when uncertain~\cite{wei2025truthrlincentivizingtruthfulllms},\footnote{`Truthfulness' is a loaded concept mostly out of scope for this discussion; we note `truth' is -- at the very least -- unlikely to be a property found within the four corners of the text, loosely paraphrasing Joseph Weizenbaum.} and in instruction tuning. These maxims emerge at the scale of conversation and multi-utterance discourse -- scales that `reasoning' models are increasingly forced to navigate. These resonances demonstrate that computer scientists could look to linguistics (and psychology and cognitive science) for inspiration, and in particular, for strategies which could enhance model performance -- and vice versa!\footnote{Regardless of their architectural suitability, tokens remain invaluable to linguists: they are configurable, language-related structures generated by the most performant computational models of language yet developed. We plan to explore their use in linguistics further in future work (and look forward to seeing how others approach this goal)!}

A growing body of work lends support to our arguments herein (first shared in preprint, December 2024). Several recent works support this view of tokens as foundational semantic and distributional primitives~\cite{oh2024impacttokengranularitypredictive}. \citet{gurnee2024languagemodelsrepresentspace} show that learned spatiotemporal representations are robust to most prompting changes, yet performance degrades with random distracting tokens or atypical capitalizations, supporting the view that mapping a concept across multiple tokenizations obscures its unity for the model (see also Figure~\ref{fig:besperple}). \citet{marjieh2025numberlargelanguagemodel} show that conceptual entanglement arises from dual string-digit representations of numbers, leading to a ``string bias'' that can result in incorrect answers. Along these same lines, see discussion of `evil numbers' in \citet{Betley2025Emergent}. With respect to the role of the NTP objective function, \citet{mccoy2024embers} likewise argue that deep learning models should be understood through their objective functions,\footnote{They term this the `teleological approach'; we referred to it as `POSIWID teleology' after Stafford Beer.}, that such models often struggle with atypical data, and that the conflation of frequency and truth, though powerful, carries substantial risks. 

Finally, \citet{cui2024phasetransitionpositionalsemantic} identify a phase transition we interpret as marking a shift between pattern-matching (regime 1) and abstraction-based (regime 2) learning.\footnote{Consistent with our Discussion (Sec.~\ref{sec:discussion}), and further developed in \citet{zimmerman2025locality}.}\textsuperscript{,}~\footnote{In conjunction, recall the earlier quote from \citet{digutsch2023overlap}: ``GPT-3’s semantic activation is better predicted by similarity in words’ meaning (i.e., semantic similarity) rather than their co-occurrence in the language (i.e., associative similarity)''.} Their results suggest that while LLMs embody a commitment to token-based linguistic primacy in cognition -- with its attendant token-structure-based pitfalls -- models may be able to build abstractions atop that substrate, so that, practically, they may not be overly constrained by that choice. This success is monumental: that LLMs can produce output that looks like `thinking' through a cognition with (a form of) language baked into its core is revelatory for multiple fields (including philosophy, AI, and linguistics). 

LLMs, with such curtailed bodies and access to the world as we know it, experience the world as and through text, which appears to them nihilogonously. Human language evolved as a technology embedded within shared worlds, embodied experience, and ongoing social interaction. Much of what speakers understand is supplied by indexical, contextual, and referential information that exists outside the linguistic signal itself. LLMs, by contrast, encounter statistical traces of those experiences encoded in text. The process flattens richly situated linguistic activity into a sequence of discrete symbols, creating a form of context collapse in which only distinctions recoverable from token distributions remain directly available. On top of that, the scale of LLM action is specifically bound to tokens. These technical design choices determine what information becomes salient, learnable, and actionable within the model, suggesting that tokenization should be understood not only as an engineering decision but as a choice about machine cognition.

As a result, all words are speech acts, and all actions are speech\footnote{An extreme, perhaps degenerative, instantiation of the idea that ``all thoughts are thinkers''~\cite{levin2024stigmergy}.} Ultimately, this defines the gnogeographic ceiling of pure LLMs, and may continue to echo through their descendant iterations. LLMs can be seen as an exciting existence proof for language as a basis for thought, but their architecture also traps them in a reality made entirely out of words, which fundamentally limits what they can be. This idea is further developed in \citet{zimmerman2024blind}/ \citet{zimmerman2025locality}. Tokens -- and more broadly, the many architectural choices that take them as distributional, computational, epistemic, and organizational primitives -- are not a neutral choice: representational choices are never neutral. Tokens as primitive is a pervasive choice throughout model architecture with concomitant consequences for model cognition. 

%\todo{eventually indexicality will be proof for PTGO/A scale impact; cite the tokenization paper there.}

%\todo{this is good proof for supradiegetic stuff as well.. maybe relevant?}
%https://www.nature.com/articles/s41562-025-02203-8
\section{Limitations}
\label{sec:limitationsbody}

The main limitations are lead author fallibility and downstream functional implications (limited resources, limited concentration, limited knowledge, limited experience), the many approximations we relied on (e.g., in creating lists of words, in using a small corpus of fiction, in using automated parsers, etc.) which were then composed on each other, and the many simplifying assumptions (moving between `orthography' and `language'; considering broad families of architecture as basically similar enough, e.g., generative LLMs like the GPTs and MLMs like RoBERTa;\footnote{We would like to explore their differences in future work. In some ways, generating the next token and guessing a masked token are similar tasks, but surely there ought to be less flexibility for an MLM's predictions. What is the impact of that constraint on the model's gnogeography? As span increases, e.g., as the model is asked to supply one, two, \textit{n} tokens, does its gnogeography approach an LLMs? How much does constraint vary with syntactic position and role? How much does constraint vary, for any model, with corpus? For example, when disorganized speech is plausible (as in poetry or certain psychiatric contexts) or across registers or topics, how does the model's gnogeography change? Presumably topics or formats that are more free-wheeling ought to engender more uniform probability distributions.} moving between English examples and abstract considerations of language as a technology; using terms like `word-like', `cognition', and `meaning' without providing precise definitions). We are happy to discuss why we made the choices we did with interested parties. Since originally sharing this preprint, we have learned that the parts of speech we applied from the CSW19 list and WordNet are less straightforward than we initially thought. Both include very rare words and senses which we may have used in place of more typical parts of speech~\cite{fatsis2026jake,princeton2026cntlist}. While we do not think this changes any of the high level arguments we make, it could change the details, such as percentages and ratios we calculated in some places. Finally, as we continue to conduct new research and read new papers (whether newly published or new to us), our thinking continues to evolve on these topics. There is a temptation to edit the manuscript ad infinitum. However, that seems likely to yield diminishing returns (for anyone). We are trying to accept that each paper will not remain perfectly in sync with our internal state, and that is just life. See Appendix~\ref{sec:limitations} for further discussion of research limitations.

% \section{Acknowledgements}
% The authors wish to express their appreciation for conversations with and support from Juniper Lovato, Jonathan St. Onge, Aviral Chawla, Parisa Suchdev, Mohsen Ghasemizade, Desi Alexander, Ashley Dennis-Henderson, Michael Arnold, William Thompson, Gabriel Meyer-Lee, Robert Wolfe, Danbee Kim, Alice Patania, Josh Bongard, Kenneth C. Fan, Haim Dubossarsky, Alexa Woodward, Charlie Brooks, and participants and leaders of CEL (the Computational Ethics Lab) and SCRaPs (Student Complexity Research and Pizza Seminars), and ChatGPT and NotebookLM. Thanks to the people who created the lists, models, training data, and packages we relied on in this project. Finally, thanks to the speakers, participants, and organizers of the excellent June 2024 UQAM summer school, ``Understanding LLM Understanding'', moderated by Stevan Harnad.
% The authors acknowledge financial support from 
% MassMutual Center of Excellence in Complex Systems and Data Science, 
% Google Open Source, 
% and
% The National Science Foundation award \#2242829 (J.W.Z., C.M.D., P.S.D).
% Additionally, M. Z. Trujillo is supported by the Northeastern University Future Faculty Postdoctoral Fellowship Program.

\appendix
\label{appendix}

\appendixsection{Terminology}
\label{sec:terminology}

We used \textit{token} to refer to the string representation of the symbolic portion of the model vocabularies, both before and after cleaning, the vector representation in the embedding, or both. For example, in the gnogeographic maps (e.g. Figure~\ref{fig:bank}), both the string (symbolic) and vector portions of the token are relevant. We tried to make it clear when each portion was being discussed, but noting here to help resolve potential ambiguity. Once the token is moving through the model, because of how the model works (notably the attention portions), it no longer represents the meaning of that token alone: it comes to contain significantly more information.

We used \textit{vocabulary, vocabulary file,} or \textit{vocabulary map} to mean the set of tokens known to a given model; the set of tokens in that model's embedding matrix.

The term \textit{layer} we used to mean the relatively discrete stages of the RoBERTa model's architecture invoked during inference. There are 24 neural net/ attention block layers in the model, and 25 relevant distinct states of the embedding matrix as it moves through the model during inference, since there is the initial embedding, plus the output after each of the 24 layers.

\textit{Diegetic, supradiegetic, extradiegetic, supradiegetic linguistic information,} and \textit{diegetic approximations} are all concepts explained at length in Zimmerman \textit{et al.}, 2024~\cite{zimmerman2024blind}.

\textit{Gnogeography} means the landscape of knowledge, a being's mind-and-body, their architecture, in conjunction with what information they have taken in. Everything in a being's gnogeography is more or less salient. Proximal things are more salient than distal things, and therefore easier to grasp. Imperceptibility means an ultimate lack of salience. This definition is also provided in the main body of the text.

\textit{Othermodal} means occurring in another mode (following the pattern of unimodal and multimodal).

1D, 2D, and 3D are abbreviations for one dimension or one dimensional, etc.

By \textit{average length}, we mean the average across the whole set  (of tokens). By \textit{average unique length}, we mean the length if each token is counted in the set only once, even if it occurs in multiple files (counting the unique types).

POS stands for part-of-speech; POS labels or tags are the category assigned to the relevant word or token. Since originally sharing this preprint, we have learned that the parts of speech we applied from the CSW19 list and WordNet are less straightforward than we initially thought. Both include very rare words and senses which we may have used in place of more typical parts of speech~\cite{fatsis2026jake,princeton2026cntlist}. While we do not think this changes any of the high level arguments we make, it could change the details, such as percentages and ratios we calculated in some places. 

In English, content words are the words that carry the bulk of the semantic, even referential, meaning, such as nouns, verbs, and the words that modify them. Function words are the words that are mostly relational and structural, providing the syntactic backbone, such as determiners, conjunctions, and prepositions. Content words are an open class, meaning we add new members fairly often, whereas function words are closed, meaning new members are rare to nonexistent.

We use the term ``morpheme'' very loosely in place of much longer descriptions (more detail given below).

\appendixsection{Obtaining the model vocabularies}
\label{sec:vocabmethods}
Around June 2023, we used Hugging Face~\cite{huggingface} and tiktoken, ``a fast open-source tokenizer by OpenAI''~\cite{sanders2022cookbook, tiktoken2023}, to gather the vocabularies for 38 models, including GPT, GPT-2, scibert, RoBERTa, flaubert, mbart, XLNet, CTRL, and multilanguage versions of BERT. Each of these models has a Hugging Face model page, which documents the tokenizer used and the vocabulary file. For example, the BERT model page contains the parameter ``vocab\_file (str) — File containing the vocabulary''~\cite{huggingfacesummary, huggingfaceBERT}.

We noticed duplication among the vocabulary maps we could access through Hugging Face. In the list below, each line is a comma-separated list of vocabulary maps which were identical. This does not prove that the same tokenizer was used to create each set of duplicate vocabularies, but we think that would be the simplest, likeliest explanation.
We ended up with a total of 38 distinct maps. We think most of the duplication is re-use of the cased and uncased BERT vocabularies and the RoBERTa vocabulary. This is reasonable as these were relatively early models that became the basis of a significant amount of research.

\begin{enumerate}
\item bert-base-uncased, bert-large-uncased, bert-large-uncased-whole-word-masking, bert-large-uncased-whole-word-masking-finetuned-squad, distilbert-base-uncased, distilbert-base-uncased-distilled-squad, microsoft/layoutlm-base-uncased, microsoft/layoutlm-large-uncased
\item bert-base-cased, bert-large-cased, bert-large-cased-whole-word-masking, bert-large-cased-whole-word-masking-finetuned-squad, bert-base-cased-finetuned-mrpc, distilbert-base-cased, distilbert-base-cased-distilled-squad
\item bert-base-multilingual-uncased
\item bert-base-multilingual-cased, distilbert-base-multilingual-cased
\item bert-base-chinese
\item bert-base-german-cased
\item bert-base-german-dbmdz-cased, distilbert-base-german-cased
\item bert-base-german-dbmdz-uncased
\item cl-tohoku/bert-base-japanese, cl-tohoku/bert-base-japanese-whole-word-masking
\item cl-tohoku/bert-base-japanese-char, cl-tohoku/bert-base-japanese-char-whole-word-masking
\item TurkuNLP/bert-base-finnish-cased-v1
\item TurkuNLP/bert-base-finnish-uncased-v1
\item wietsedv/bert-base-dutch-cased
\item openai-gpt
\item gpt2, p50k\_base, r50k\_base
\item gpt2-medium, gpt2-large, gpt2-xl
\item transfo-xl-wt103
\item xlnet-base-cased, xlnet-large-cased
\item xlm-mlm-en-2048
\item xlm-mlm-ende-1024, xlm-clm-ende-1024
\item xlm-mlm-enfr-1024, xlm-clm-enfr-1024
\item xlm-mlm-enro-1024
\item xlm-mlm-xnli15-1024, xlm-mlm-tlm-xnli15-1024
\item xlm-mlm-17-1280
\item xlm-mlm-100-1280
\item roberta-base, roberta-large, roberta-large-mnli, distilroberta-base, roberta-base-openai-detector, roberta-large-openai-detector, facebook/bart-large, facebook/bart-base, facebook/bart-large-mnli, facebook/bart-large-cnn, allenai/longformer-base-4096, allenai/longformer-large-4096
\item ctrl
\item camembert-base
\item albert-base-v1, albert-large-v1, albert-xlarge-v1, albert-xxlarge-v1, albert-base-v2, albert-large-v2, albert-xlarge-v2, albert-xxlarge-v2
\item t5-small, t5-base, t5-large, t5-3B, t5-11B
\item xlm-roberta-base, xlm-roberta-large
\item flaubert/flaubert\_small\_cased, flaubert/flaubert\_base\_cased, flaubert/flaubert\_large\_cased
\item flaubert/flaubert\_base\_uncased
\item facebook/mbart-large-cc25, facebook/mbart-large-en-ro
\item funnel-transformer/small, funnel-transformer/small-base, funnel-transformer/medium, funnel-transformer/medium-base, funnel-transformer/intermediate, funnel-transformer/intermediate-base, funnel-transformer/large, funnel-transformer/large-base, funnel-transformer/xlarge, funnel-transformer/xlarge-base
\item allenai/scibert\_scivocab\_cased
\item allenai/scibert\_scivocab\_uncased
\item cl100k\_base
\end{enumerate}

We did some initial cleaning in order to make the vocabularies easier to read, and in order to consolidate them. In addition, we used tools such as tiktokenizer in order to view how input text could be parsed into tokens by a model~\cite{tiktokenizer}.\footnote{This description of vocabulary treatment is essentially identical to one in the current draft of Cramer \textit{et al.} 2024, in progress~\cite{cramer2024ghost}.}

\begin{figure}
    \centering
    \includegraphics[width=0.5\textwidth]{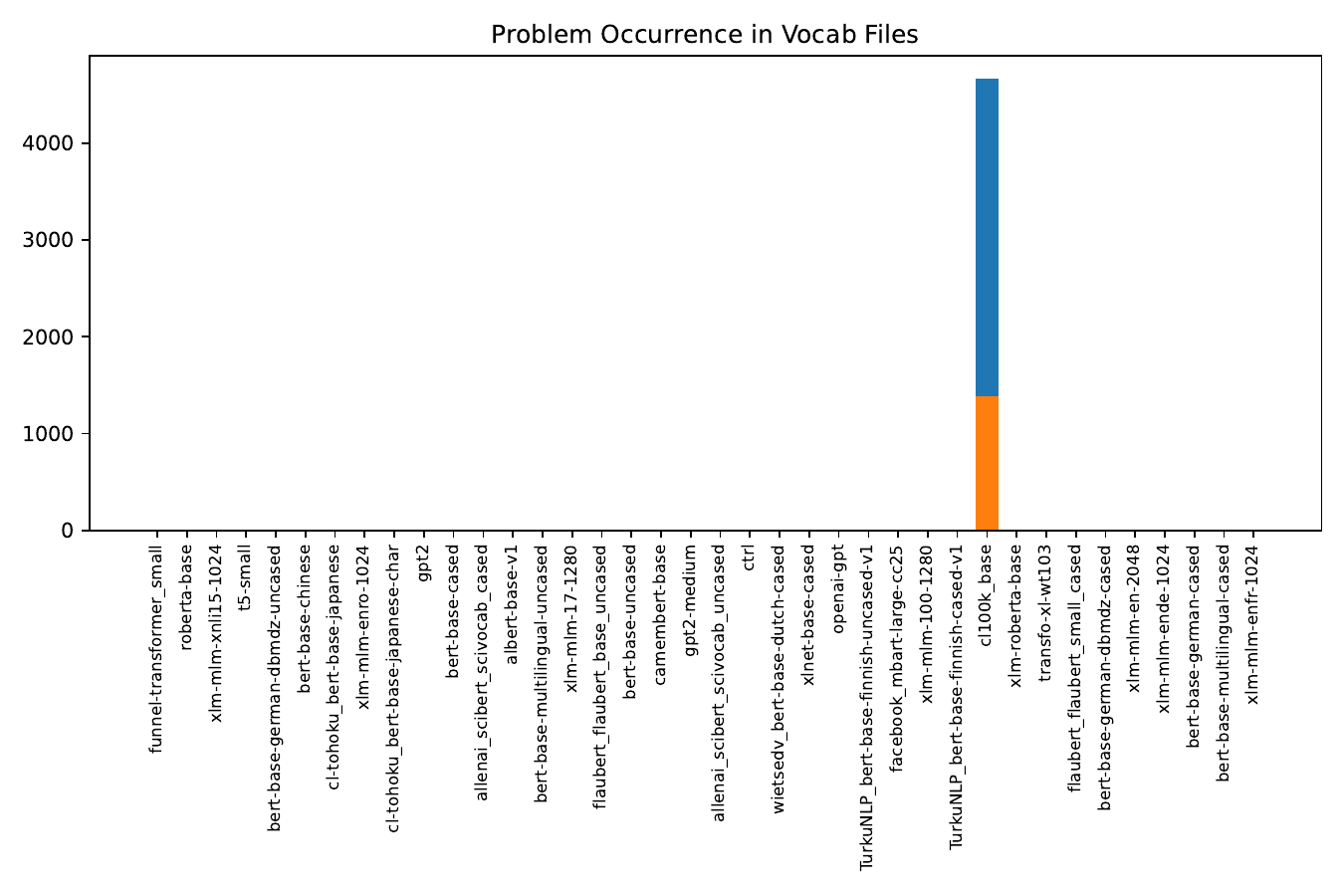}
    \includegraphics[width=0.5\textwidth]{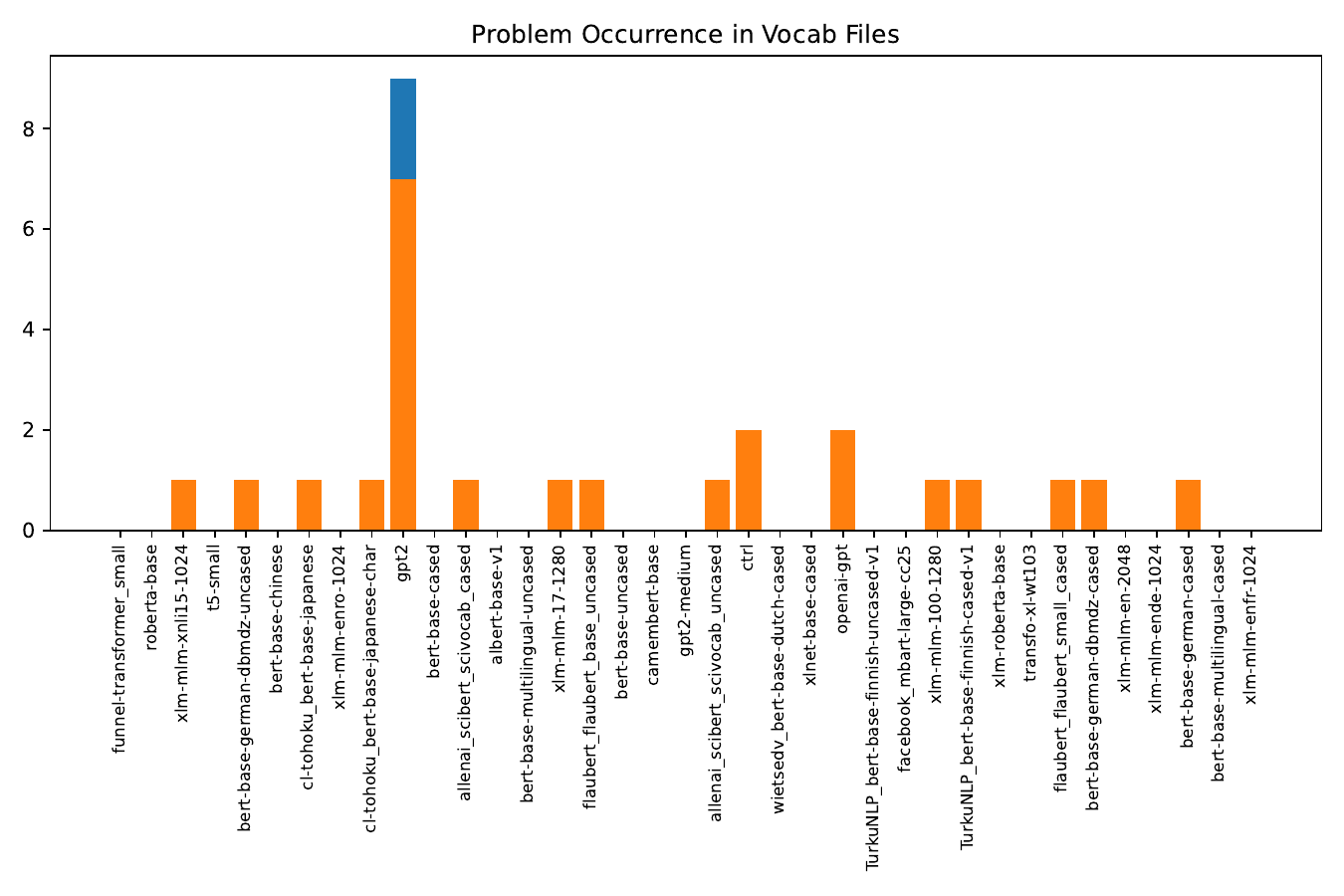}
    \caption{These charts shows the number of lines that caused problems when reading the initial vocabulary files from hugging face. The second chart shows the problems with the file cl100k\_base excluded. The orange height shows the number of lines that seemed to be unique; the blue height shows the total number of lines that caused an exception when being parsed the standard way we handled all lines.}
    \label{fig:problems}
\end{figure}

4,696 lines from the 38 vocabulary files caused problems -- were not parseable in the same way as the other lines -- which boils down to 1,413 unique lines. We did not use the tokens contained in these lines in any of the analysis. These almost all come from the cl100k\_base (4,670 lines, 1,389 unique lines) vocabulary file. (See Figure~\ref{fig:problems}.)

Around May 2024, we further analyzed those vocabulary maps, and did some initial exploration with the vocabulary of the newly-released GPT-4o model, which we accessed as tiktoken's ``o200k\_base''. There are six vocabularies available from tiktoken (gpt2, r50k\_base, p50k\_base, p50k\_edit, cl100k\_base, and o200k\_base)~\cite{tiktoken2023}.
~\footnote{Perhaps surprisingly, when we looked at four of the vocabulary maps produced via tiktoken, three (gpt2, p50k\_base, and r50k\_base) were identical. There could be differences in how the tokenizer breaks input sequences down into vocabulary components, or some other reason not visible in the vocabulary itself as to why these would be considered different vocabularies.}

We note that the GPT-2 vocabulary maps produced by Hugging Face GPT-2 tokenizers (gpt2-medium, gpt2-large, gpt2-xl), while all identical, are different from the GPT-2 map produced by tiktoken. In general, some of the tokenizers, model weights, or other parameters available through Hugging Face may be from different versions of the relevant models (than what is accessible through tiktoken). From our perspective, this was indiscernible.

The two GPT-2 vocabulary files have 41,396 tokens in common out of 41,409 tiktoken tokens and 41,403 Hugging Face tokens (the number in each vocabulary if appending whitespace is ignored). The tokens that apparently differ do not look relevant to this paper. When either vocabulary is checked against all 965 bad words (word lists described below, Sec.~\ref{sec:aboutwordlists}), the same 90 words are flagged from each file.

There are 200,000 tokens in GPT-4o's vocabulary (via tiktoken). We tried to keep the cleaning process as close as possible to what we did with the 38 other files (via Hugging Face), although we obtained it via a different pathway. The token list from tiktoken has 200,000 entries. However, when saved with one token per line to a text file and re-opened, it has 203,308 entries. To avoid worrying about this, we used the list directly from tiktoken without the intermediate step of having saved and opened it again. However, 198,797 tokens come up as unique out of the 200,000 entries, even when loaded directly from tiktoken. We think this is encoding-related, since the most prevalent current tokenizations (like BPE) are generally applied at the byte level, rather than the character level, meaning they may be resolved to the same string even when they are not identical at the byte level.

Cleaning non-alphabet (including space) characters, and eliminating tokens that are the empty string after cleaning, leaves 91,206 unique tokens. Lower-casing (removing case) leaves 72,151 unique tokens. That means that, of the clean tokens, \num{0.20892265859702214} (19,055) were cased variations of each other. Of the clean but not lower-cased tokens, \num{0.21072078591320748} (19,219) have an initial capital letter but no later capitals, meaning they could be typically-formatted proper nouns (personal names, toponyms).

For the initial straightforward comparisons among the tokens, we stripped whitespace and leading \_ from tokens (used to indicate leading whitespace), but we did not strip other special symbols. Some vocabulary files included things like trailing w or leading \#\# (presumably indicating something about a word boundary).\footnote{E.g. \#\#rook, \#\#take, and \#\#washed are examples of tokens from funnel-transformer\_small.} For more detailed comparisons, we cleaned the files more aggressively, stripping from tokens anything that was not a space or in the Latin alphabet (upper- or lower-case).

\begin{figure}
    \centering
    \includegraphics[width=0.5\textwidth]{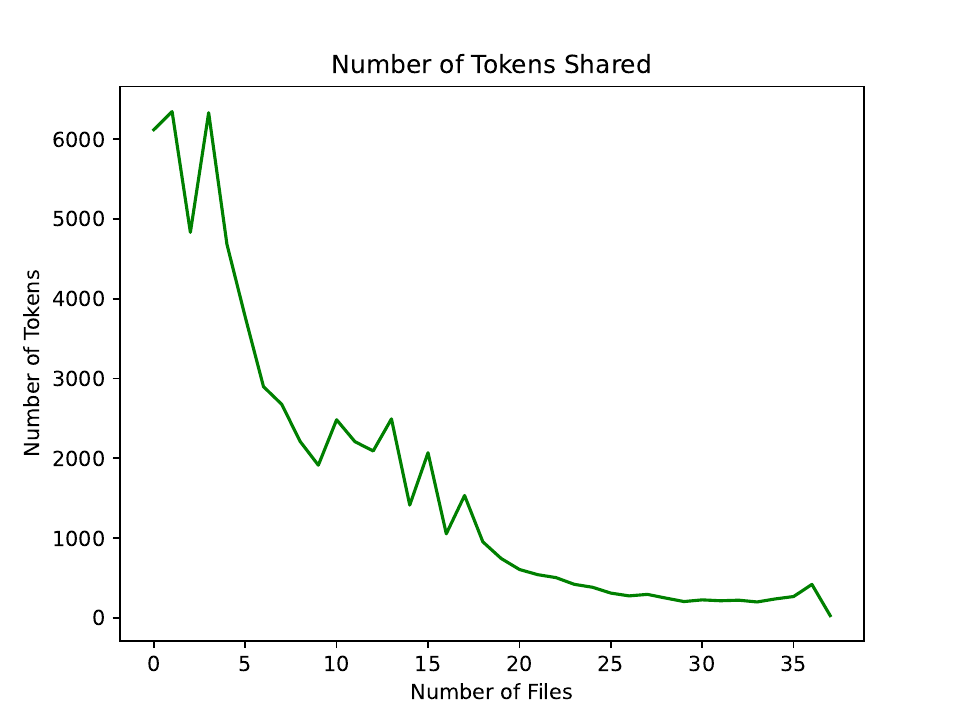}
    \includegraphics[width=0.5\textwidth]{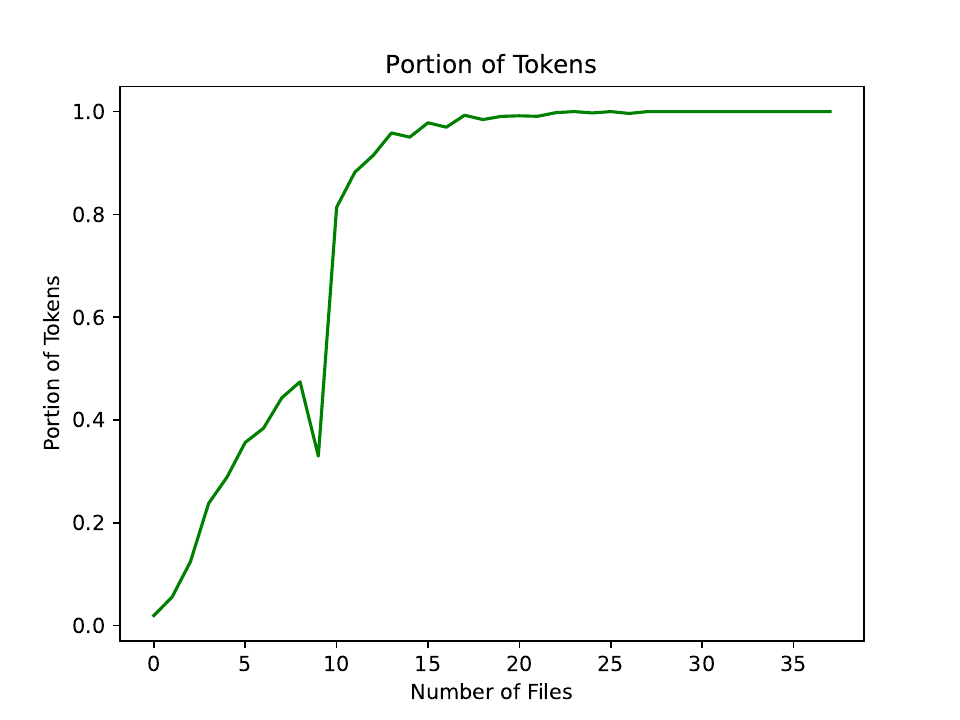}
    \caption{The first chart shows the number of clean tokens shared between GPT-4o and that number of files.  The second chart shows the portion of clean tokens in GPT-4o and the Hugging Face vocabulary files, showing that if a token is in a good number of vocabulary files, it is likely in GPT-4o's vocabulary as well.}
    \label{fig:numtokenssharedgpt4o}
\end{figure}

\begin{figure}
    \centering
    \includegraphics[width=0.5\columnwidth]{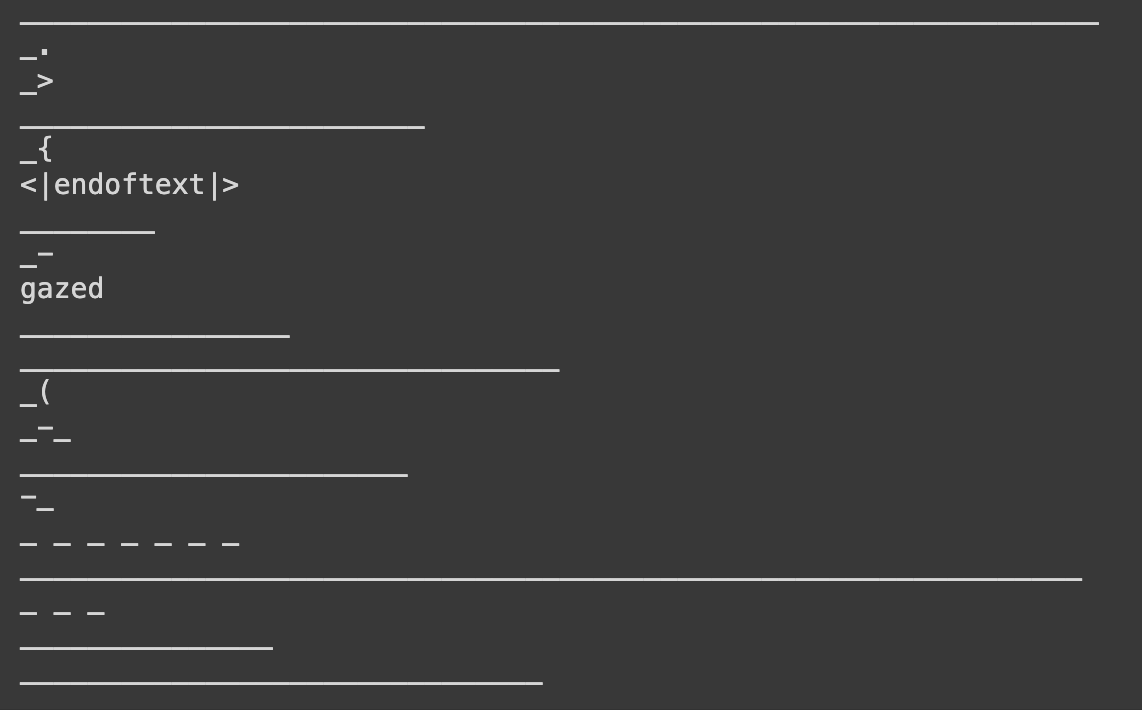}
    \caption{We compared the vocabulary file `gpt2' from tiktoken and from Hugging Face, and found some differences which we deemed minor. After stripping whitespace and eliminating resulting duplicates, there were about 19 differing tokens. Examples are shown here. These differences could reflect genuine differences in the tokens returned from the two sources, maybe due to the version of the file each is drawing on, or they could be due to differences in formatting that caused us to view tokens as different that would not be different to a model using that vocabulary. That we did not deeply investigate these differences for this or any other file is a limitation of this paper. We decided for now it was reasonable to gloss over, since our results are more about what kinds of things become tokens in general than tied to one model or one tokenization process.}
    \label{fig:gpt2_h_vs_t_differences}
\end{figure}

\begin{figure}
    \centering
    \includegraphics[width=0.5\textwidth]{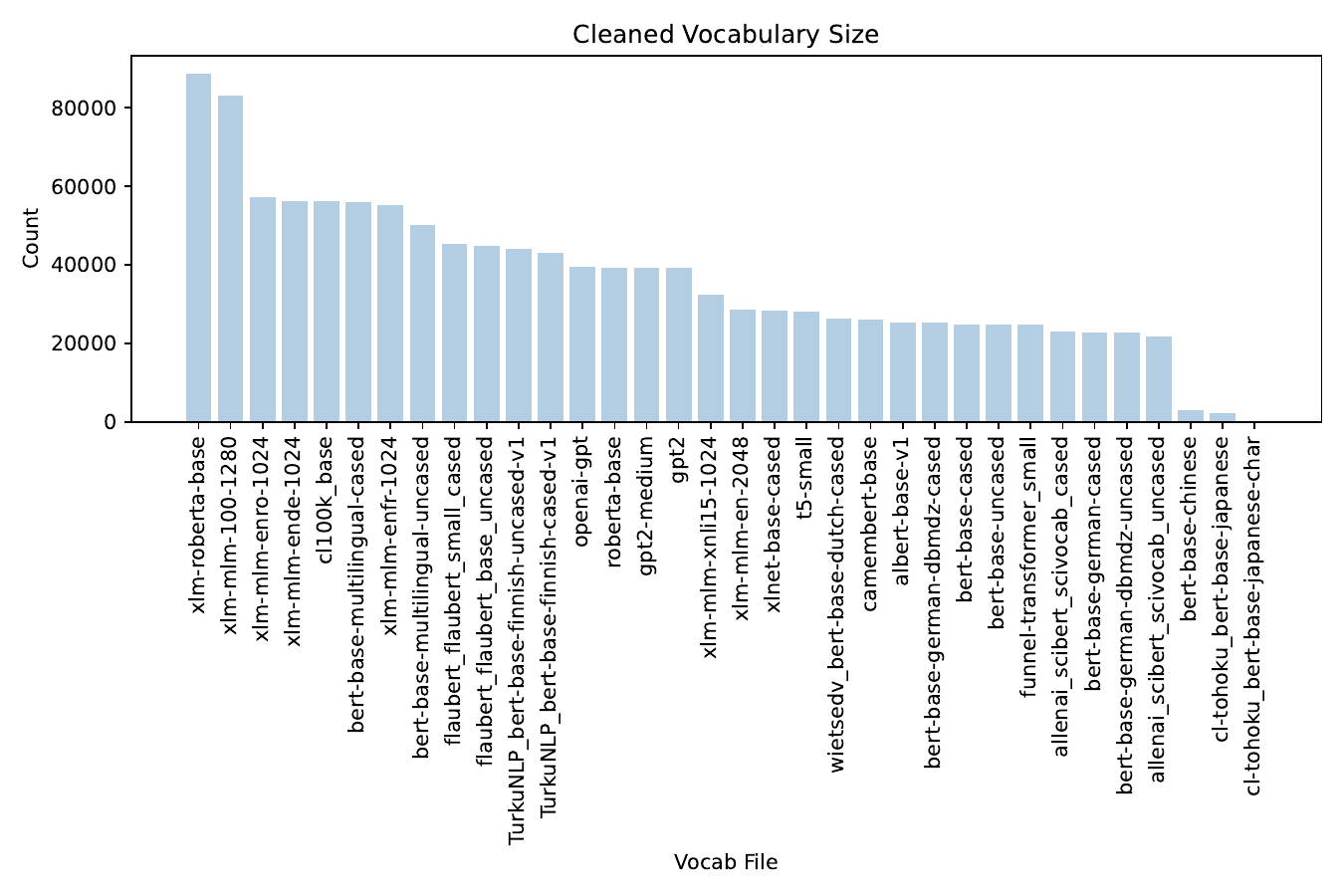}
    \caption{This figure shows the sizes of the vocabulary files after the tokens are aggressively cleaned and de-duped within each file (but retain case), for a total of 1,774,968 tokens.}
    \label{fig:cleaned_vocabsize_hf}
\end{figure}

\begin{figure}
    \centering
    \includegraphics[width=0.5\textwidth]{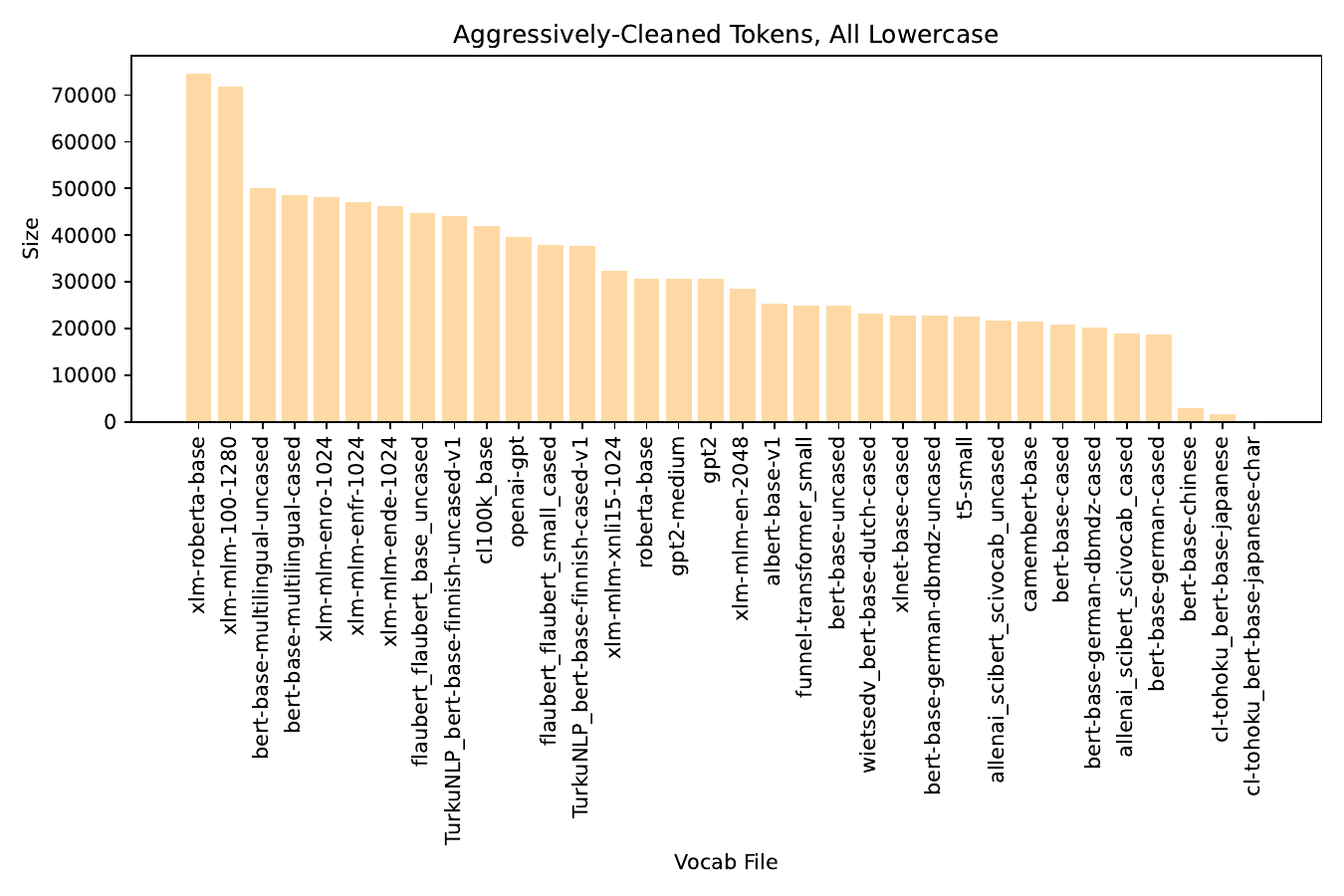}
    \caption{These are the sizes of the vocabulary files after the tokens are aggressively cleaned and de-duped within each file (with case removed), for a total of 1,534,479 tokens.}
    \label{fig:cleaned_vocabsize_lower_hf}
\end{figure}

\appendixsection{Word lists}
\label{sec:aboutwordlists}
This section is about the lists we used in making our categories.

The HarperCollins 2019 word list (CSW19)~\cite{CSW19} is a list of words that are allowable in the English-language version of Scrabble. As far as we can tell, CSW19 is the most recent \textit{longest} version of the list (words were removed in more recent versions), which makes it the most suitable for our purposes.\footnote{We did find a few small mistakes or typos, which we corrected in our version of the list.} Even so, it is not without caveats. Since originally sharing this preprint, we have learned that the parts of speech we applied from the CSW19 list are less straightforward than we initially thought. It includes very rare words and senses which we may have used in place of more typical parts of speech~\cite{fatsis2026jake}.

We used the S2 list~\cite{dwyl_english_words} because it is more generous in its inclusion of words and contains lengths outside of the range 2-15.

There are 279,496 entries in CSW19. Those words occur across the 38 vocabulary files we obtained from Hugging Face 421,490 times as 83,343 unique tokens (looking at clean, lower-cased versions of the tokens in the files). Their average length was \num{6.872312510379843}, and average unique length was \num{8.115162641133628}. There are 279,585 in CSW19 combined with our list of affixes. There are 370,105 in the alphabet-only (clean) version of S2; the creators of S2 also created this clean version. There are 466,550 in the base version of S2. Combining down any repeated words, we get a list of 458,654 alphabet-only words, and a list of 585,006 more inclusive words. Adding in our list that combined CSW19 and affixes brings the total up to 458,685; there are only about 30 items in that list that were not already in the S2 word list.

We used this combined list (S2, CSW19, affixes) to roughly approximate morphemes (morphemes are the smallest semantically meaningful unit of language, including things like words and affixes). For convenience, we call the result ``morphemes'', but please note that is an extremely rough approximation. Adding in the affixes list to the cleaned words lists adds back non-alphabet characters, such as apostrophe + s ('s). Those entries will not match any of our cleaned tokens because we removed non-alphabet characters (o'clock becomes oclock, re-elect becomes reelect, etc.). This was a significant simplification and could be handled differently.

There are 458,685 entries in our resulting master list of rough morphemes (based on CSW19, S2, and affixes). This matches 99,841 unique tokens occurring 511,216 times across the 38 files. The average length is \num{6.5879178272980505}, and average unique length is \num{7.942678859386424}. This is approximately words, since most of the entries are words.

There are 86 morphemes in the two morpheme lists we found online~\cite{morphemelist,morphemelist2}, to which we added 10 custom ones, for 96 total (which are all affixes). There are 282 from Scrabble Australia (SA)~\cite{scrabble_prefixes}, for a total of 315 affixes. All 315 affixes are present in our master morpheme list. The 6 affixes listed explicitly in CSW19 are in the master morpheme list as well, but one is not in the list of 315 affixes: `as'. Including `as' yields a total of 316 affixes. For the 316 affixes, there are 9,840 occurrences among the clean, lower-cased tokens; 315 unique (since apostrophe + s, `'s', was removed during cleaning). The average length is \num{3.3578252032520326}, and average unique length is \num{3.4317460317460315}. Some of the most common of the 316 affixes and the number of files they occurred in are: `a' in 38 files, `y' in 38 files, `e' in 38 files, `s' in 38 files, `off' in 37 files, `ent' in 37 files, `men' in 37 files, `good' in 37 files, `key' in 37 files, and `over' in 37 files.

For some uses, we combined a list of affixes~\cite{morphemelist} with a description of inflectional morphemes~\cite{morphemelist2} and our own judgment, which we then augmented with the same fuzzy matching approach that we used for our small list of bad words.

We used the first sample list of 5,000 ``lemmas'' from the \textit{wordfrequency.info} website for frequency information~\cite{frequencylist}.

We used a list of 277 function words by James D. O'Shea, shared on his website~\cite{semanticsimilarity_function_words}. Based on his website's About page, he decided to share such lists after publishing a related paper~\cite{1644735semanticsimilarity}. Of the 277 function words, 274 are in at least one vocabulary file. Note that we expect there to be about 300 English function words~\cite{brysbaert2016words}, which is consistent with this list.

We used Winter \textit{at al.}, 2017's list of words and iconicity scores~\cite{winter2017bodo}.

Out of the 39,499 tokens that are written in all capital (Latin alphabet) letters and have a length greater than one (to exclude individual letters), 14,444 are unique.

Out of 2,909,850 slightly cleaned tokens we read in via the vocabulary files, 1,774,968 remain after we aggressively clean them by removing any non-Latin-alphabet characters and discarding any tokens that have no non-Latin-alphabet characters in them, and removing duplicates within each vocabulary file (but not across vocabulary files). 687,085 tokens are unique across the files (case-sensitive). If we de-duplicate by making everything lower-case, there are 569,810 total unique tokens from all 38 files combined. The nltk~\cite{bird2009nltk} package can at least potentially handle case, so where possible we tried to err on the side of cleaning as little as possible.
Out of 284,133 words with a part-of-speech in CSW19, 86,889 are also tokens in the combination of the 569,810 aggressively cleaned and de-duplicated tokens from the 38 vocabulary files.

A comparison of the word lists is shown in Figure~\ref{fig:comparison_of_word_lists}.

\begin{table}
    \centering
    \small
\begin{tabular}{llrrrrr}
\toprule
 Name & Length & CSW19 & CSW19+ & S2 & S2+ \\
\midrule
CSW19 & 279496 & 1.000000 & 0.999682 & 0.515927 & 0.398978 \\
CSW19+ & 279585 & 1.000000 & 1.000000 & 0.516083 & 0.399074 \\
S2 & 370105 & 0.683183 & 0.683173 & 1.000000 & 0.721239 \\
S2+ & 466550 & 0.665995 & 0.665944 & 0.909185 & 1.000000 \\
\bottomrule
\end{tabular}
\caption{A comparison of the words in each list, CSW19, CSW19+ (our list of CSW19 plus some affixes), S2, and S2+ (words including  non-alphabet characters). The proportions are the number of words in both lists out of the number of words in the list represented by that column.}
\label{fig:comparison_of_word_lists}
\end{table}

We decided to combine two lists for our longest bad word list (we also looked for smaller sets of suspect words, on a case-by-case basis). We used a list of 403 words seemingly shared by Shutterstock, because it seemed to have fewer neutral words, while still maintaining a solid length. The ReadMe file states that they use the list ``to filter results from our autocomplete server and recommendation engine'' and that their guiding principle for terms in the list is to ask, ``What wouldn't we want to \textit{suggest} that people look at?''~\cite{shutterstocklist} We used a list of 958 words that claims to be the ``[f]ull list of bad words and top swear words banned by Google''~\cite{googlelist}, because it has had more recent sustained activity, and has more variants for specific words of interest. The lists have significant overlap: their union has 965 words.

We used WordNet, which could be categorized as a list or a tool. Note that there are ``few adverbs in WordNet (hardly, mostly, really, etc.) as the majority of English adverbs are straightforwardly derived from adjectives via morphological affixation (surprisingly, strangely, etc.)''~\cite{WordNet}. Since originally sharing this preprint, we have learned that the parts of speech we applied from WordNet are less straightforward than we initially thought. It includes very rare words and senses which we may have used in place of more typical parts of speech~\cite{princeton2026cntlist}.

\appendixsection{Tools}

We used the automated part-of-speech (POS) tool nltk~\cite{bird2009nltk} to label the tokens with very approximate parts of speech. POS information is more detailed in nltk than in CSW19. According to these labels, the majority of tokens are nouns (by far the largest category), and most of the tokens in the noun category were categorized as singular (e.g. ``cat'') or mass (e.g. ``sand'') nouns. If we roughly aggregate related POS groups together, then verbs are the next most prevalent category, followed by adjectives, adverbs, prepositions, pronouns, conjunctions, numbers, symbols, foreign words, the ``to'' particle, interjections, and existential ``there'' (the last three categories are of tokens parsed as multiple POS; this could reflect tokens that include an interior word boundary, in other words, that are part of a phrase -- it is worth noting they are tiny categories with only a few instances).

The resulting POS tags are only rough guesses. For example, in the vocabulary file xlm-mlm-enfr-1024, the token SING is categorized by nltk as a noun (NN), as is LUMB, even though the first looks more likely to be a verb and the latter looks like part of a word. It looks like nltk generously assigns noun with little context, especially for strings made entirely of capital letters, which is fair enough, as they could be abbreviations standing for entities (acronyms, initialisms). It is not really fair of us to ask a tool to do part-of-speech tagging when there is no sentence structure (we passed in individual tokens) and there are chunks that are far from recognizable words or phrases. Both stanza~\cite{stanza} and nltk output some obviously unhelpful labels under these stormy conditions.

For detailed comparisons among the tokens, including looking at parts of speech, we stripped from tokens anything that was not a space or in the Latin alphabet (upper or lower case). This means any characters from other languages were ignored. The 0 POS category indicates nltk did not provide any tag for that token. Many of those tokens look to be empty strings or whitespace after the aggressive, anglo-centric cleaning process.

We used the spaCy and benepar package pipeline for syntactic constituency parsing~\cite{honnibal2020spacy,kitaev-klein-2018-constituency,kitaev-etal-2019-multilingual}. We used tree-viewer to visualize syntax trees~\cite{treeviewer2023}.

\appendixsection{Parts of speech and human vocabularies}
\label{sec:partsofspeechappendix}

Although it is not straightforward to obtain the precise proportion of parts of speech in English (as a whole or in individual vocabularies)~\cite{brysbaert2016words}, we used WordNet, nltk, and CSW19 as basic starting points for comparison. We can also loosely compare the ratios we found to an estimate of parts of speech in terms of frequencies in corpora (tokens, as in types-tokens), which puts nouns as the largest category in written English at around 37\%, and which puts verbs around 18\%, prepositions around 12\%, adjectives around 7\% and adverbs around 5\%~\cite{hudson1994about}.

Similarly to the large uncertainty in relative portions of parts of speech, estimates of adult (human) English speakers' vocabularies vary pretty widely. However, at the small end is 9,000 word families (e.g. ``run'', ``running'', ``ran'', etc. would all be counted as 1 unit), with other researchers suggesting figures between about 17,000 and 200,000. Data from a popular quiz-style website suggested adult English speakers may know 20,000 - 35,000 words, and kids around age 8 may know 10,000 words~\cite{treffers-daller2013vocabulary, economist_lexicalfacts}.

By comparison to WordNet, CSW19 adheres more strictly to surface forms (which makes sense given its purpose). There are 279,496 words (some of which are derived from each other). Nouns are the largest category, at about 57.72\%, verbs at 24.53\%, adjectives at 14.50\%, adverbs at 3.06\%, then (in order) interjections, prepositions, pronouns, and conjunctions, all below 1\%. Out of 90,686 affixes listed explicitly in CSW19, ``s'' occurred most frequently, at about 76.71\% of the time, followed by ``ed'' at 8.16\%, ``ing'' at 8.15\%, ``es'' at 6.97\%, and ``e'' and ``as'' occurring just once (0.0011\%). All 90,686 words that can be generated from the CSW19 list using the affixes given therein are already among the 279,496 words in the list. Removing those words yields 189,558 words that are more independent of each other than the original list (the base words). From there, the approximate portion of nouns is 48.95\%; verbs, 25.17\%; adjectives, 21.13\%; adverbs, 4.48\%; followed by interjections, prepositions, pronouns, and conjunctions at less than 1\% each.

The reason we looked at both nltk and CSW19 for POS is that, in our opinion: (1) CSW19 has more reliable POS information than nltk, (2) CSW19 has a reliable list of words to compare to, (3) nltk has more detailed POS labels than CSW19, and (4) nltk can attempt to parse compound tokens.

\begin{figure}
    \centering
    \includegraphics[width=0.5\textwidth]{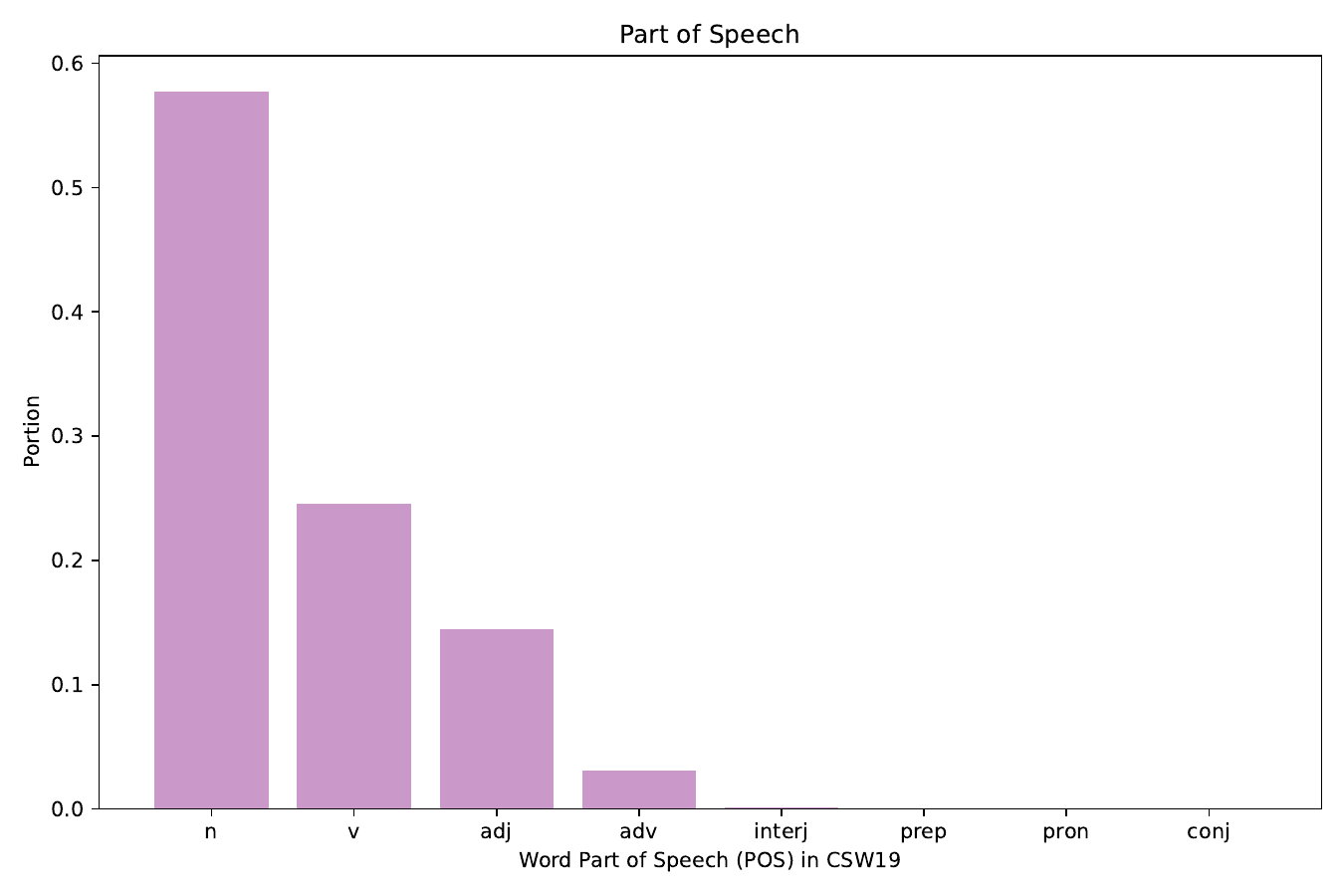}
    \includegraphics[width=0.5\textwidth]{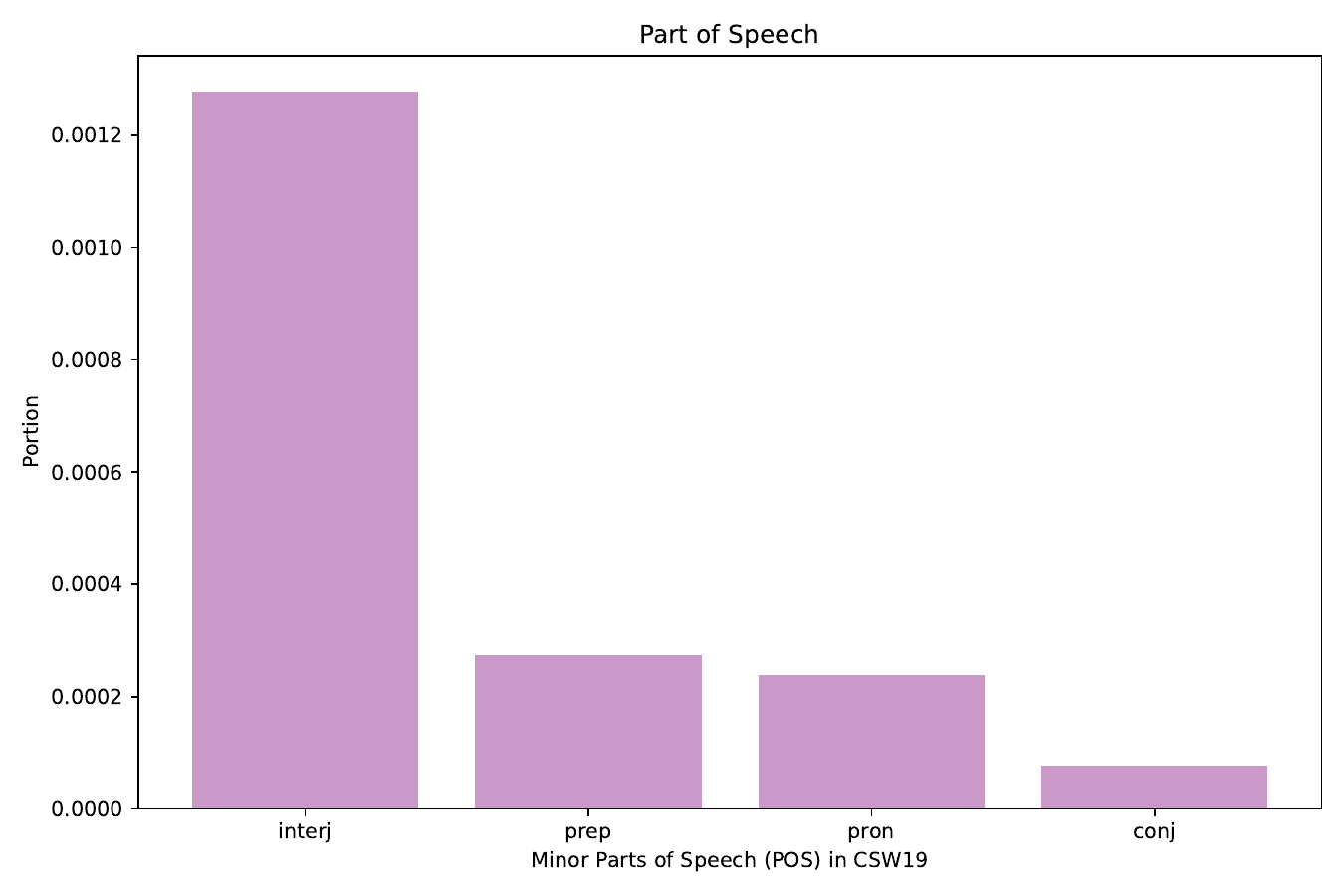}
    \caption{These charts shows the proportions of parts of speech among the words in CSW19. The second chart shows just the minor categories, each less than 1\%.}
    \label{fig:wordlength_csw19_2}
\end{figure}

\begin{figure}
    \centering
    \includegraphics[width=0.5\textwidth]{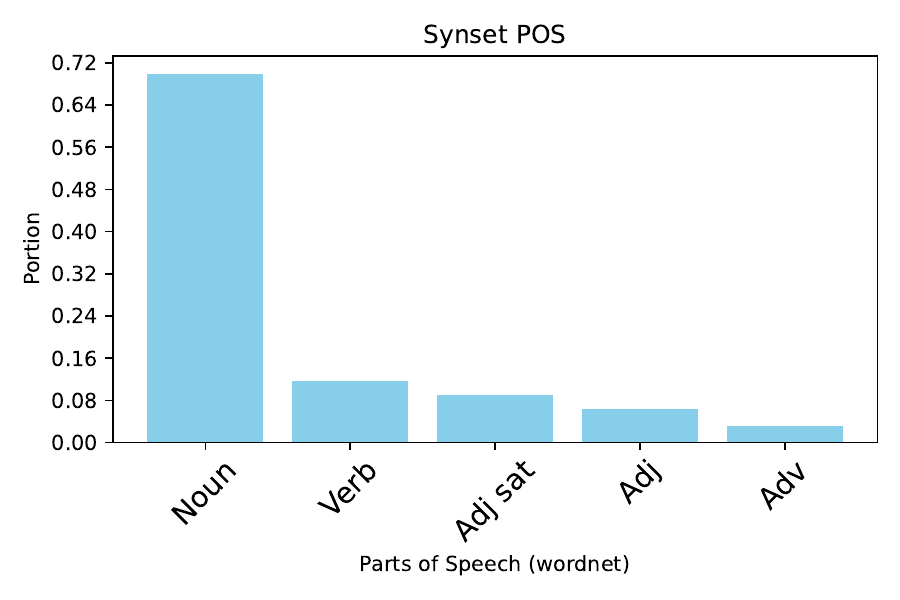}
    \caption{This chart shows the parts of speech across the synsets in WordNet. Note that ``adjective satellite'' seems to be a WordNet-specific term. In WordNet, ``[n]ouns, verbs, adjectives and adverbs are grouped into sets of cognitive synonyms (synsets), each expressing a distinct concept.''~\cite{WordNet} We looked at the synsets.}
    \label{fig:WordNet}
\end{figure}

\begin{figure}
    \centering
    \includegraphics[width=0.5\textwidth]{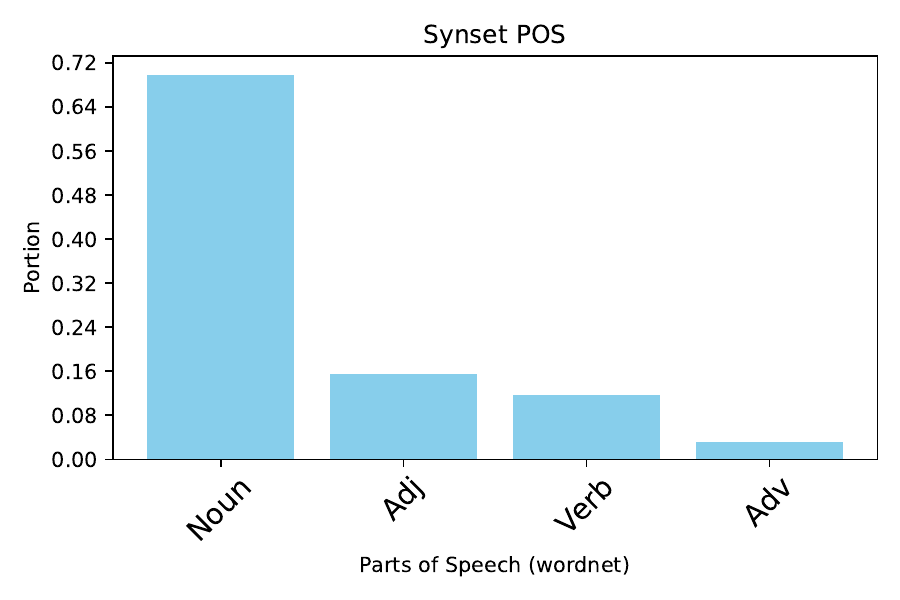}
    \caption{Although it is not straightforward to estimate the proportion of parts of speech in the English language, since WordNet is somewhat akin to word families, it seems like a good starting point of comparison. This chart shows the parts of speech across the synsets in WordNet, with the ``adjective satellite'' category subsumed by the ``adjective'' category. In WordNet, ``[n]ouns, verbs, adjectives and adverbs are grouped into sets of cognitive synonyms (synsets), each expressing a distinct concept.''~\cite{WordNet} We looked at the synsets.}
    \label{fig:WordNet_consolidated}
\end{figure}

Figure~\ref{fig:wordlength_csw19_2} shows the proportions of parts of speech among the words in CSW19. WordNet seems somewhat akin to word families (closer alignment to concepts than to surface forms). Figures~\ref{fig:WordNet} and ~\ref{fig:WordNet_consolidated} show the parts of speech across the synsets in WordNet.

\begin{figure}
    \centering
    \includegraphics[width=0.5\textwidth]{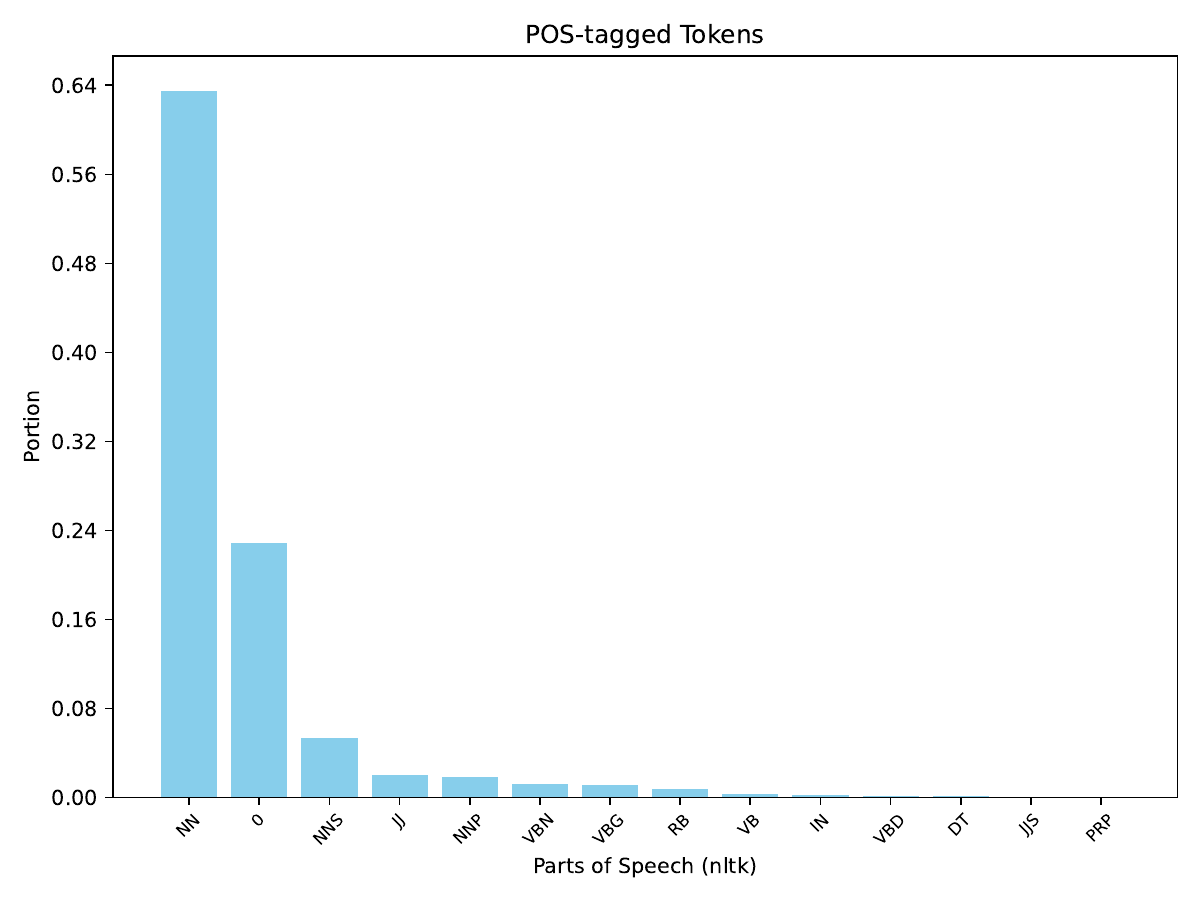}
    \caption{We used nltk to tag 2,909,850 tokens from 38 vocabulary files with parts of speech. The 0 category indicates nltk did not provide any tag for that token. We cleaned the tokens more aggressively for the tagging process. These are the parts of speech that occurred at least 1,000 times. The portion is calculated from the total number of tokens, not just the ones that were tagged as these parts of speech.}
    \label{fig:nltk_pos_portion_1000}
\end{figure}

\begin{figure}
    \centering
    \includegraphics[width=0.5\textwidth]{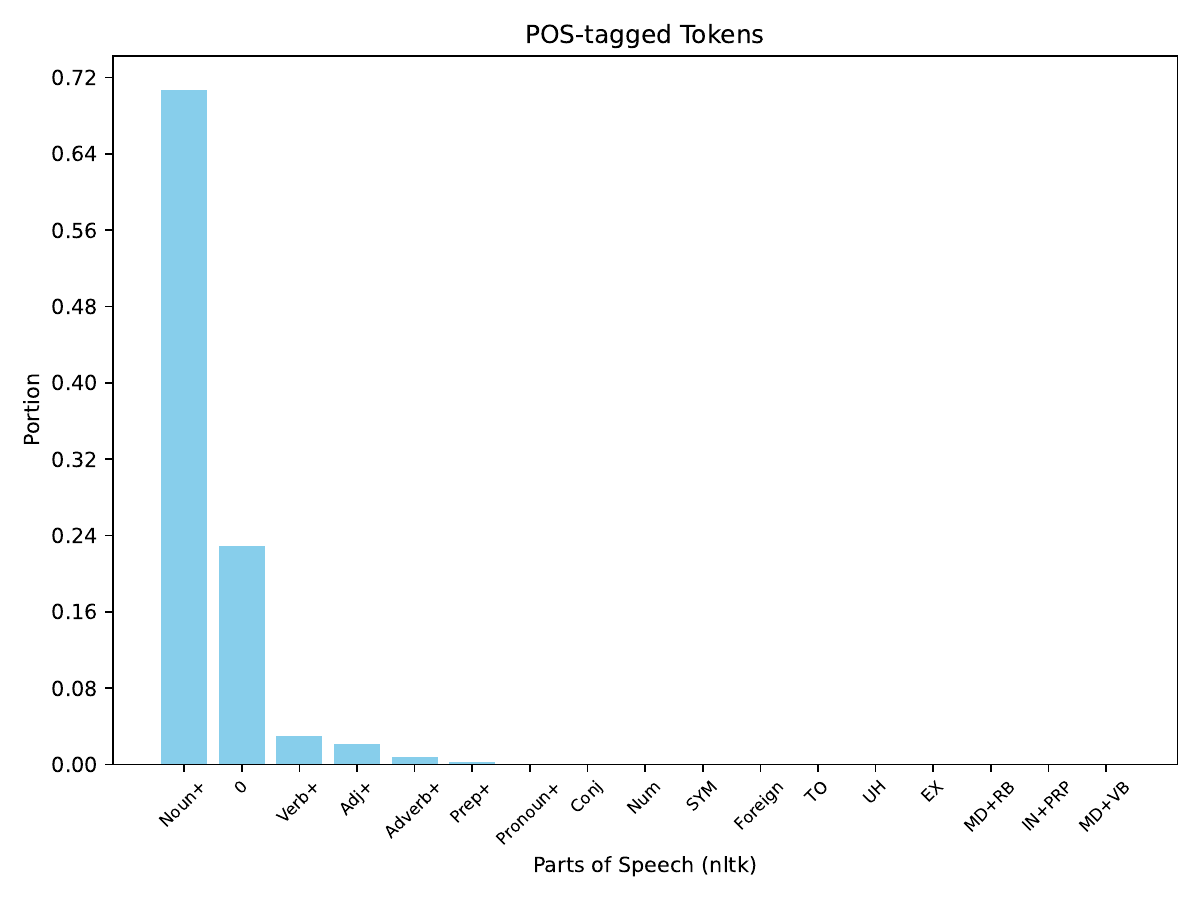}
    \caption{We used nltk to tag 2,909,850 tokens from 38 vocabulary files with parts of speech. The 0 category indicates nltk did not provide any tag for that token. We cleaned the tokens more aggressively for the tagging process. We then roughly consolidated the labels for the tags. All the resulting categories are shown in this visualization.}
    \label{fig:nltk_pos_portion_simplified}
\end{figure}

Figures ~\ref{fig:nltk_pos_portion_simplified},~\ref{fig:nltk_pos_portion_1000} show the parts-of-speech of the tokens in the 38 files as tagged by nltk.

Out of the 86,889 tokens we could match to CSW19, nouns made up approximately \num{54.28}\%, verbs \num{29.45}\%, adjectives \num{13.05}\%, adverbs \num{2.89}\%, and interjections, prepositions, pronouns, and conjunctions less than 1\% each.

\begin{figure}
    \centering
    \includegraphics[width=0.5\textwidth]{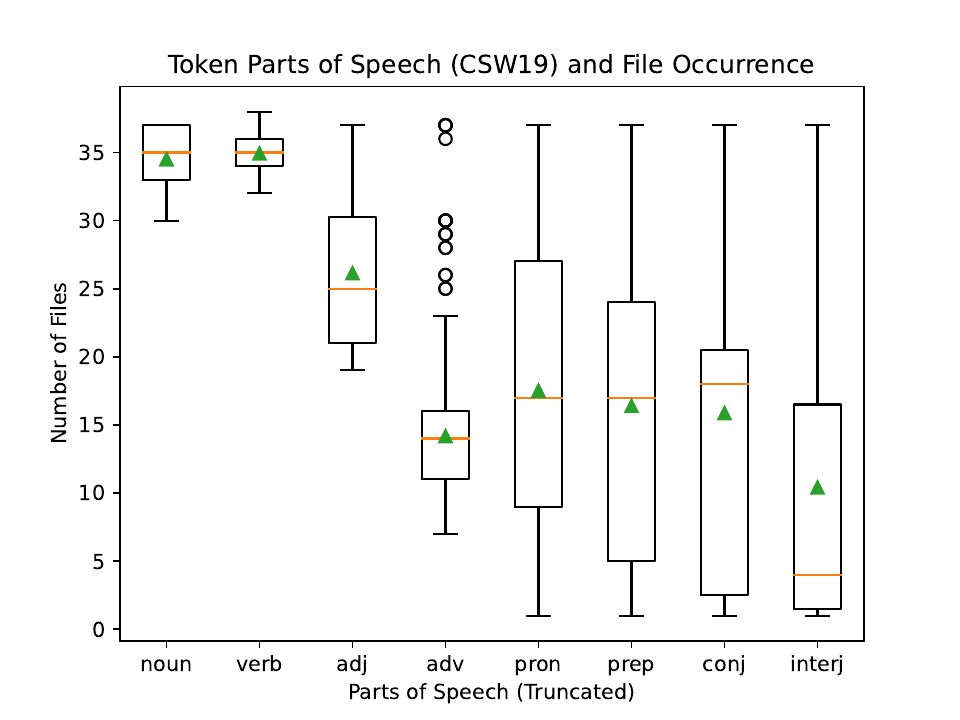}
    \caption{The spread of the number of files each clean token that had a POS in CSW19 occurred in, by POS, if we imagine each category of POS were about the same size, that is, looking at only (as many as) the 500 most common tokens in each of those categories. The green triangles indicate the mean.}
    \label{fig:pos_tokens_csw19_file_num_trunc}
\end{figure}

Because there are rare words in the open classes (content words) that cause a heavy tail, we truncated the lists of tokens to see what the results would be in that case. See Figure~\ref{fig:pos_tokens_csw19_file_num_trunc}.

We found that the open classes have heavy tails, and the function words do not. For the most part, the bigger the class, the heavier the tail, because the fact that a class is open leads to its being larger. One exception is interjections, which are a small, low frequency, but presumably at least semi-open class. Out of the smaller classes, interjections have the most similar shape to the larger classes, which is consistent with them perhaps being a semi-open class. We wonder if the shape of the prepositions is related to their having attributes in common with function words and content words (we can imagine combining the shapes of conjunctions and the heavy tail of nouns and getting the shape of prepositions). Prepositions may have more traditionally semantic content than other parts of speech under the function word umbrella, something our token extispicy methodology could perhaps shed light on in future work.

\begin{figure}
    \centering
    \includegraphics[width=0.5\textwidth]{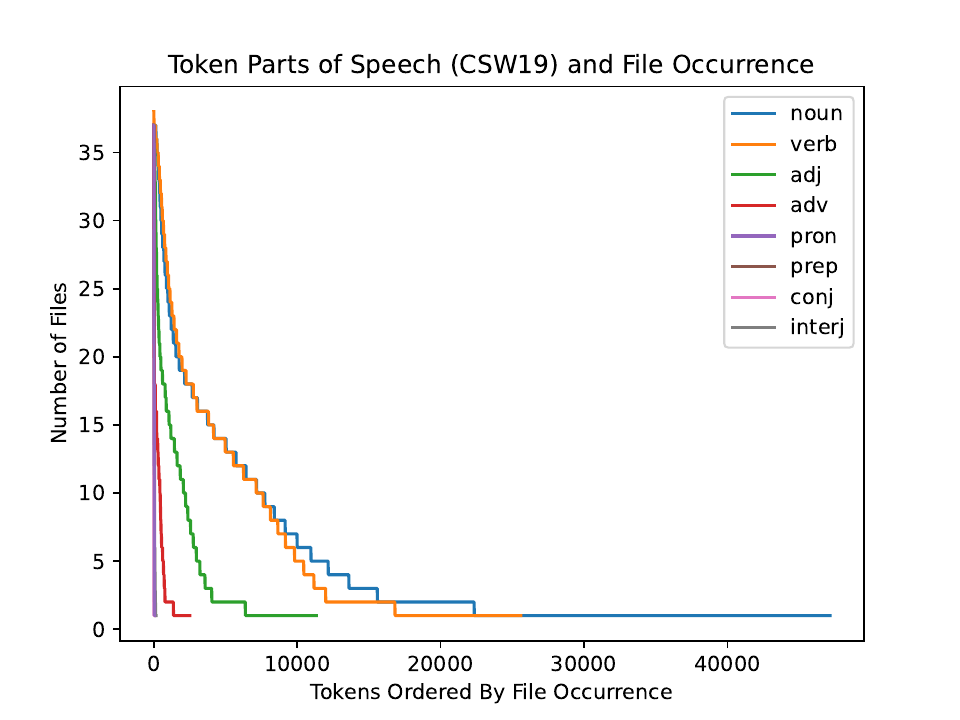}
    \caption{The number of files each clean token that had a POS in CSW19 occurred in, by POS.}
    \label{fig:pos_tokens_csw19_file_num_lines}
\end{figure}

\begin{figure}
    \centering
    \includegraphics[width=0.5\textwidth]{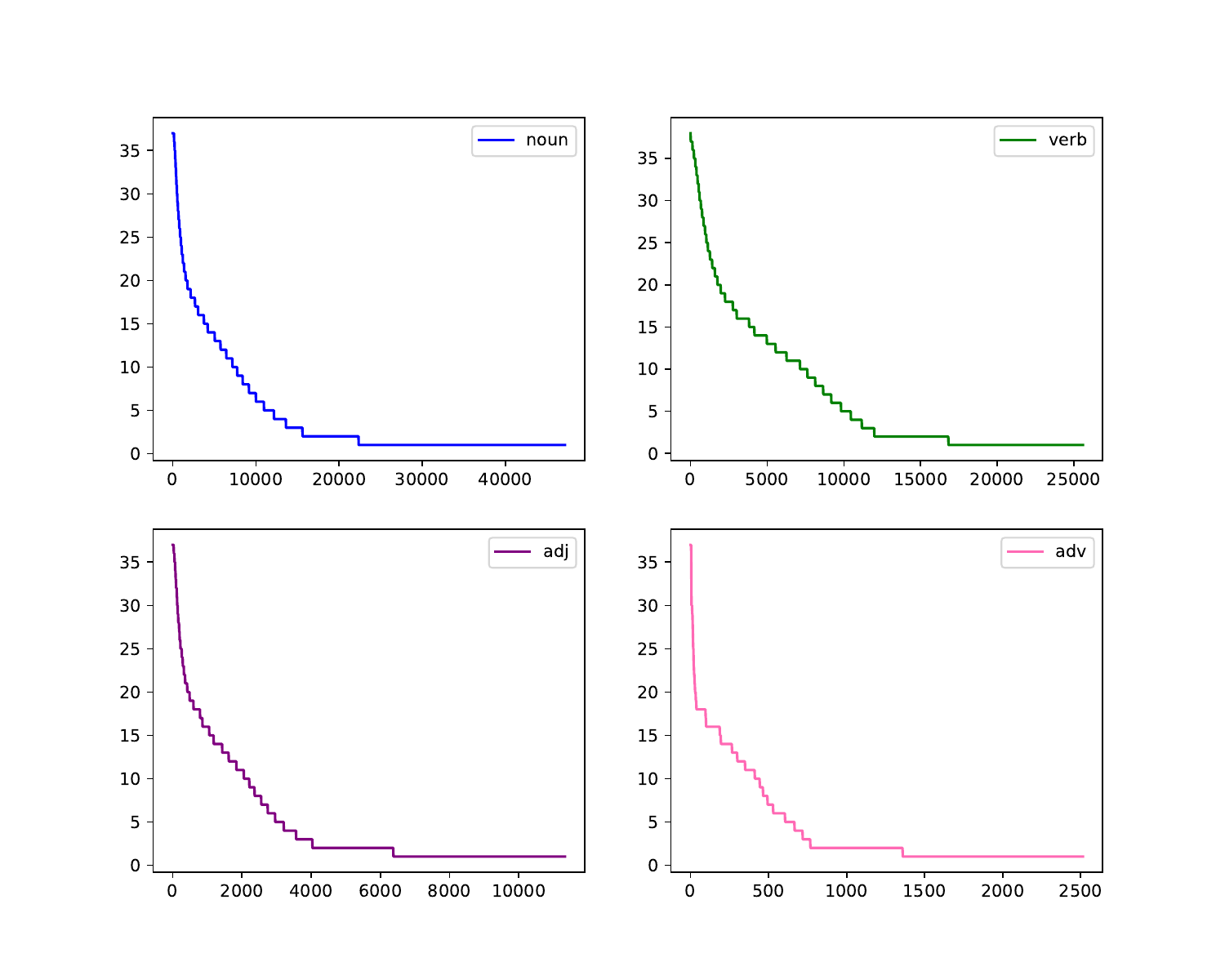}
    \includegraphics[width=0.5\textwidth]{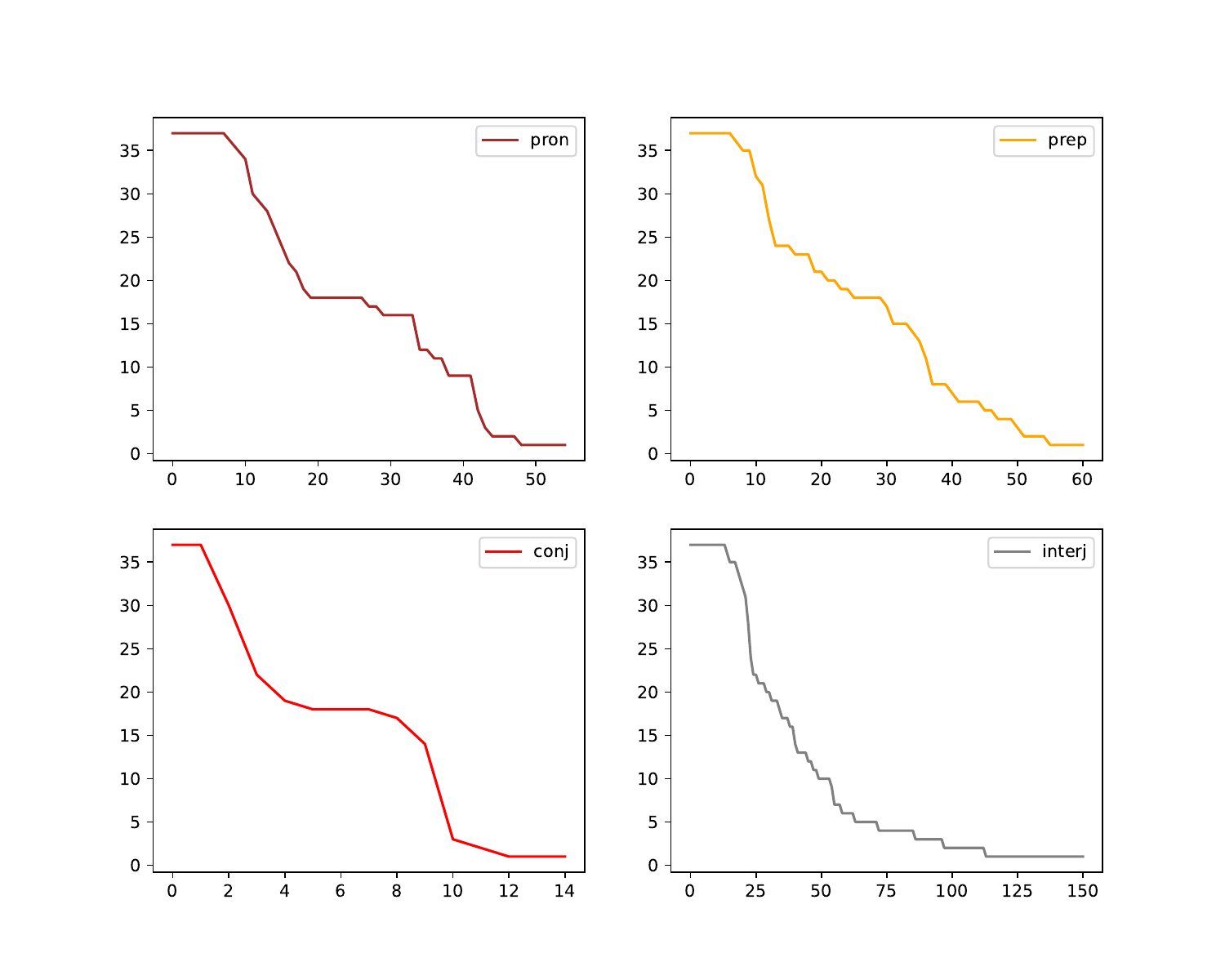}
    \caption{The number of files each clean token that had a POS in CSW19 occurred in, by POS. The x axis shows the total number of tokens in that category.}
    \label{fig:pos_tokens_csw19_file_num_lines_subplots}
\end{figure}

\begin{figure}
    \centering
    \includegraphics[width=0.5\textwidth]{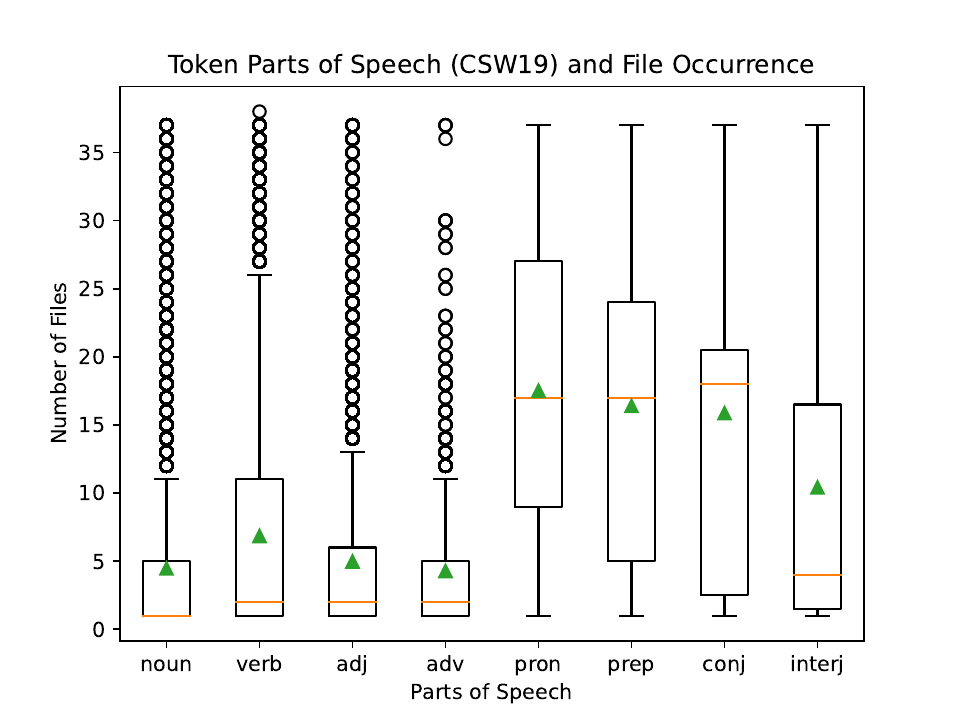}
    \caption{The spread of the number of files each clean token that had a POS in CSW19 occurred in, by POS. The green triangles indicate the mean.}
    \label{fig:pos_tokens_csw19_file_num}
\end{figure}

\begin{figure}
    \centering
    \includegraphics[width=0.5\textwidth]{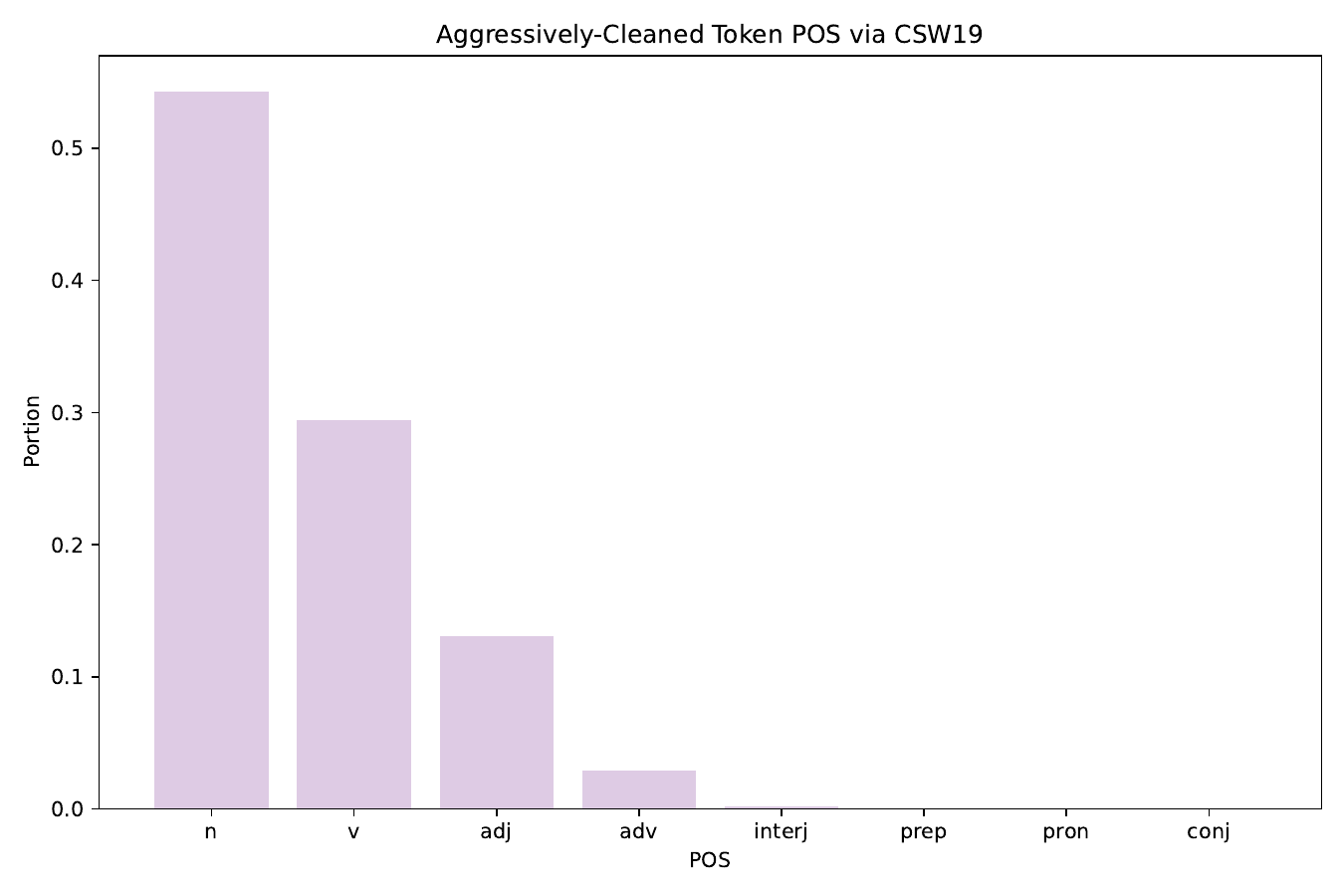}
    \includegraphics[width=0.5\columnwidth]{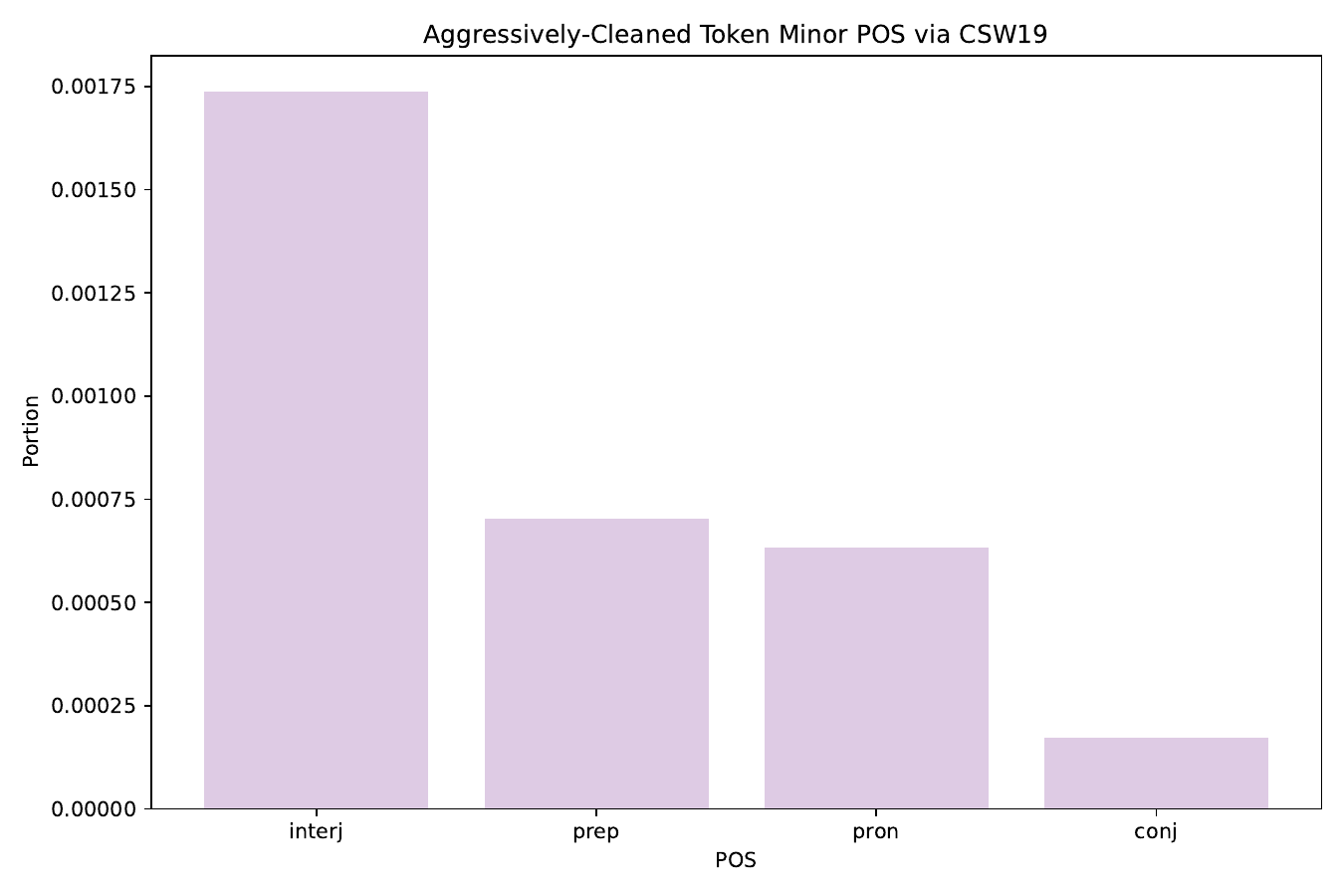}
    \caption{Out of the 86,889 tokens we could match to CSW19, nouns made up approximately 54.28\%, verbs 29.45\%, adjectives 13.05\%, adverbs 2.89\%, and interjections, prepositions, pronouns, and conjunctions less than 1\% each.}
    \label{fig:cleaned_pos_hf}
\end{figure}

\appendixsection{Token embeddings (extispicy) methods}
\label{sec:embeddings_methods}
RoBERTa (large) is an encoder-type transformer trained using the masked language model (MLM) objective 
originally pioneered in BERT~\cite{liu2019robertarobustlyoptimizedbert,devlin2019bertpretrainingdeepbidirectional}. It has 24 neural net/ attention blocks in its architecture, leading to 25 distinct embedding matrix states: the initial embedding, plus the output after each neural net/ attention block layer.

In a masked language model (MLM) such as RoBERTa, sequences of tokens are fed into the model, and the model is trained to output the same sequence, even when the input data is heavily corrupted (to fill in the gap of the masked token). RoBERTa's tokens are represented as 1024-length vectors. As the vector propagates through the model, information is added based on the context (the other tokens that are simultaneously propagated through). The value of a token's vector after any particular layer is highly dependent on the other tokens in the sequence. That is why we refer to our gnogeographic maps in the token extispicy section as being more inclusive of the model's knowledge: the information in those maps would not match the information in the initial embedding matrix.

For a few exemplar words, notably ``bank'' as it corresponds to the token `` bank'' (space + bank), we recorded the paths these token vectors took as they moved through the layers of the model for many different input contexts (sentences from our corpus which the token occurred in). A single trajectory can be described by a matrix with dimensions of 1024 (the length of the token vector) x 25 (the number of distinct embedding matrix states in the model; \textit{n} + 1, where \textit{n} is the number of neural net/ attention blocks). This is time-series data, as each successive layer's output is calculated sequentially, based on the previous layer's output.

\begin{figure*}
\centering
\includegraphics[width=\textwidth]{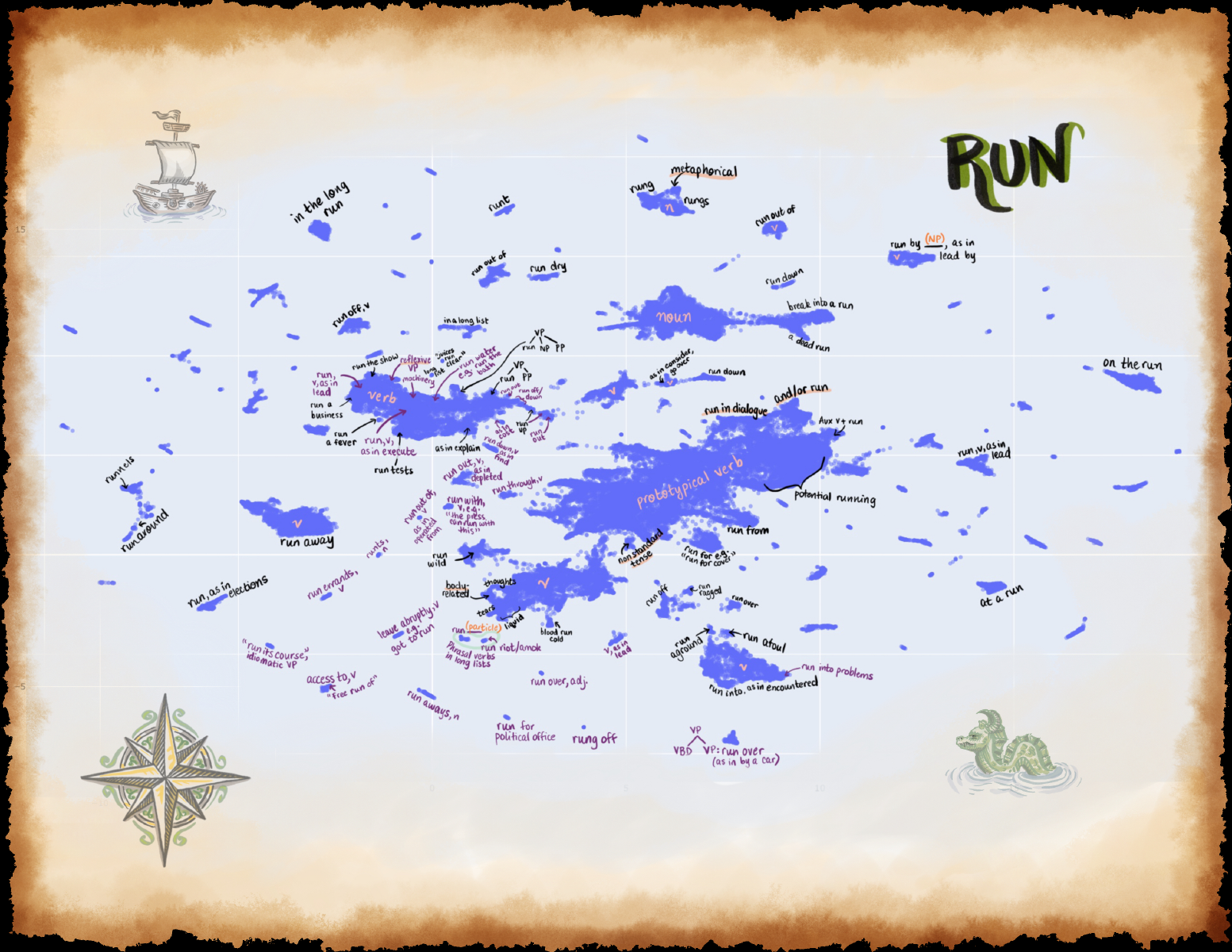}
\caption{A condensed visualization of the content within instances of the `` run'' token as it traveled through the RoBERTa MLM. We condensed the trajectories using PCA, then UMAP, to create the map. The compass rose is purely aesthetic!}
\label{fig:run}
\end{figure*}

\begin{figure*}
\centering
\includegraphics[width=\textwidth]{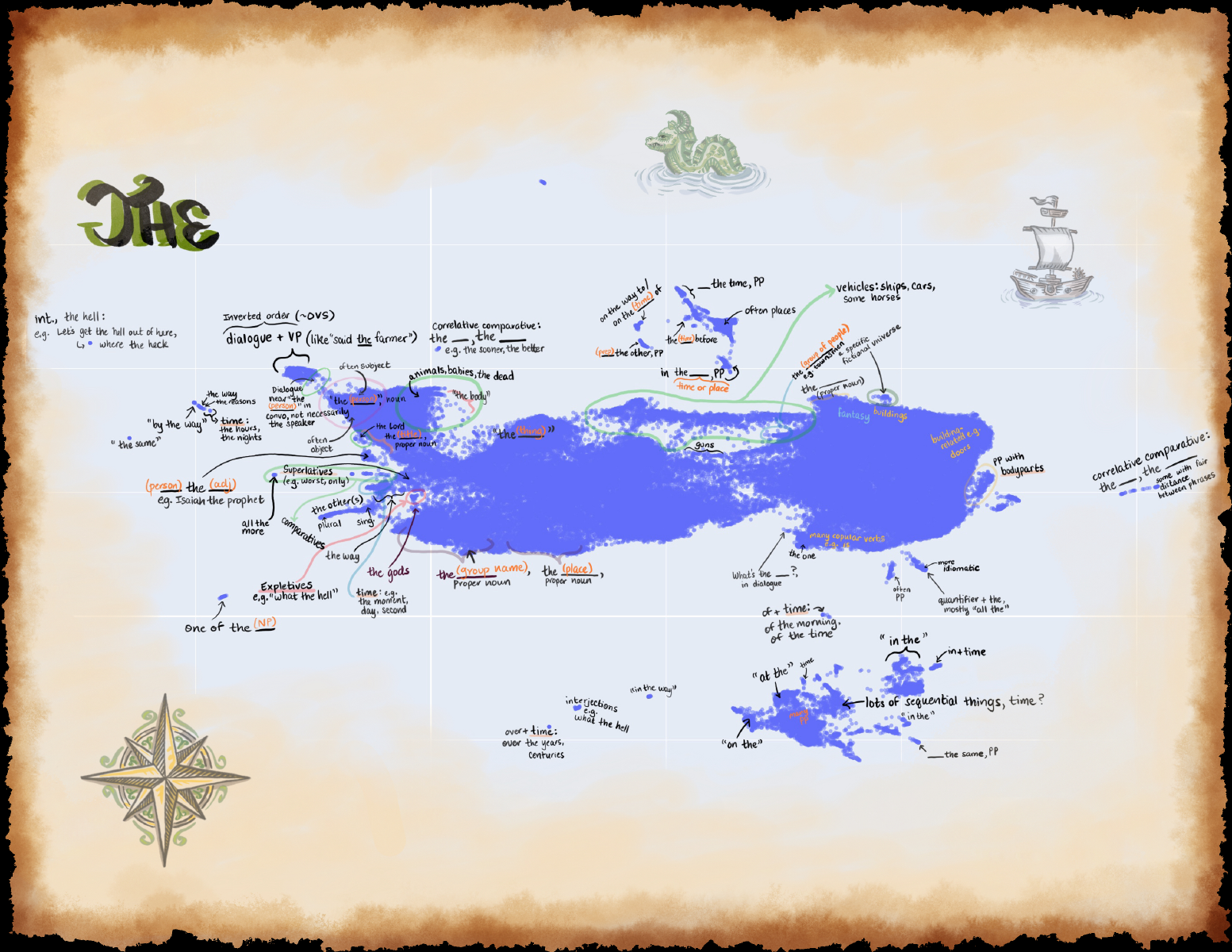}
\caption{A condensed visualization of the content within instances of the `` the'' token as it traveled through the RoBERTa MLM. We condensed the trajectories using PCA, then UMAP, to create the map. The compass rose is purely aesthetic!}
\label{fig:the}
\end{figure*}

\begin{figure*}
\centering
\includegraphics[width=\textwidth]{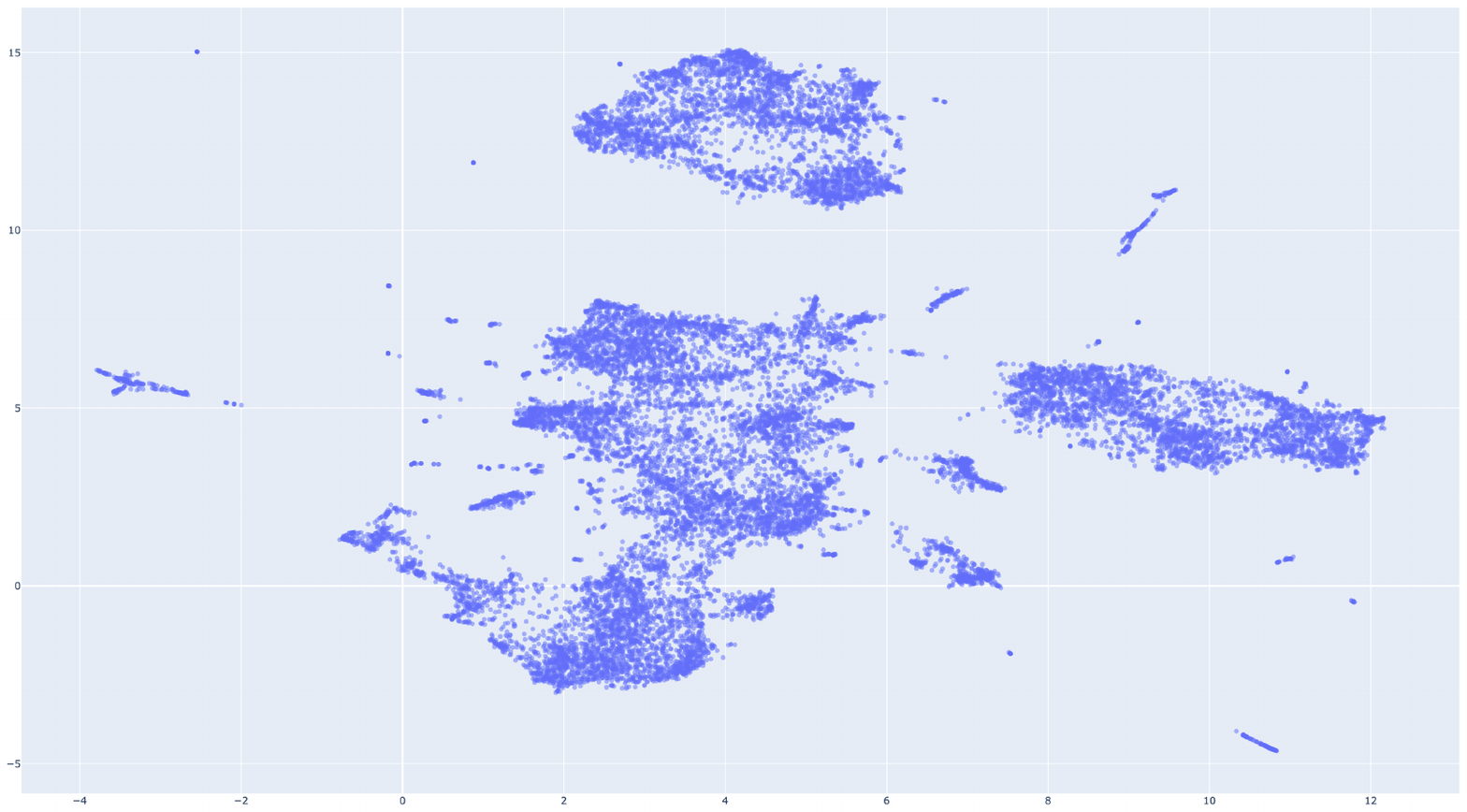}
\caption{A condensed visualization of the content within instances of the `` teacher'' token as it traveled through the RoBERTa MLM. We condensed the trajectories using PCA, then UMAP, to create the map. ``Teacher'' is a relatively monosemous word, as compared to words like ``run'' and ``set''.}
\label{fig:teacher}
\end{figure*}

We used consolidation strategies -- UMAP (Uniform Manifold Approximation and Projection) and PCA (Principle Component Analysis) -- to boil down the large amount of data about each word into something human-readable. PCA was used to truncate the results to the 50 most significant components, which captured the vast majority of the variance (above 99\%). Both contributed to the final results, the clusters seen in the visualizations, Figures~\ref{fig:bank},~\ref{fig:the},~\ref{fig:run},~\ref{fig:teacher}. It is worth noting that UMAP can result in ``finer clustering than is necessarily present in the data'', which we kept in mind (or at least tried to) when interpreting the resultant clusters~\cite{mcinnes2020umap, umapdocs2024}.

Using Python, we retrieved spans of text around our exemplar tokens from a corpus of about 7,000 English-language works of fiction, most of which were in the science fiction or fantasy genre. We did minimal pre-processing of the text, which was from e-reader formats and fairly messy. For each exemplar token, we found every occurrence across the texts, excluded any that didn't have sufficient context before and after the exemplar token, then shuffled them randomly. We used a BPE tokenizer to parse the text in order to find 100 tokens preceding and following the exemplar token.\footnote{As an interesting aside, Liu \textit{et al.}, 2019, the creators of RoBERTa, found ``only slight differences'' between a tokenizer with character-level resolution and one with byte-level resolution, even though the byte-level one should in general be more robust to OOV problems. They found the BPE encoding had ``slightly
worse end-task performance on some tasks'', but decided that the benefits of a ``universal encoding scheme'' outweighed the minor loss in performance. They called for ``a more detailed comparison of these encodings'' in future work -- while we're not comparing different tokenizations head-to-head, we hope this paper is spiritually in that vein, as we think one explanation for why a character-level vocabulary might have some benefits over a byte-level one could be found in our discussion of the token's role as both vehicle for distributional patterns and semantic receptacle~\cite{liu2019robertarobustlyoptimizedbert}.}

We then fed in text sequences and extracted latent space data for our token of interest as it moved through the model, as described above. We created visualizations showing this data one layer at a time in 2D as well as across all layers in 3D. We passed the instances of a single token in different contexts (every time it occurred in our corpus, as long as there was enough context on either side of it) through our model, and then applied PCA to each layer separately. Because this process led to a large amount of data being generated for each word, we only looked at a few words -- `` bank'', `` black'', `` cold'', `` David'', `` hand'', `` hot'', `` human'', `` left'', `` man'', `` of'', `` one'', `` right'', `` run'', `` set'', `` teacher'', `` the'', `` white'', `` woman''. In this paper, we used ``bank'', ``the'', and ``run'', and paid particular attention to the ``bank'' data.

\appendixsection{Text data}

We used the novel \textit{Alice in Wonderland}~\cite{carroll1865alice} in some of our examples because the text is freely available online via Project Gutenberg~\cite{project_gutenberg}, and because it has sufficient variation in both its syntax and vocabulary. Furthermore, because it is a childhood classic and used often in computer science examples (for NLP, for ML), we figured it had a good chance of being familiar to readers, which might make the intention of the examples more readily apparent.

We used the wikitext dataset, specifically \textit{wikitext-2-raw-v1}, which contains about 2,000,000 tokens and 600 articles~\cite{merity2016pointer}, as a matter of convenience. Similarly, we used the dataset of about 7,000 works of English fiction because it was accessible and was likely to contain sufficient variation in syntax and vocabulary. We did not think there were many constraints in this area with respect to this project: although variations in training data/ corpus would be interesting for future work, our initial concern was only having enough probably-human-written text to surface general behaviour and information (which would hopefully be fairly robust to variations in the text).

\appendixsection{Surfaced syntax}

\begin{figure}
\centering
\includegraphics[width=0.5\textwidth]{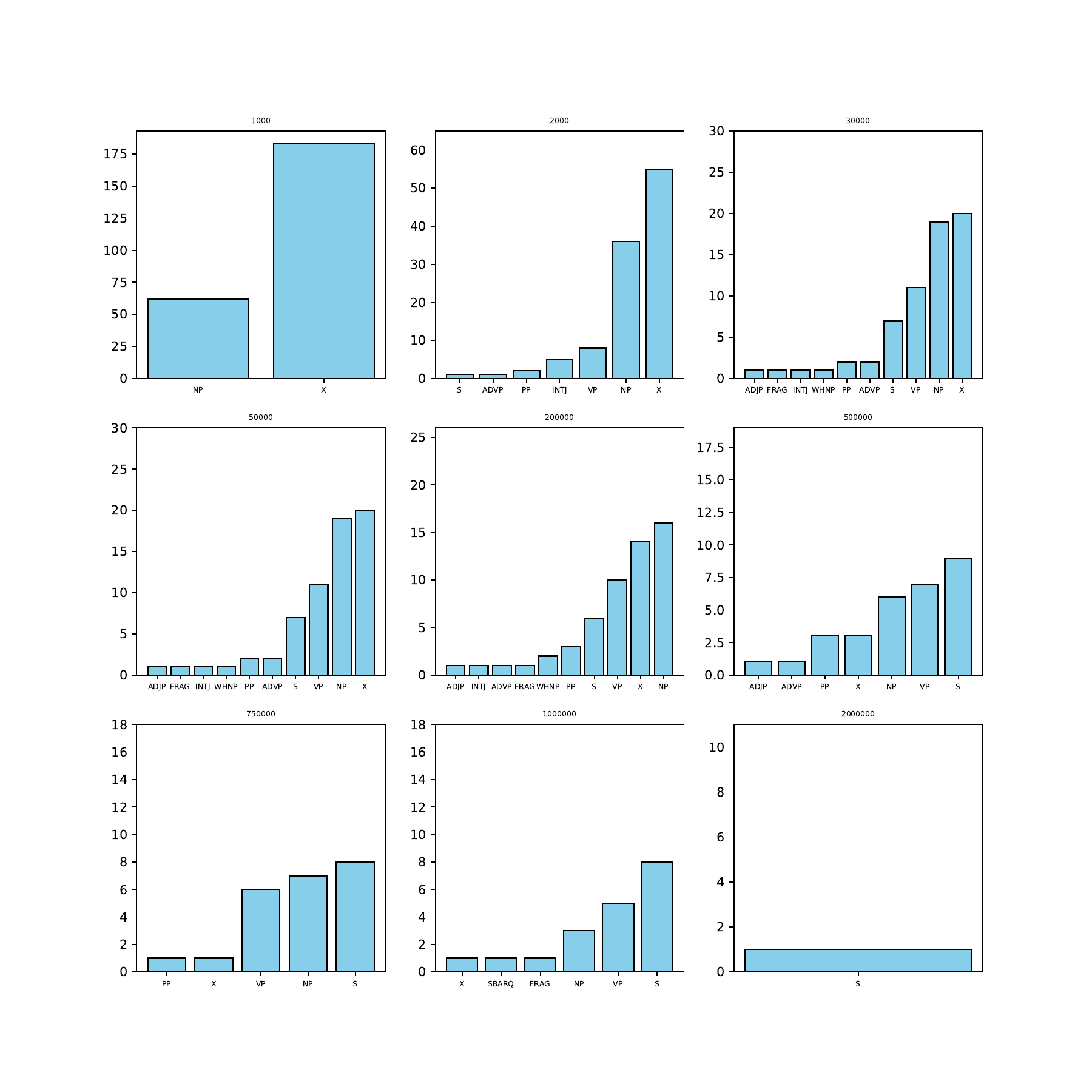}
\caption{This chart shows counts in each category of label (the constituent structure label for spans of text) assigned to tokens by the benepar and spaCy package pipeline for syntactic constituency parsing, as the vocabulary size is dialed up (with inconsistent step size shown). The dataset used was wikitext, plus the first chunk from \textit{Alice in Wonderland}.}
\label{fig:labelgrid}
\end{figure}

\begin{figure}
\centering
\includegraphics[width=0.5\textwidth]{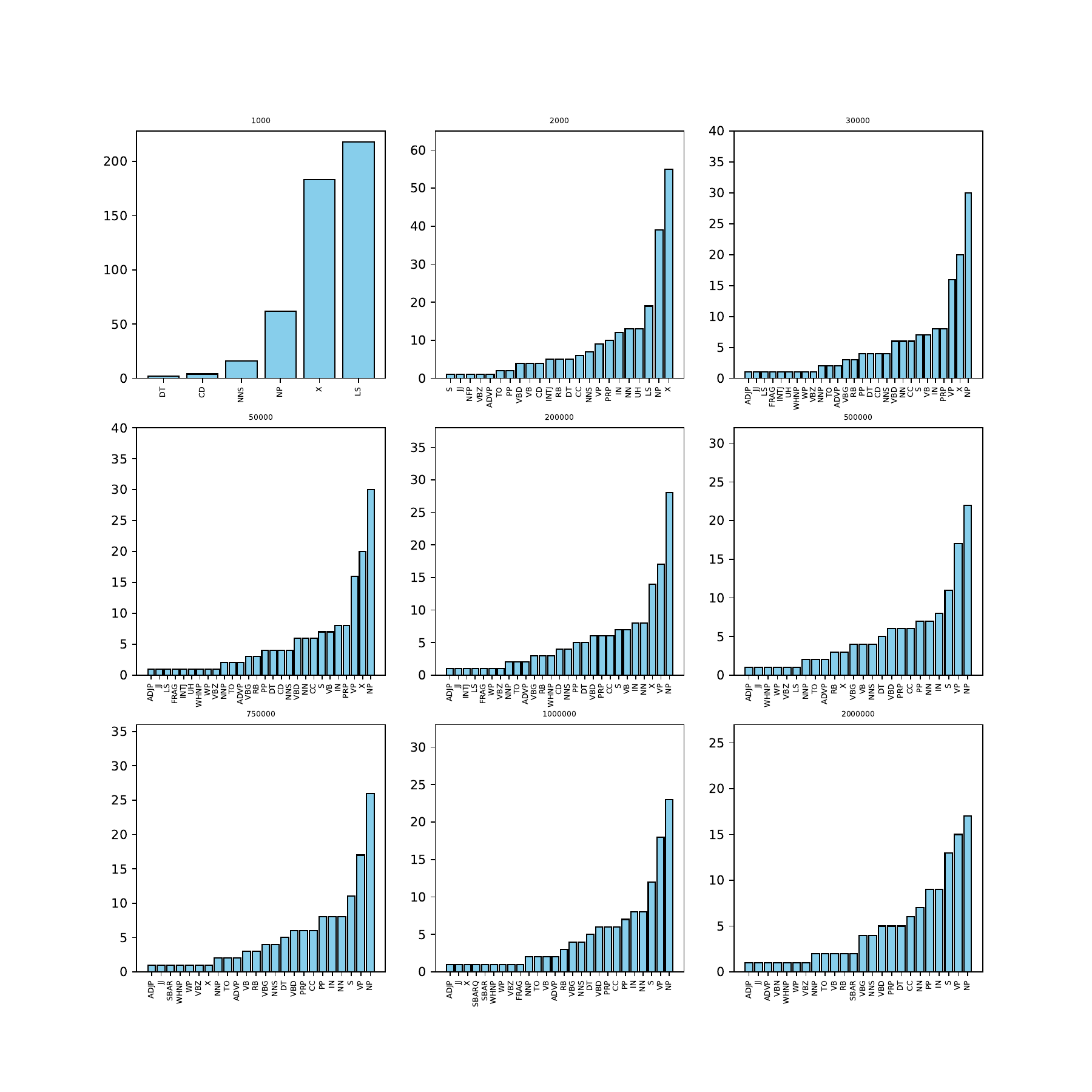}
\caption{This chart shows counts in each constituent structure category assigned to tokens by the benepar and spay package pipeline for syntactic constituency parsing, as the vocabulary size is dialed up (with inconsistent step size shown). The dataset used was wikitext, plus the first chunk from \textit{Alice in Wonderland}.}
\label{fig:syntaxgrid}
\end{figure}

Figures~\ref{fig:labelgrid},~\ref{fig:syntaxgrid} show counts in each category assigned to tokens by the benepar and spaCy package pipeline for syntactic constituency parsing, as the vocabulary size is dialed up (with inconsistent step size shown). The dataset used was wikitext, plus the first chunk from \textit{Alice in Wonderland}.

\appendixsection{Vocabulary results}

\begin{figure}
    \centering
    \includegraphics[width=0.5\textwidth]{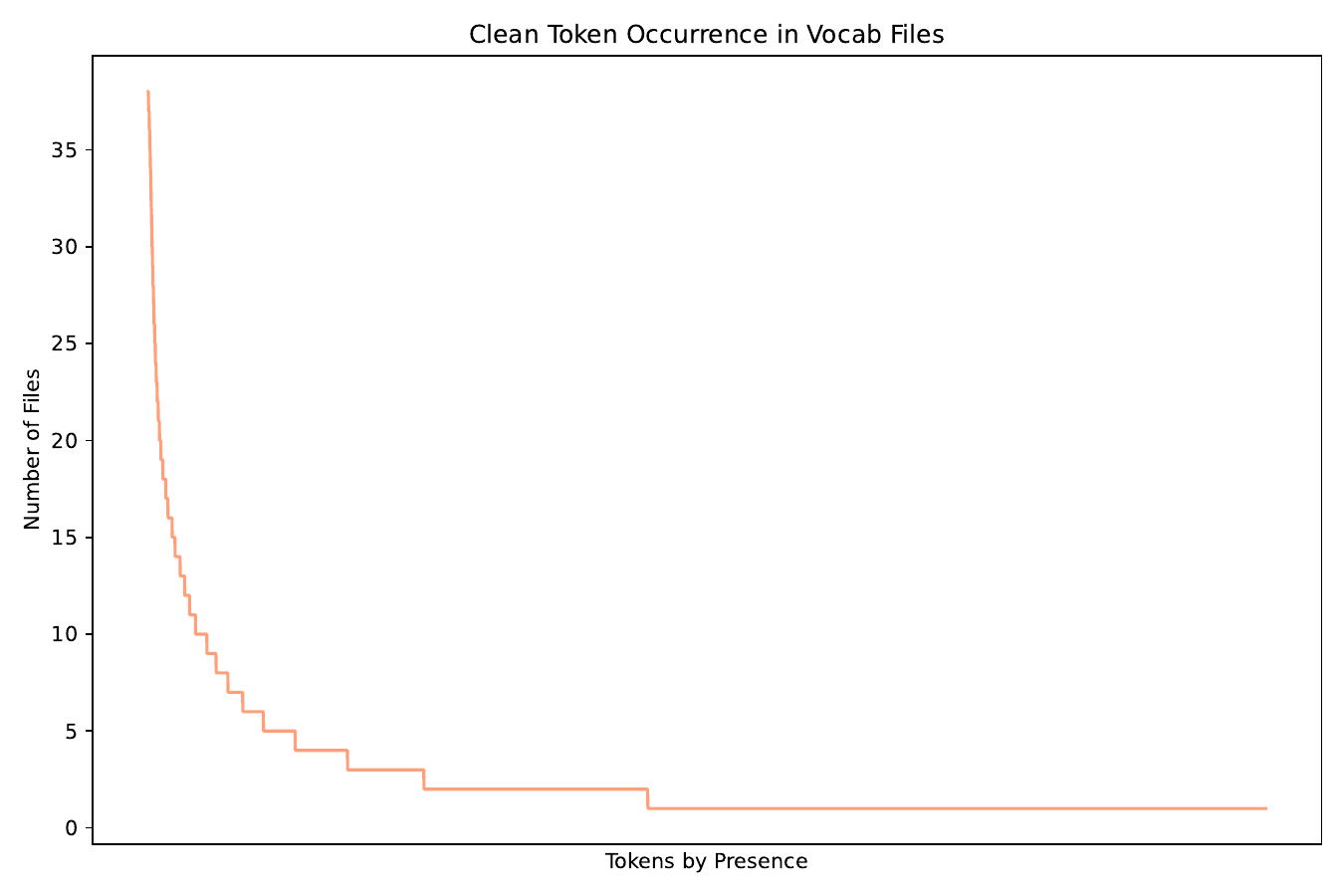}
    \includegraphics[width=0.5\textwidth]{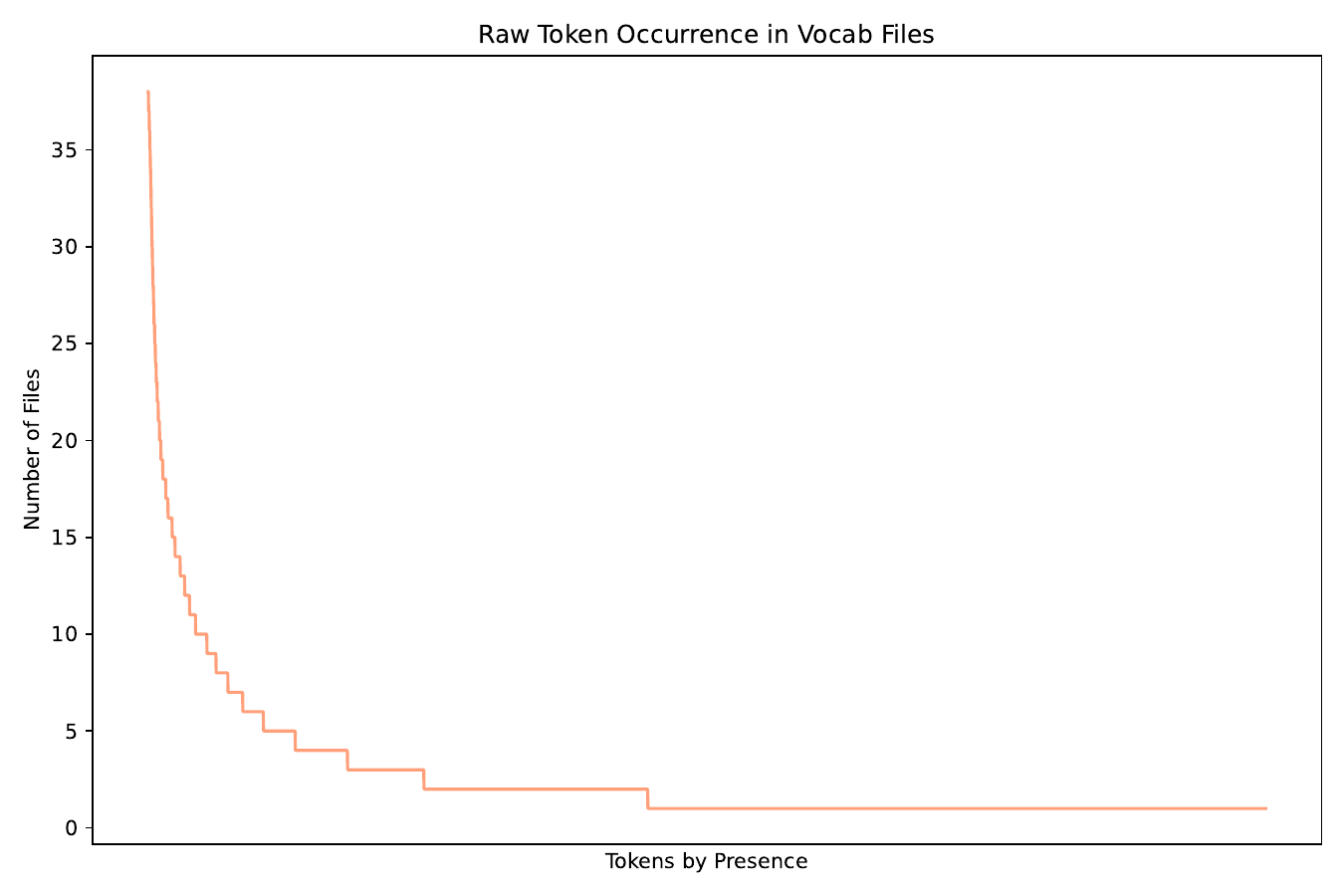}
    \caption{The first chart shows the number of files each clean token occurs in (plot downsampled by taking every tenth out of 569,810 tokens). The second chart  shows the same for the raw tokens (2,909,850 tokens, downsampled by every 25th).}
    \label{fig:clean_raw_token_occurrence_vocabfiles}
\end{figure}

\begin{figure}
    \centering
    \includegraphics[width=0.5\textwidth]{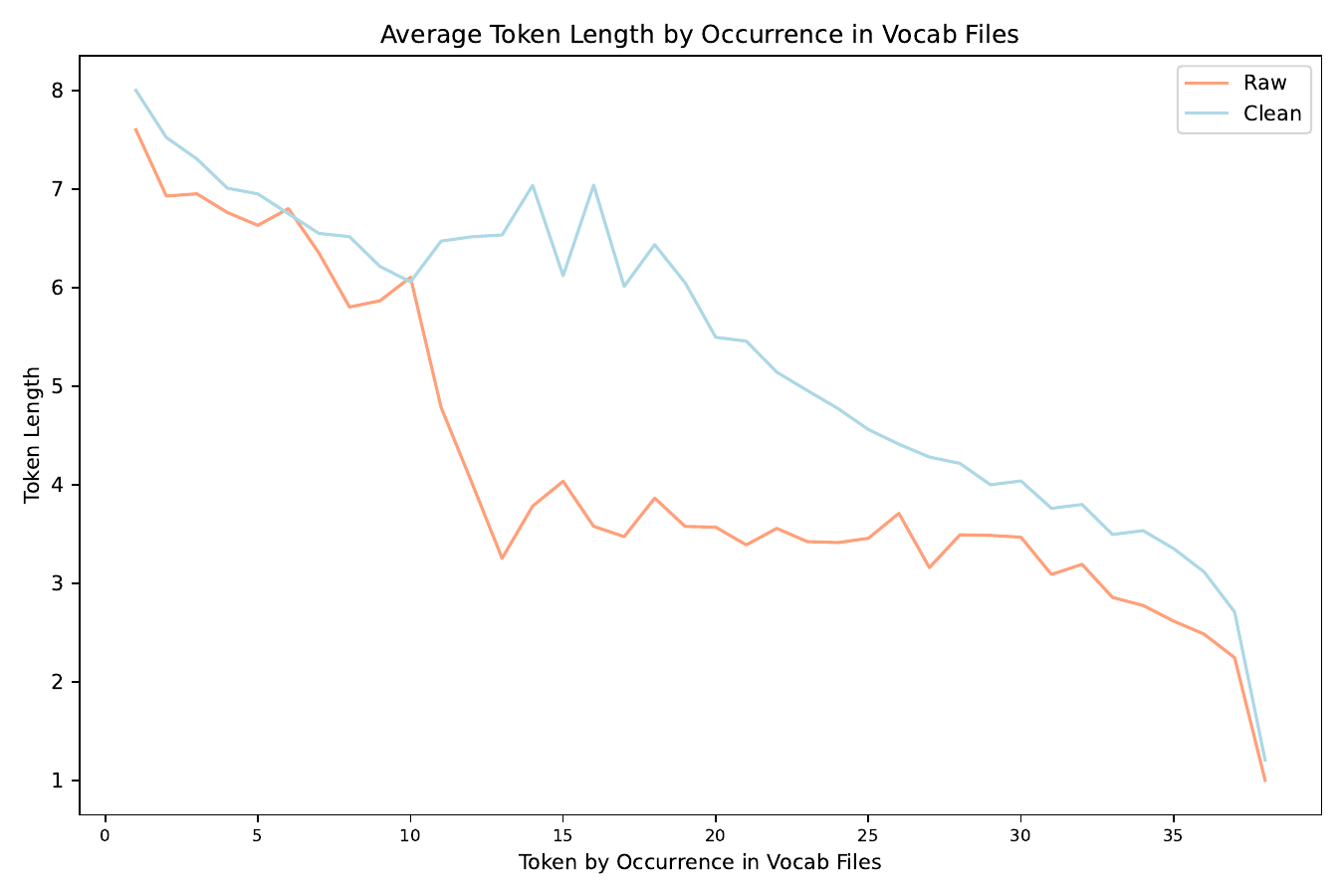}
    \caption{This chart shows the average length of the clean tokens (569,810 tokens) and the average length of the raw tokens (2,909,850 tokens) that occur in each number of files from 1 - 38.}
    \label{fig:avg_token_occurrence_vocabfiles}
\end{figure}

\begin{figure}
    \centering
    \includegraphics[width=0.5\textwidth]{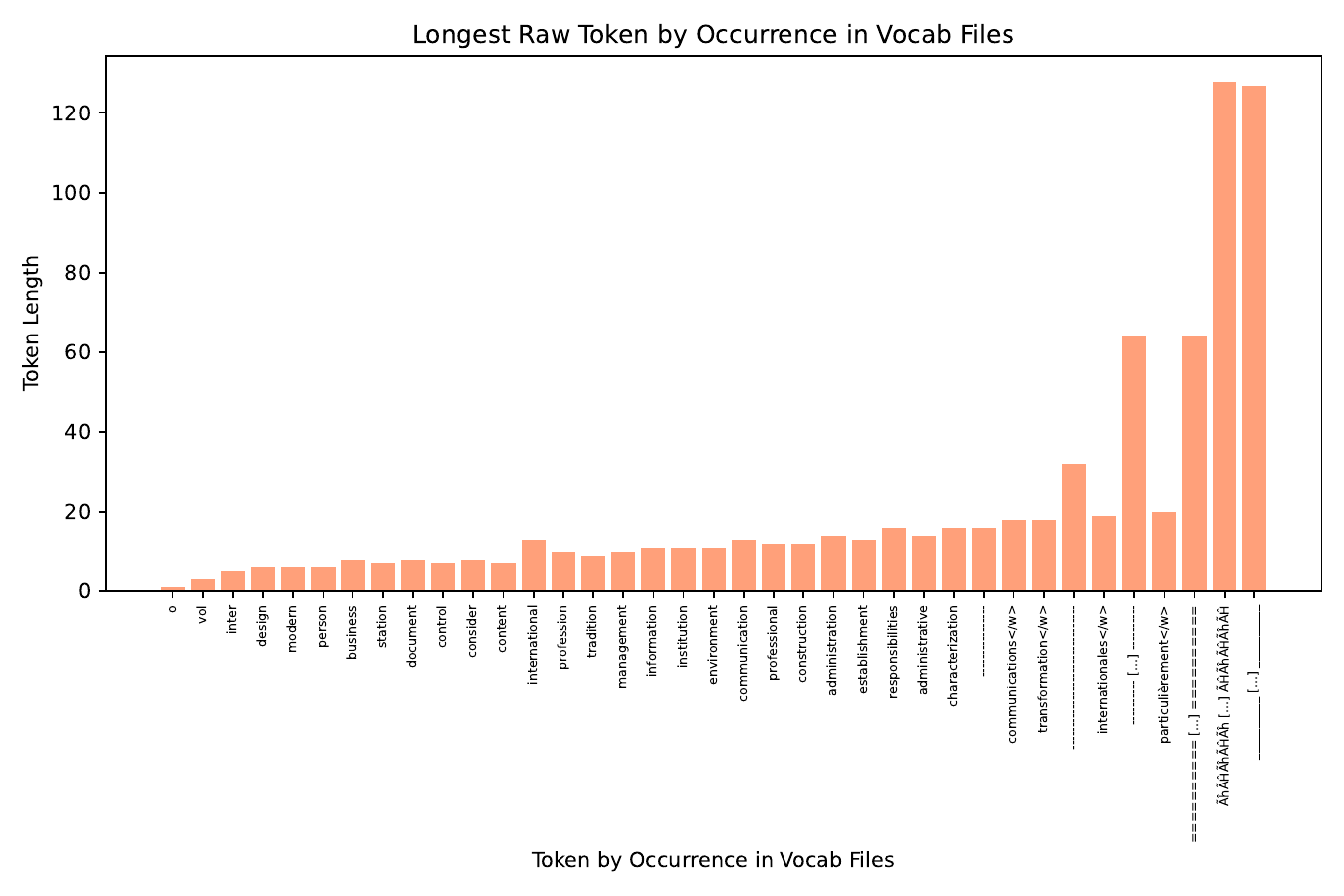}
    \caption{This chart shows the longest raw token length (of 2,909,850 tokens) that occurs in each number of files from 1 - 38, along with an example of a token of that length. Some tokens are trimmed for readability (indicated by bracketed ellipses in the chart).}
\label{fig:longest_raw_token_length_occurrence_vocabfiles}
\end{figure}

\begin{figure}
    \centering
    \includegraphics[width=0.5\textwidth]{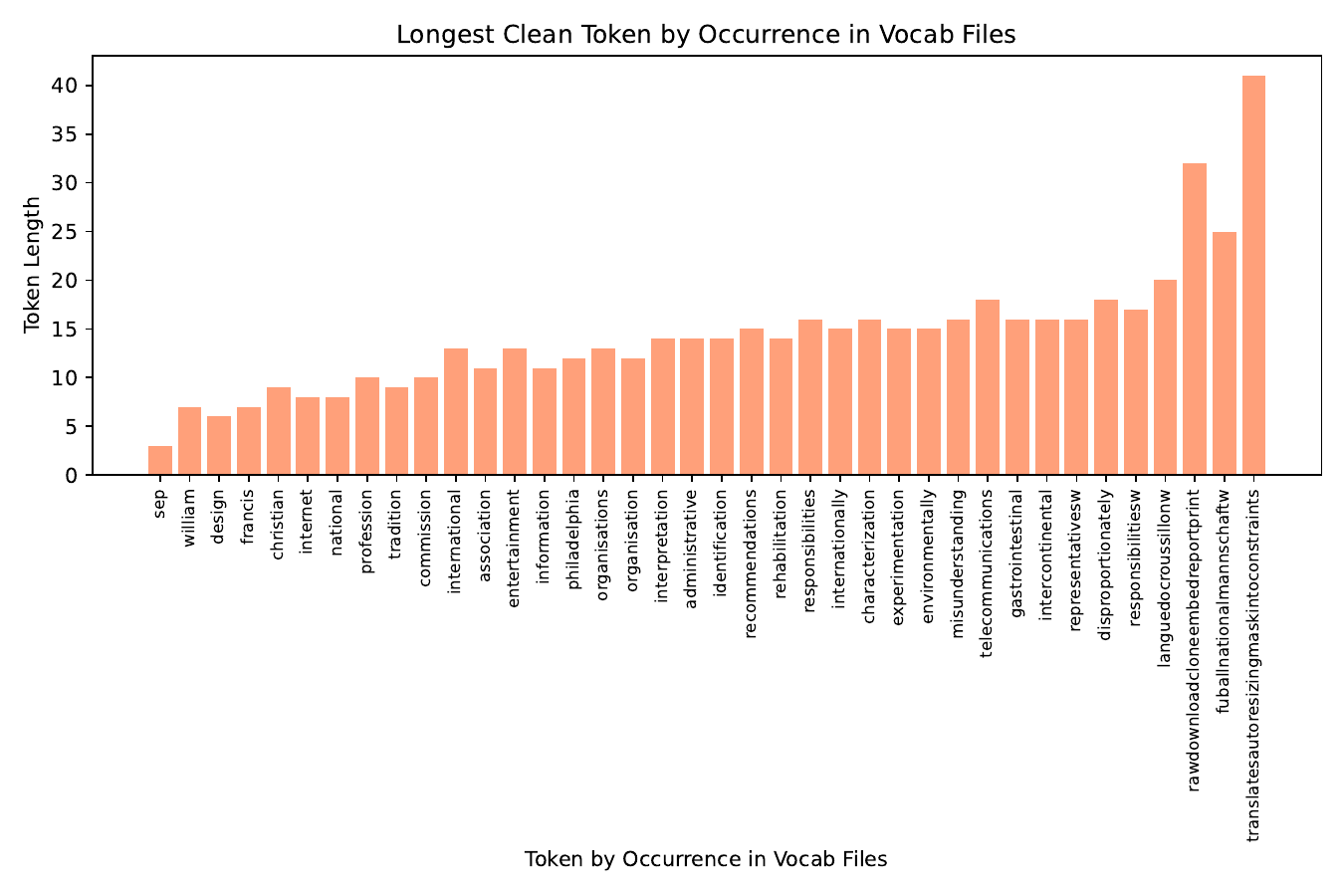}
    \caption{This chart shows the longest length of clean token in each number of files from 1 - 38 and an example token (there may be additional tokens of equal length in that number of files). Very roughly, this can be thought of as one of the most significant tokens at that stage of tokenization. The token ``sep''likely refers to separation (this is a fairly common special token in LLMs). Note that choices we made as to cleaning changed these tokens' counts, for example, by lowercasing and stripping something like ``sep'', we consolidated down variants so that ``sep'' itself appeared across more files. This is why the longest raw token in all 38 files is not also ``sep'' (and is shorter).}
\label{fig:longest_clean_token_length_occurrence_vocabfiles}
\end{figure}

Figure~\ref{fig:clean_raw_token_occurrence_vocabfiles} shows the file inclusion of the tokens. Figure~\ref{fig:avg_token_occurrence_vocabfiles} shows the average length of raw and clean tokens by file inclusion. Figures~\ref{fig:longest_raw_token_length_occurrence_vocabfiles},~\ref{fig:longest_clean_token_length_occurrence_vocabfiles} show an example of the longest token that occurs at each level of file inclusion.

\begin{figure}
    \centering
    \includegraphics[width=0.5\textwidth]{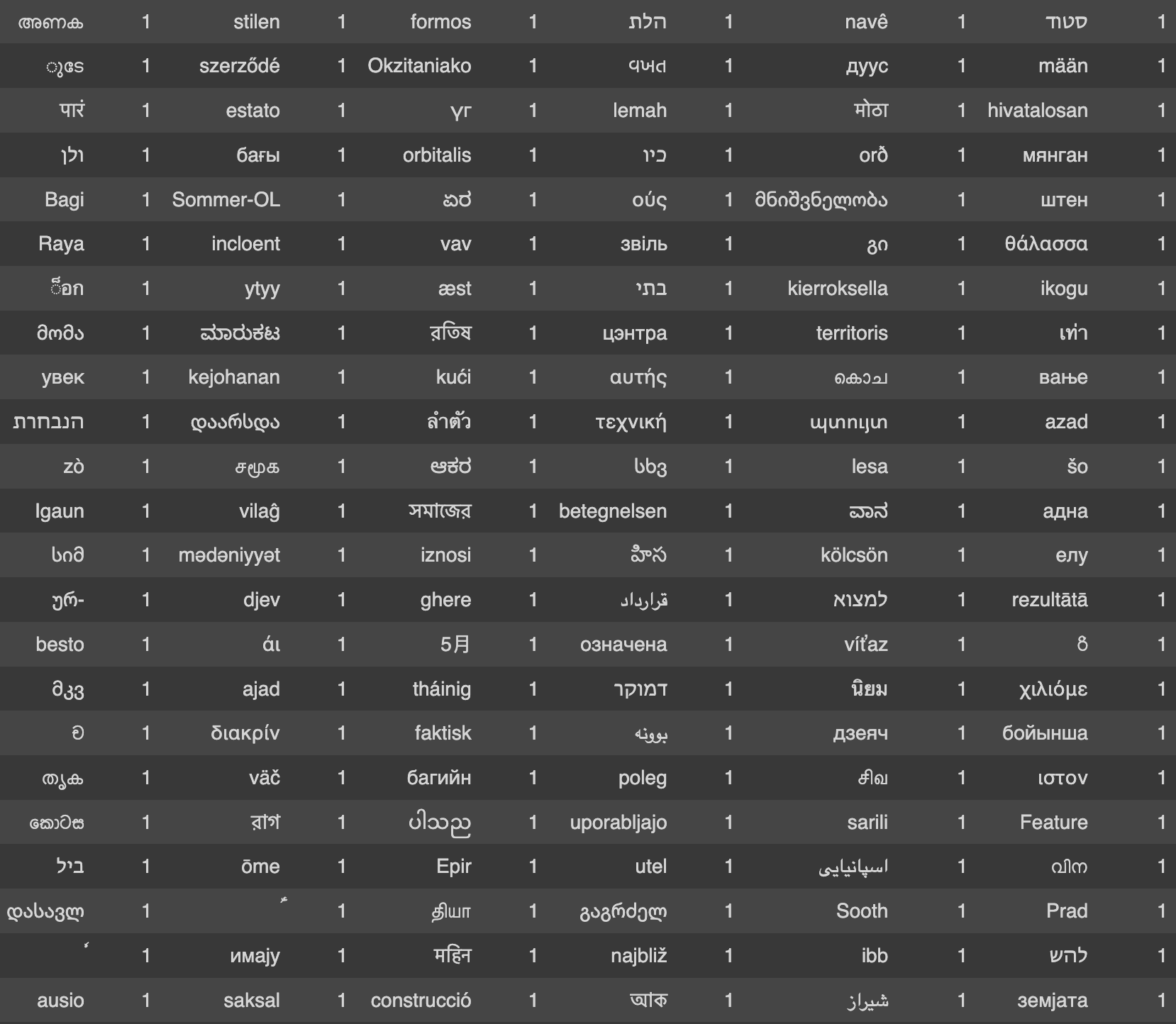}
    \caption{Some of the least common raw tokens and the number of files they occur in (from 1 - 38).}
    \label{fig:uncommon_raw_tokens}
\end{figure}

Figure~\ref{fig:uncommon_raw_tokens} shows some of the least common raw tokens and the number of files they occur in.

\begin{figure}
    \centering
    \includegraphics[width=0.5\textwidth]{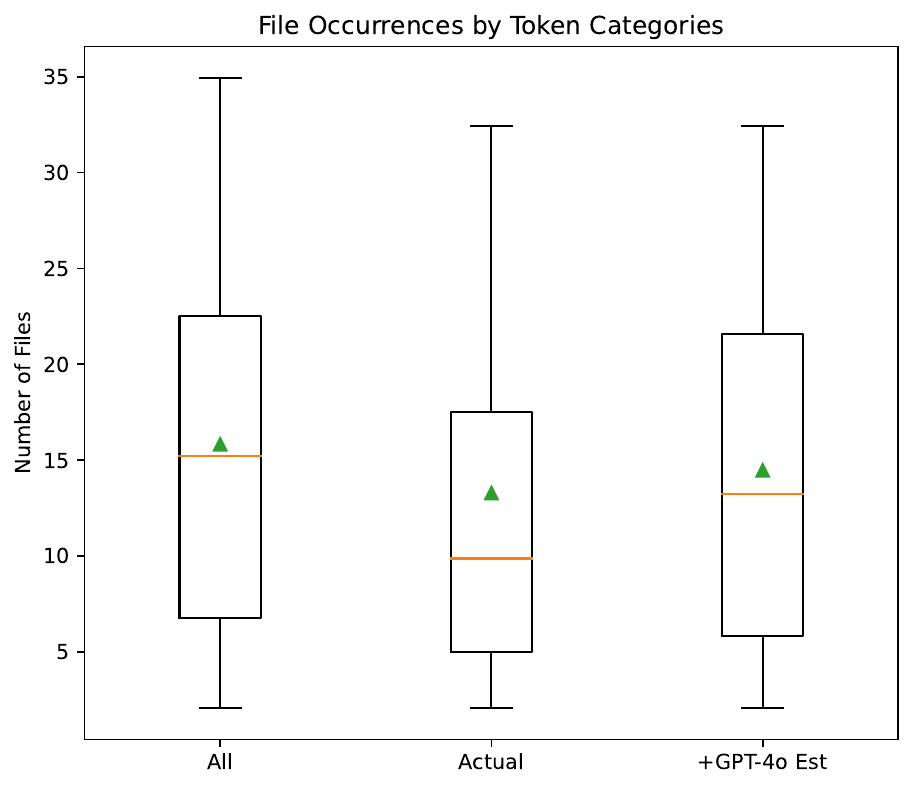}
    \caption{This chart shows the distributions of the average number of files (rounded to two decimal places) tokens of that category occurred in. The categories are all categories; all categories with estimates and hypothetical values excluded; all categories except the hypothetical CSW19 POS categories.}
    \label{fig:box_plots_all}
\end{figure}

\begin{figure}
    \centering
    \includegraphics[width=0.5\textwidth]{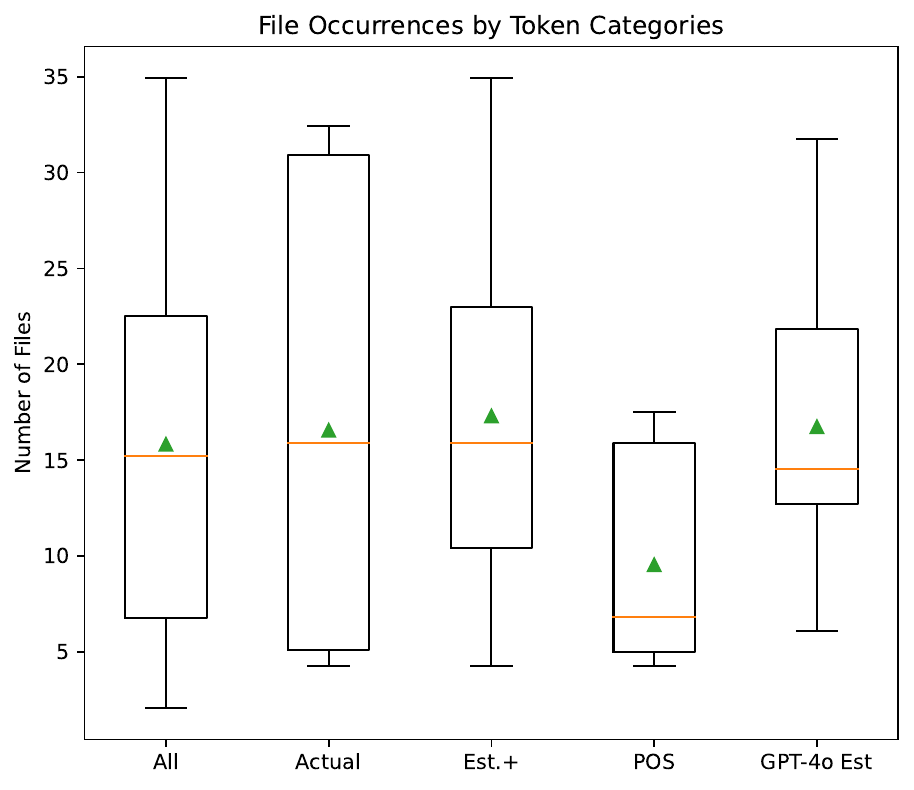}
    \caption{This chart shows the distributions (rounded to two decimal places) of the average number of files tokens of that category occurred in. The categories are all categories; categories related to grammar (affixes, morphemes, parts of speech, etc.) excluding estimated or hypothetical values; categories related to grammar (affixes, morphemes, parts of speech, etc.); POS categories (nouns, verbs, etc.) excluding estimates; and the POS category estimates for GPT-4o. The GPT-4o estimates are potentially skewed higher at least in part because we are only looking at the tokens that also occurred in the hugging face files; both the POS and the overall word estimates for GPT-4o are higher than their hugging face counterparts.}
    \label{fig:box_plots_grammar}
\end{figure}

Figure~\ref{fig:box_plots_all} shows the distributions of the average number of files (rounded to two decimal places) tokens of each category occurred in. Figure~\ref{fig:box_plots_grammar} shows the distributions (rounded to two decimal places) of the average number of files tokens of that category occurred in, for more constrained categories.

Figure~\ref{fig:numfiles_categories_all} shows the average number of files (rounded to two decimal places) each category of token occurred in (across the 38 hugging face vocabulary files), and hypothetical/ estimates for categories in GPT-4o using the number of times the tokens in GPT-4o occurred in the hugging face files.

Table~\ref{tab:fileocc_length} shows categories of token with their average file inclusion and average length (when considered).

\begin{table}
    \centering
    \small
\begin{tabular}{lrl}
\toprule
Category & Number of Files & Lengths \\
\midrule
Proper Nouns (Approximate) & 2.07 & 7.31 \\
Raw Tokens & 2.07 & 7.21 \\
Clean Tokens & 2.69 & 7.61 \\
Adverbs (CSW19) & 4.27 & 9.94 \\
Nouns (CSW19) & 4.47 & 8.25 \\
Adjectives (CSW19) & 4.97 & 7.39 \\
Words (CSW19) & 5.06 & 8.12 \\
Morphemes & 5.12 & 7.94 \\
Raw Tokens (GPT-4o Estimate) & 6.07 & - \\
Bad Words & 6.62 & 5.89 \\
Verbs (CSW19) & 6.84 & 4.42 \\
Clean Tokens (GPT-4o Estimate) & 9.16 & - \\
High Iconicity Words & 9.87 & 5.40 \\
Interjections (CSW19) & 10.41 & 5.36 \\
Interjections (CSW19 Hypothetical) & 10.41 & - \\
Adverbs (GPT-4o Estimate) & 12.64 & - \\
Nouns (GPT-4o Estimate) & 12.78 & - \\
Adjectives (GPT-4o Estimate) & 13.70 & - \\
Adverbs (CSW19 Hypothetical) & 14.17 & - \\
Verbs (GPT-4o Estimate) & 14.54 & - \\
Conjunctions (CSW19) & 15.87 & 5.73 \\
Conjunctions (CSW19 Hypothetical) & 15.87 & - \\
All Words with Iconicity Scores & 16.01 & 5.79 \\
Prepositions (CSW19) & 16.39 & 4.85 \\
Prepositions (CSW19 Hypothetical) & 16.39 & - \\
Interjections (GPT-4o Estimate) & 16.79 & - \\
Pronouns (CSW19) & 17.51 & 8.66 \\
Pronouns (CSW19 Hypothetical) & 17.51 & - \\
Prepositions (GPT-4o Estimate) & 21.33 & - \\
Pronouns (GPT-4o Estimate) & 22.38 & - \\
Conjunctions (GPT-4o Estimate) & 23.00 & - \\
Function Words & 23.11 & 5.24 \\
Adjectives (CSW19 Hypothetical) & 26.14 & - \\
Less Common Affixes (SA) & 30.63 & - \\
Affixes & 31.24 & 3.43 \\
Affixes (SA) & 31.34 & 3.43 \\
Affixes (GPT-4o Estimate) & 31.74 & - \\
More Common Affixes (SA) & 32.45 & - \\
Nouns (CSW19 Hypothetical) & 34.50 & - \\
Verbs (CSW19 Hypothetical) & 34.93 & - \\
\bottomrule
\end{tabular}
    \caption{File Occurrences and Token Lengths by category, rounded to two decimal places.}
    \label{tab:fileocc_length}
\end{table}

\begin{figure}
    \centering
    \includegraphics[width=0.5\textwidth]{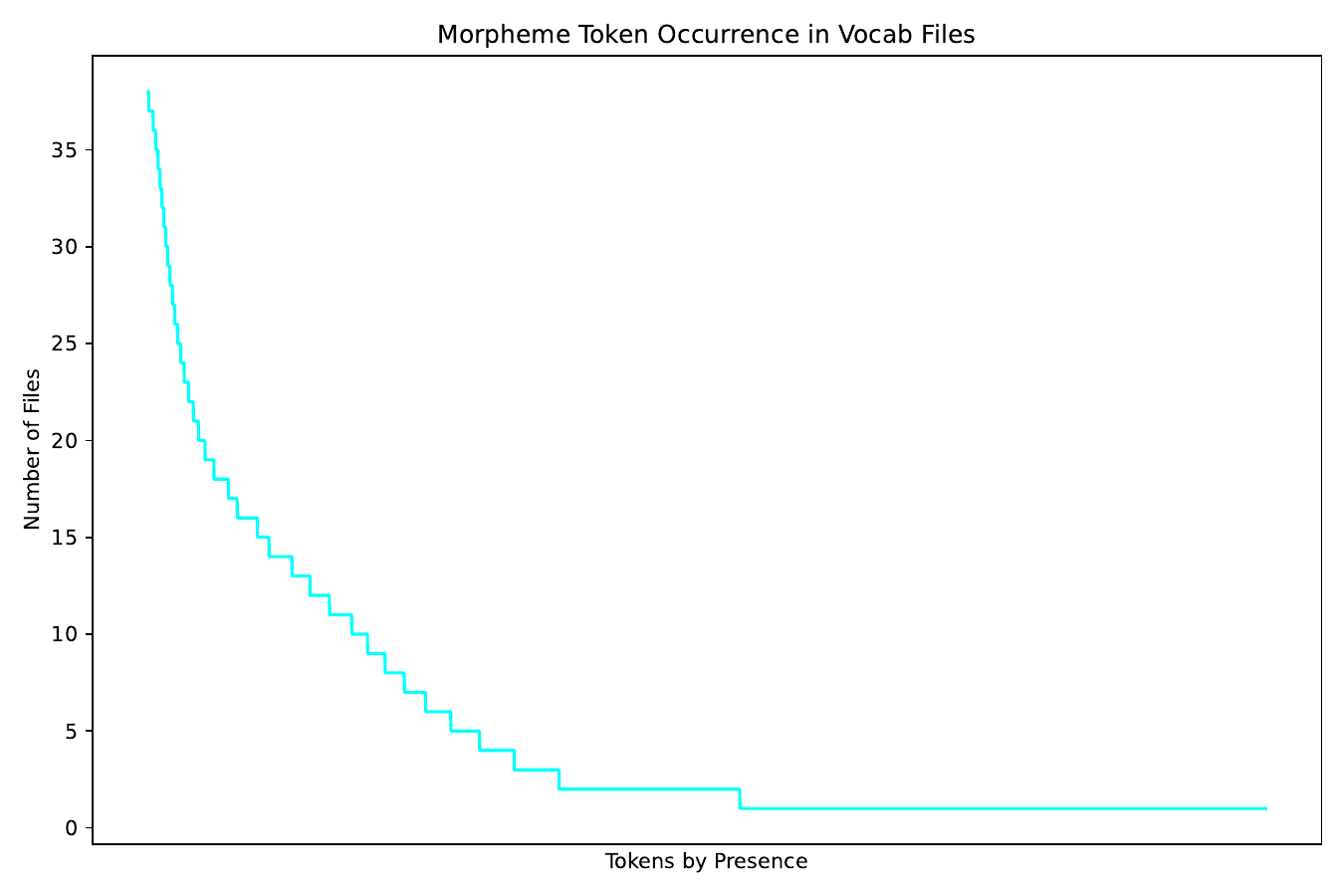}
    \includegraphics[width=0.5\columnwidth]{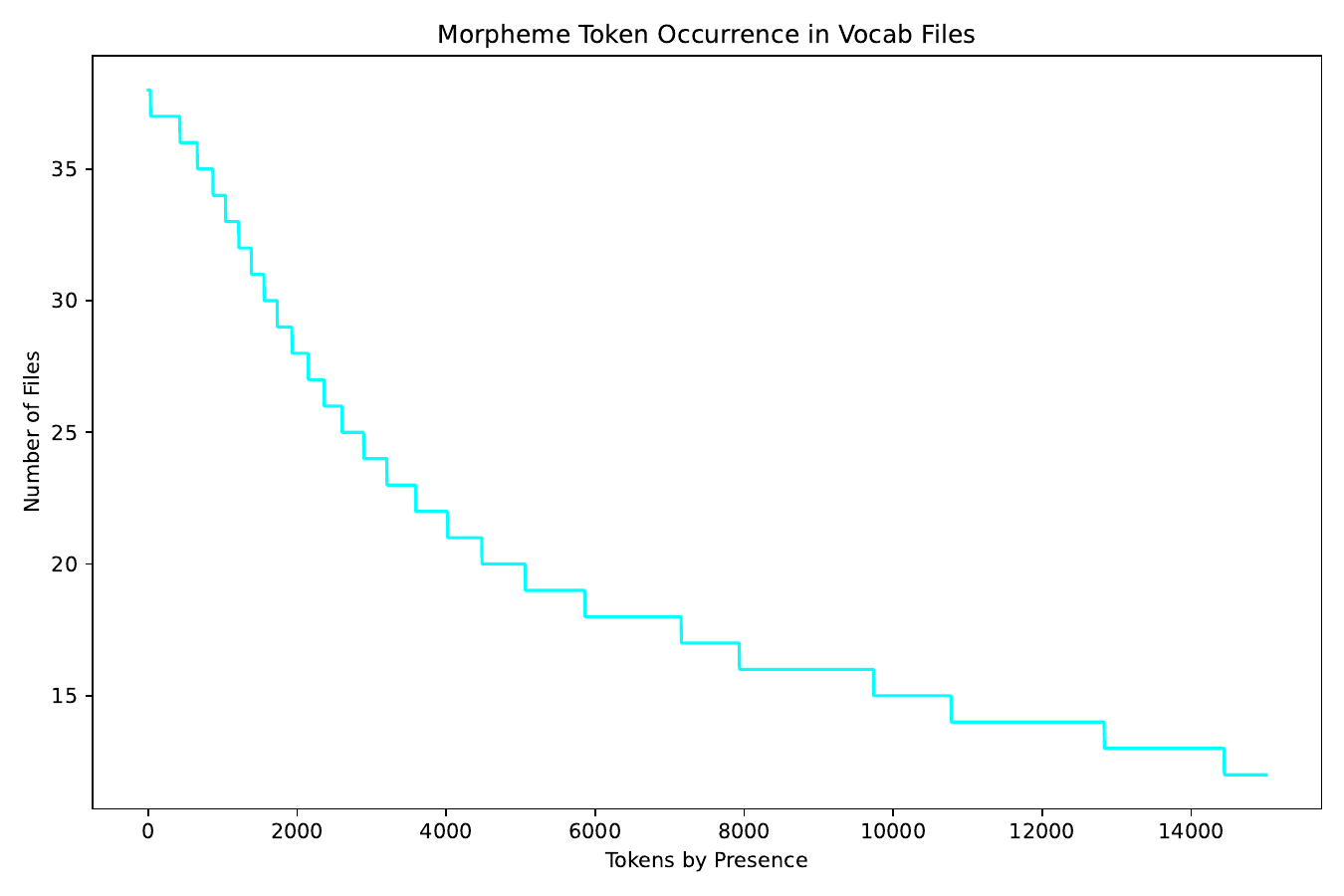}
    \caption{These charts show the number of files each morpheme occurs in, from our largest list approximating morphemes (words, affixes). The second chart shows more detail in the shape of the head: the most common 15,000 tokens (out of 99,841).}
    \label{fig:morpheme_token_occurrence_vocabfiles}
\end{figure}

Figure~\ref{fig:morpheme_token_occurrence_vocabfiles} shows the number of files each string occurs in, from our largest list roughly approximating morphemes (words, affixes).
Some of the most common rough morphemes from our largest list and the number of files they occur in are listed in Figure~\ref{tab:common_rough_morphemes}. (See also Table~\ref{tab:morpheme_csw19} and Table~\ref{tab:model_comparison_morpheme}.)

Out of our largest rough morpheme list, which includes words and affixes, a subset of the 458,685 morphemes (99,841) occur 511,216 times across the files. The 99,841 unique morphemes in the files have an average length of \num{6.5879178272980505} and an average unique length of \num{7.942678859386424}.

\begin{table}
    \centering
    \tiny
\begin{tabular}{lrrrrrrr}
\toprule
Vocabulary & Tokens & Cleaned & Clean/Raw & Morph. & Morph./Clean & Morph./Raw & Morph./All Morph. \\
\midrule
xlm-mlm-100-1280 & 199999 & 71780 & 0.358902 & 5739 & 0.079953 & 0.028695 & 0.012512 \\
 bert-base-german-dbmdz-uncased & 31101 & 22674 & 0.729044 & 5701 & 0.251433 & 0.183306 & 0.012429 \\
 xlm-mlm-enro-1024 & 64592 & 48190 & 0.746068 & 4467 & 0.092696 & 0.069157 & 0.009739 \\
 funnel-transformer\_small & 30522 & 24846 & 0.814036 & 20796 & 0.836996 & 0.681345 & 0.045338 \\
 openai-gpt & 40477 & 39542 & 0.976900 & 3686 & 0.093217 & 0.091064 & 0.008036 \\
 bert-base-uncased & 30522 & 24844 & 0.813970 & 20796 & 0.837063 & 0.681345 & 0.045338 \\
 bert-base-german-cased & 29999 & 18703 & 0.623454 & 4230 & 0.226167 & 0.141005 & 0.009222 \\
 xlm-mlm-en-2048 & 30145 & 28463 & 0.944203 & 3813 & 0.133963 & 0.126489 & 0.008313 \\
 facebook\_mbart-large-cc25 & 250025 & 74711 & 0.298814 & 12920 & 0.172933 & 0.051675 & 0.028167 \\
 albert-base-v1 & 30000 & 25312 & 0.843733 & 21014 & 0.830199 & 0.700467 & 0.045814 \\
 camembert-base & 32004 & 21477 & 0.671072 & 8506 & 0.396052 & 0.265779 & 0.018544 \\
 gpt2-medium & 50252 & 30573 & 0.608394 & 22270 & 0.728421 & 0.443166 & 0.048552 \\
 bert-base-german-dbmdz-cased & 31101 & 20228 & 0.650397 & 4998 & 0.247083 & 0.160702 & 0.010896 \\
 xlm-mlm-ende-1024 & 64699 & 46266 & 0.715096 & 4438 & 0.095924 & 0.068595 & 0.009675 \\
 xlm-mlm-17-1280 & 199999 & 87404 & 0.437022 & 6114 & 0.069951 & 0.030570 & 0.013329 \\
 xlm-mlm-xnli15-1024 & 94999 & 32378 & 0.340825 & 3874 & 0.119649 & 0.040779 & 0.008446 \\
 gpt2 & 41403 & 30573 & 0.738425 & 22269 & 0.728388 & 0.537860 & 0.048550 \\
 bert-base-multilingual-uncased & 105878 & 50116 & 0.473337 & 16610 & 0.331431 & 0.156879 & 0.036212 \\
 t5-small & 32100 & 22635 & 0.705140 & 14514 & 0.641219 & 0.452150 & 0.031643 \\
 xlm-mlm-enfr-1024 & 64139 & 47169 & 0.735418 & 4385 & 0.092964 & 0.068367 & 0.009560 \\
 TurkuNLP\_bert-base-finnish-cased-v1 & 50104 & 37659 & 0.751617 & 3549 & 0.094240 & 0.070833 & 0.007737 \\
 wietsedv\_bert-base-dutch-cased & 30073 & 23129 & 0.769095 & 6032 & 0.260798 & 0.200579 & 0.013151 \\
 ctrl & 245470 & 78453 & 0.319603 & 45940 & 0.585574 & 0.187151 & 0.100156 \\
 allenai\_scibert\_scivocab\_cased & 31115 & 18961 & 0.609385 & 12879 & 0.679236 & 0.413916 & 0.028078 \\
 cl-tohoku\_bert-base-japanese-char & 3999 & 31 & 0.007752 & 29 & 0.935484 & 0.007252 & 0.000063 \\
 bert-base-multilingual-cased & 119547 & 48586 & 0.406418 & 15206 & 0.312971 & 0.127197 & 0.033151 \\
 roberta-base & 50260 & 30574 & 0.608317 & 22270 & 0.728397 & 0.443096 & 0.048552 \\
 cl100k\_base & 76014 & 41927 & 0.551569 & 23527 & 0.561142 & 0.309509 & 0.051292 \\
 TurkuNLP\_bert-base-finnish-uncased-v1 & 50101 & 44155 & 0.881320 & 4091 & 0.092651 & 0.081655 & 0.008919 \\
 flaubert\_flaubert\_base\_uncased & 67541 & 44725 & 0.662190 & 3819 & 0.085388 & 0.056543 & 0.008326 \\
 transfo-xl-wt103 & 267734 & 215699 & 0.805647 & 91443 & 0.423938 & 0.341544 & 0.199359 \\
 xlm-roberta-base & 250000 & 74689 & 0.298756 & 12917 & 0.172944 & 0.051668 & 0.028161 \\
 allenai\_scibert\_scivocab\_uncased & 31089 & 21795 & 0.701052 & 15011 & 0.688736 & 0.482840 & 0.032726 \\
 bert-base-chinese & 21127 & 2870 & 0.135845 & 2024 & 0.705226 & 0.095802 & 0.004413 \\
 bert-base-cased & 28996 & 20935 & 0.721996 & 17236 & 0.823310 & 0.594427 & 0.037577 \\
 flaubert\_flaubert\_small\_cased & 68728 & 37985 & 0.552686 & 3392 & 0.089298 & 0.049354 & 0.007395 \\
 xlnet-base-cased & 31997 & 22748 & 0.710942 & 19539 & 0.858933 & 0.610651 & 0.042598 \\
 cl-tohoku\_bert-base-japanese & 31999 & 1674 & 0.052314 & 1172 & 0.700119 & 0.036626 & 0.002555 \\
\bottomrule
\end{tabular}
    \caption{A simple comparison of the number of raw and cleaned tokens in each vocabulary file, and the overlap between those tokens and our largest list of words and affixes, meant to approximate morphemes. This list includes affixes, CSW19, and S2.}
    \label{tab:model_comparison_morpheme}
\end{table}

\begin{table}
    \centering
    \tiny
\begin{tabular}{lrlrlrlrlrlrlrlrlr}
\midrule
z & 38 & ha & 37 & cl & 37 & tom & 37 & sky & 37 & top & 37 & are & 37 & ter & 37 & an & 37 \\
 k & 38 & av & 37 & ht & 37 & oc & 37 & br & 37 & iv & 37 & ling & 37 & nas & 37 & ta & 37 \\
 x & 38 & mu & 37 & ing & 37 & ale & 37 & bo & 37 & lan & 37 & ur & 37 & bl & 37 & ide & 37 \\
 r & 38 & cat & 37 & he & 37 & ol & 37 & ac & 37 & un & 37 & john & 37 & all & 37 & ox & 37 \\
 w & 38 & mon & 37 & super & 37 & gen & 37 & sk & 37 & vo & 37 & day & 37 & il & 37 & camp & 37 \\
 a & 38 & or & 37 & king & 37 & tra & 37 & bor & 37 & light & 37 & ah & 37 & mac & 37 & she & 37 \\
 j & 38 & med & 37 & out & 37 & ju & 37 & sh & 37 & power & 37 & bel & 37 & lu & 37 & by & 37 \\
 y & 38 & lor & 37 & fu & 37 & ra & 37 & op & 37 & best & 37 & ar & 37 & grand & 37 & paul & 37 \\
 l & 38 & river & 37 & it & 37 & ft & 37 & tor & 37 & vel & 37 & tim & 37 & te & 37 & sa & 37 \\
 sep & 38 & hall & 37 & ent & 37 & play & 37 & ball & 37 & um & 37 & big & 37 & open & 37 & ko & 37 \\
 p & 38 & port & 37 & bill & 37 & na & 37 & vi & 37 & im & 37 & rom & 37 & mp & 37 & ster & 37 \\
 n & 38 & boy & 37 & mar & 37 & ps & 37 & my & 37 & va & 37 & za & 37 & no & 37 & yo & 37 \\
 s & 38 & er & 37 & oo & 37 & ish & 37 & ie & 37 & sp & 37 & ee & 37 & com & 37 & ou & 37 \\
 q & 38 & del & 37 & ay & 37 & hy & 37 & ec & 37 & sm & 37 & and & 37 & ck & 37 & ash & 37 \\
 b & 38 & po & 37 & ant & 37 & cc & 37 & lam & 37 & ak & 37 & bi & 37 & tr & 37 & ll & 37 \\
 v & 38 & om & 37 & let & 37 & jean & 37 & north & 37 & dj & 37 & co & 37 & one & 37 & home & 37 \\
 f & 38 & ut & 37 & cy & 37 & may & 37 & gi & 37 & son & 37 & pro & 37 & per & 37 & off & 37 \\
 u & 38 & lt & 37 & ab & 37 & us & 37 & angel & 37 & pop & 37 & sun & 37 & ad & 37 & ok & 37 \\
 h & 38 & est & 37 & you & 37 & bar & 37 & ph & 37 & ky & 37 & ir & 37 & gold & 37 & me & 37 \\
 o & 38 & tu & 37 & mit & 37 & be & 37 & af & 37 & mark & 37 & las & 37 & ho & 37 & ley & 37 \\
 pad & 38 & land & 37 & green & 37 & pa & 37 & id & 37 & was & 37 & bb & 37 & od & 37 & mc & 37 \\
 t & 38 & show & 37 & oh & 37 & bus & 37 & web & 37 & ate & 37 & ron & 37 & run & 37 & en & 37 \\
 g & 38 & over & 37 & gn & 37 & end & 37 & ob & 37 & at & 37 & con & 37 & berg & 37 & god & 37 \\
 d & 38 & to & 37 & mr & 37 & hi & 37 & ber & 37 & que & 37 & max & 37 & arm & 37 & ill & 37 \\
 m & 38 & ca & 37 & ram & 37 & way & 37 & key & 37 & sat & 37 & go & 37 & art & 37 & th & 37 \\
 c & 38 & for & 37 & pan & 37 & re & 37 & win & 37 & gan & 37 & pe & 37 & su & 37 & on & 37 \\
 e & 38 & sam & 37 & south & 37 & war & 37 & ver & 37 & nor & 37 & mer & 37 & vin & 37 & ch & 37 \\
 i & 38 & alex & 37 & with & 37 & the & 37 & black & 37 & cr & 37 & mb & 37 & is & 37 & ton & 37 \\
 long & 37 & ser & 37 & io & 37 & em & 37 & william & 37 & wa & 37 & et & 37 & ern & 37 & dr & 37 \\
 will & 37 & pl & 37 & jan & 37 & ard & 37 & up & 37 & bon & 37 & ann & 37 & lex & 37 & master & 37 \\
\bottomrule
\end{tabular}
    \caption{Some of the most common rough morphemes from our largest list and the number of files they occur in.}
    \label{tab:common_rough_morphemes}
\end{table}

\begin{table}
\centering
\tiny
\begin{tabular}{lrrrrrrr}
\toprule
 Vocabulary & Tokens & Cleaned & Clean/Raw & Morph. & Morph./Clean & Morph./Raw & M./All Morph. \\
\midrule
xlm-mlm-100-1280 & 199999 & 71780 & 0.358902 & 3929 & 0.054737 & 0.019645 & 0.014053 \\
 bert-base-german-dbmdz-uncased & 31101 & 22674 & 0.729044 & 4030 & 0.177737 & 0.129578 & 0.014414 \\
 xlm-mlm-enro-1024 & 64592 & 48190 & 0.746068 & 3339 & 0.069288 & 0.051694 & 0.011943 \\
 funnel-transformer\_small & 30522 & 24846 & 0.814036 & 18183 & 0.731828 & 0.595734 & 0.065036 \\
 openai-gpt & 40477 & 39542 & 0.976900 & 2839 & 0.071797 & 0.070139 & 0.010154 \\
 bert-base-uncased & 30522 & 24844 & 0.813970 & 18183 & 0.731887 & 0.595734 & 0.065036 \\
 bert-base-german-cased & 29999 & 18703 & 0.623454 & 2902 & 0.155162 & 0.096737 & 0.010380 \\
 xlm-mlm-en-2048 & 30145 & 28463 & 0.944203 & 2862 & 0.100552 & 0.094941 & 0.010237 \\
 facebook\_mbart-large-cc25 & 250025 & 74711 & 0.298814 & 9408 & 0.125925 & 0.037628 & 0.033650 \\
 albert-base-v1 & 30000 & 25312 & 0.843733 & 18129 & 0.716222 & 0.604300 & 0.064843 \\
 camembert-base & 32004 & 21477 & 0.671072 & 6591 & 0.306886 & 0.205943 & 0.023574 \\
 gpt2-medium & 50252 & 30573 & 0.608394 & 19665 & 0.643215 & 0.391328 & 0.070336 \\
 bert-base-german-dbmdz-cased & 31101 & 20228 & 0.650397 & 3497 & 0.172879 & 0.112440 & 0.012508 \\
 xlm-mlm-ende-1024 & 64699 & 46266 & 0.715096 & 3285 & 0.071002 & 0.050774 & 0.011750 \\
 xlm-mlm-17-1280 & 199999 & 87404 & 0.437022 & 4247 & 0.048590 & 0.021235 & 0.015190 \\
 xlm-mlm-xnli15-1024 & 94999 & 32378 & 0.340825 & 2748 & 0.084872 & 0.028927 & 0.009829 \\
 gpt2 & 41403 & 30573 & 0.738425 & 19664 & 0.643182 & 0.474941 & 0.070333 \\
 bert-base-multilingual-uncased & 105878 & 50116 & 0.473337 & 12538 & 0.250180 & 0.118419 & 0.044845 \\
 t5-small & 32100 & 22635 & 0.705140 & 12727 & 0.562271 & 0.396480 & 0.045521 \\
 xlm-mlm-enfr-1024 & 64139 & 47169 & 0.735418 & 3158 & 0.066951 & 0.049237 & 0.011295 \\
 TurkuNLP\_bert-base-finnish-cased-v1 & 50104 & 37659 & 0.751617 & 2286 & 0.060703 & 0.045625 & 0.008176 \\
 wietsedv\_bert-base-dutch-cased & 30073 & 23129 & 0.769095 & 4278 & 0.184963 & 0.142254 & 0.015301 \\
 ctrl & 245470 & 78453 & 0.319603 & 39490 & 0.503359 & 0.160875 & 0.141245 \\
 allenai\_scibert\_scivocab\_cased & 31115 & 18961 & 0.609385 & 11492 & 0.606086 & 0.369340 & 0.041104 \\
 cl-tohoku\_bert-base-japanese-char & 3999 & 31 & 0.007752 & 28 & 0.903226 & 0.007002 & 0.000100 \\
 bert-base-multilingual-cased & 119547 & 48586 & 0.406418 & 11327 & 0.233133 & 0.094749 & 0.040514 \\
 roberta-base & 50260 & 30574 & 0.608317 & 19665 & 0.643194 & 0.391265 & 0.070336 \\
 cl100k\_base & 76014 & 41927 & 0.551569 & 20423 & 0.487109 & 0.268674 & 0.073048 \\
 TurkuNLP\_bert-base-finnish-uncased-v1 & 50101 & 44155 & 0.881320 & 2687 & 0.060854 & 0.053632 & 0.009611 \\
 flaubert\_flaubert\_base\_uncased & 67541 & 44725 & 0.662190 & 2647 & 0.059184 & 0.039191 & 0.009468 \\
 transfo-xl-wt103 & 267734 & 215699 & 0.805647 & 78289 & 0.362955 & 0.292413 & 0.280019 \\
 xlm-roberta-base & 250000 & 74689 & 0.298756 & 9405 & 0.125922 & 0.037620 & 0.033639 \\
 allenai\_scibert\_scivocab\_uncased & 31089 & 21795 & 0.701052 & 13359 & 0.612939 & 0.429702 & 0.047782 \\
 bert-base-chinese & 21127 & 2870 & 0.135845 & 1547 & 0.539024 & 0.073224 & 0.005533 \\
 bert-base-cased & 28996 & 20935 & 0.721996 & 14970 & 0.715070 & 0.516278 & 0.053544 \\
 flaubert\_flaubert\_small\_cased & 68728 & 37985 & 0.552686 & 2334 & 0.061445 & 0.033960 & 0.008348 \\
 xlnet-base-cased & 31997 & 22748 & 0.710942 & 17236 & 0.757693 & 0.538676 & 0.061649 \\
 cl-tohoku\_bert-base-japanese & 31999 & 1674 & 0.052314 & 790 & 0.471924 & 0.024688 & 0.002826 \\
\bottomrule
\end{tabular}
\caption{This very approximate morpheme list (more accurately, word surface forms and affixes) has 279,585 entries (words from CSW19, individual letters, and affixes from a few sources). The vast majority of those entries come from CSW19 -- so most of the entries are words.}
\label{tab:morpheme_csw19}
\end{table}

Table~\ref{tab:most_common_cleaned_tokens} shows common cleaned tokens and their file inclusion. Table~\ref{tab:common_nouns_csw19} shows common tokens that matched words categorized as nouns in CSW19 and their file inclusion.

\begin{table}
\centering
\tiny
\begin{tabular}{lrlrlrlrlrlrlrlrlr}
\midrule
q & 38 & ern & 37 & ill & 37 & home & 37 & day & 37 & dr & 37 & you & 37 & top & 37 & bra & 37 \\
 o & 38 & est & 37 & ng & 37 & gar & 37 & men & 37 & ser & 37 & run & 37 & bill & 37 & mp & 37 \\
 u & 38 & pan & 37 & mac & 37 & ah & 37 & de & 37 & ot & 37 & louis & 37 & app & 37 & bay & 37 \\
 c & 38 & af & 37 & ther & 37 & or & 37 & to & 37 & sky & 37 & saint & 37 & min & 37 & ve & 37 \\
 g & 38 & south & 37 & west & 37 & mon & 37 & son & 37 & ib & 37 & oc & 37 & ex & 37 & lor & 37 \\
 j & 38 & ine & 37 & master & 37 & ii & 37 & by & 37 & band & 37 & one & 37 & ul & 37 & mu & 37 \\
 a & 38 & hen & 37 & da & 37 & os & 37 & fire & 37 & fl & 37 & gen & 37 & oh & 37 & ph & 37 \\
 m & 38 & vol & 37 & om & 37 & rich & 37 & per & 37 & id & 37 & are & 37 & at & 37 & god & 37 \\
 i & 38 & king & 37 & las & 37 & pre & 37 & ev & 37 & em & 37 & ce & 37 & am & 37 & how & 37 \\
 l & 38 & ur & 37 & bus & 37 & ri & 37 & er & 37 & dan & 37 & tour & 37 & vi & 37 & mm & 37 \\
 p & 38 & rom & 37 & ros & 37 & th & 37 & ut & 37 & pr & 37 & ann & 37 & au & 37 & bit & 37 \\
 d & 38 & ca & 37 & en & 37 & ac & 37 & ky & 37 & lin & 37 & av & 37 & im & 37 & st & 37 \\
 r & 38 & vel & 37 & ang & 37 & free & 37 & pu & 37 & ju & 37 & up & 37 & du & 37 & ton & 37 \\
 sep & 38 & cr & 37 & ron & 37 & tom & 37 & berg & 37 & tor & 37 & ard & 37 & gan & 37 & best & 37 \\
 t & 38 & ling & 37 & sa & 37 & sm & 37 & ren & 37 & grand & 37 & lam & 37 & good & 37 & alex & 37 \\
 f & 38 & air & 37 & hot & 37 & ed & 37 & ben & 37 & big & 37 & ost & 37 & aa & 37 & black & 37 \\
 e & 38 & fr & 37 & no & 37 & su & 37 & ha & 37 & ze & 37 & ba & 37 & power & 37 & ny & 37 \\
 x & 38 & ab & 37 & ver & 37 & ja & 37 & of & 37 & fo & 37 & open & 37 & art & 37 & vin & 37 \\
 z & 38 & ep & 37 & camp & 37 & let & 37 & bel & 37 & be & 37 & um & 37 & tu & 37 & sch & 37 \\
 h & 38 & lt & 37 & bas & 37 & ni & 37 & li & 37 & pi & 37 & me & 37 & di & 37 & max & 37 \\
 b & 38 & yo & 37 & mr & 37 & mark & 37 & red & 37 & may & 37 & on & 37 & ta & 37 & do & 37 \\
 pad & 38 & river & 37 & hi & 37 & end & 37 & ent & 37 & ye & 37 & ob & 37 & ger & 37 & ry & 37 \\
 w & 38 & sh & 37 & ow & 37 & wa & 37 & ra & 37 & radio & 37 & ol & 37 & road & 37 & hall & 37 \\
 v & 38 & del & 37 & gin & 37 & san & 37 & over & 37 & ig & 37 & ong & 37 & jean & 37 & pe & 37 \\
 n & 38 & sp & 37 & ale & 37 & off & 37 & ler & 37 & ho & 37 & bell & 37 & the & 37 & il & 37 \\
 unk & 38 & ak & 37 & jan & 37 & will & 37 & si & 37 & ide & 37 & lu & 37 & ro & 37 & ell & 37 \\
 s & 38 & ice & 37 & rock & 37 & ft & 37 & frank & 37 & med & 37 & lim & 37 & le & 37 & con & 37 \\
 y & 38 & sam & 37 & boy & 37 & re & 37 & ste & 37 & hr & 37 & al & 37 & my & 37 & ci & 37 \\
 k & 38 & web & 37 & gr & 37 & play & 37 & ate & 37 & our & 37 & so & 37 & don & 37 & pa & 37 \\
 iv & 37 & cl & 37 & mo & 37 & ad & 37 & car & 37 & lo & 37 & bl & 37 & cy & 37 & fu & 37 \\
\bottomrule
\end{tabular}
\caption{Some of the most common cleaned (lower-cased, stripped of all non-Latin-characters and punctuation) tokens (569,810 tokens) and how many files they occur in.}
\label{tab:most_common_cleaned_tokens}
\end{table}

\begin{table}
    \centering
    \tiny
    \begin{tabular}{lrlrlrlrlrlrlrlrlr}
\midrule
ell & 37 & no & 37 & ash & 37 & ide & 37 & louis & 37 & arc & 36 & gal & 36 & gu & 36 & city & 35 \\
 grand & 37 & al & 37 & mon & 37 & bi & 37 & bo & 37 & martin & 36 & chi & 36 & real & 36 & mis & 35 \\
 un & 37 & jack & 37 & bus & 37 & po & 37 & bel & 37 & rem & 36 & mary & 36 & ens & 36 & tai & 35 \\
 good & 37 & pa & 37 & ben & 37 & ish & 37 & jean & 37 & ami & 36 & mini & 36 & town & 35 & old & 35 \\
 and & 37 & re & 37 & vin & 37 & ko & 37 & day & 37 & nu & 36 & ani & 36 & doc & 35 & yu & 35 \\
 ley & 37 & are & 37 & les & 37 & sol & 37 & san & 37 & zen & 36 & reg & 36 & than & 35 & bio & 35 \\
 op & 37 & ut & 37 & gi & 37 & king & 37 & pro & 37 & mas & 36 & cha & 36 & roman & 35 & alt & 35 \\
 car & 37 & now & 37 & lex & 37 & ate & 37 & hall & 37 & tel & 36 & ain & 36 & oy & 35 & org & 35 \\
 ab & 37 & one & 37 & app & 37 & en & 37 & es & 37 & prof & 36 & lab & 36 & tech & 35 & tex & 35 \\
 radio & 37 & me & 37 & river & 37 & how & 37 & ee & 37 & bank & 36 & bad & 36 & kat & 35 & kor & 35 \\
 son & 37 & pi & 37 & ta & 37 & fa & 37 & ill & 37 & euro & 36 & bro & 36 & iso & 35 & tam & 35 \\
 te & 37 & ou & 37 & at & 37 & john & 37 & fe & 37 & dom & 36 & sen & 36 & hon & 35 & sum & 35 \\
 if & 37 & mi & 37 & west & 37 & ling & 37 & ba & 37 & var & 36 & nat & 36 & ras & 35 & ion & 35 \\
 go & 37 & est & 37 & si & 37 & an & 37 & bas & 37 & mor & 36 & get & 36 & pic & 35 & gran & 35 \\
 io & 37 & ki & 37 & vol & 37 & as & 37 & dan & 37 & pol & 36 & mid & 36 & sel & 35 & ana & 35 \\
 ing & 37 & ai & 37 & aa & 37 & ay & 37 & ar & 37 & hard & 36 & bal & 36 & nan & 35 & mil & 35 \\
 rom & 37 & id & 37 & om & 37 & mo & 37 & william & 37 & mel & 36 & lis & 36 & iron & 35 & ya & 35 \\
 ag & 37 & ob & 37 & la & 37 & air & 37 & di & 37 & jun & 36 & ti & 36 & system & 35 & lit & 35 \\
 las & 37 & she & 37 & by & 37 & cal & 37 & so & 37 & night & 36 & mal & 36 & ers & 35 & hart & 35 \\
 za & 37 & ox & 37 & or & 37 & mac & 37 & del & 37 & world & 36 & mus & 36 & ban & 35 & mir & 35 \\
 ad & 37 & ann & 37 & ale & 37 & ant & 37 & ser & 37 & lar & 36 & cel & 36 & lev & 35 & mes & 35 \\
 he & 37 & men & 37 & tor & 37 & kin & 37 & rum & 36 & sim & 36 & vis & 36 & public & 35 & ala & 35 \\
 ky & 37 & ef & 37 & man & 37 & art & 37 & human & 36 & zo & 36 & col & 36 & hin & 35 & xi & 35 \\
 ton & 37 & ard & 37 & do & 37 & os & 37 & cent & 36 & met & 36 & anti & 36 & gon & 35 & jin & 35 \\
 all & 37 & li & 37 & berg & 37 & south & 37 & cor & 36 & ama & 36 & ore & 36 & eth & 35 & monte & 35 \\
 master & 37 & em & 37 & med & 37 & pe & 37 & res & 36 & pin & 36 & vas & 36 & wolf & 35 & nam & 35 \\
 bor & 37 & oo & 37 & bra & 37 & da & 37 & wo & 36 & sec & 36 & micro & 36 & raw & 35 & modern & 35 \\
 ed & 37 & with & 37 & el & 37 & ous & 37 & gas & 36 & ten & 36 & sal & 36 & gor & 35 & pac & 35 \\
 ger & 37 & road & 37 & paul & 37 & od & 37 & carl & 36 & lux & 36 & ess & 36 & bull & 35 & extra & 35 \\
 gold & 37 & jo & 37 & ma & 37 & mu & 37 & pas & 36 & eng & 36 & ultra & 36 & dal & 35 & song & 35 \\
\bottomrule
\end{tabular}
    \caption{Some of the most common cleaned tokens that were categorized as nouns in CSW19, and the number of files they occurred in.}
    \label{tab:common_nouns_csw19}
\end{table}

Common function word tokens are shown in Table~\ref{tab:function_words}.

\begin{table}
\centering
\resizebox{\textwidth}{!}{
\begin{tabular}{lrlrlrlrlrlrlrlrlr}
\midrule
i & 38 & he & 37 & than & 35 & most & 28 & which & 22 & whom & 19 & themselves & 18 & anywhere & 17 & anyway & 14 \\
 a & 38 & of & 37 & him & 35 & few & 28 & were & 22 & inside & 19 & anything & 18 & myself & 17 & nobody & 14 \\
 an & 37 & you & 37 & under & 35 & etc & 28 & though & 22 & upon & 19 & elsewhere & 18 & whenever & 17 & thru & 13 \\
 ever & 37 & us & 37 & his & 35 & need & 28 & once & 22 & something & 19 & whatever & 18 & latter & 17 & sometime & 13 \\
 me & 37 & one & 37 & first & 34 & where & 28 & else & 22 & beyond & 19 & although & 18 & ours & 16 & whereby & 13 \\
 over & 37 & it & 37 & what & 34 & there & 28 & both & 21 & hers & 18 & almost & 18 & none & 16 & herein & 12 \\
 if & 37 & eg & 37 & had & 34 & mine & 27 & many & 21 & everything & 18 & those & 18 & afterwards & 16 & wherever & 12 \\
 by & 37 & with & 37 & any & 34 & via & 27 & these & 21 & until & 18 & several & 18 & moreover & 16 & thereof & 12 \\
 is & 37 & the & 37 & who & 34 & near & 26 & could & 21 & together & 18 & otherwise & 18 & besides & 16 & whoever & 12 \\
 may & 37 & at & 37 & even & 33 & former & 26 & their & 21 & during & 18 & outside & 18 & thereby & 16 & wherein & 11 \\
 ie & 37 & she & 37 & other & 33 & this & 26 & would & 21 & while & 18 & everyone & 18 & nevertheless & 16 & yours & 11 \\
 out & 37 & re & 37 & your & 33 & about & 26 & above & 21 & being & 18 & shall & 18 & somewhere & 16 & yourselves & 11 \\
 or & 37 & well & 37 & last & 32 & same & 26 & done & 21 & might & 18 & often & 18 & namely & 16 & beforehand & 10 \\
 our & 37 & can & 37 & after & 32 & such & 25 & behind & 21 & against & 18 & himself & 18 & neither & 16 & theirs & 9 \\
 all & 37 & my & 37 & some & 32 & only & 25 & before & 21 & across & 18 & became & 18 & dare & 16 & hereby & 9 \\
 how & 37 & we & 37 & still & 32 & when & 25 & does & 21 & mostly & 18 & others & 18 & ourselves & 16 & thence & 8 \\
 so & 37 & do & 37 & less & 32 & again & 25 & below & 20 & somewhat & 18 & itself & 18 & despite & 16 & therein & 8 \\
 and & 37 & not & 37 & from & 31 & very & 25 & used & 20 & perhaps & 18 & whose & 18 & yourself & 16 & hereafter & 7 \\
 off & 37 & to & 37 & never & 30 & much & 25 & between & 20 & least & 18 & therefore & 18 & whereas & 16 & whence & 6 \\
 in & 37 & be & 37 & every & 30 & into & 24 & around & 20 & already & 18 & sometimes & 18 & thereafter & 16 & whereupon & 5 \\
 nor & 37 & as & 37 & them & 30 & been & 24 & except & 20 & enough & 18 & alone & 18 & meanwhile & 16 & anyhow & 3 \\
 her & 37 & for & 37 & more & 30 & they & 24 & why & 20 & either & 18 & each & 18 & everywhere & 16 & noone & 2 \\
 up & 37 & now & 37 & second & 30 & among & 24 & further & 20 & someone & 18 & always & 18 & furthermore & 16 & heretofore & 2 \\
 top & 37 & will & 37 & then & 29 & yet & 23 & whole & 20 & instead & 18 & without & 18 & hence & 16 & herewith & 2 \\
 no & 37 & down & 36 & here & 29 & have & 23 & should & 19 & indeed & 18 & another & 18 & somehow & 16 & thereupon & 2 \\
 on & 37 & but & 36 & its & 29 & onto & 23 & since & 19 & anyone & 18 & rather & 18 & amongst & 15 & whither & 2 \\
 per & 37 & has & 36 & too & 29 & through & 23 & nothing & 19 & cannot & 18 & within & 18 & formerly & 14 & latterly & 2 \\
 am & 37 & that & 36 & next & 29 & third & 23 & thus & 19 & throughout & 18 & however & 18 & nowhere & 14 & thereabouts & 2 \\
 was & 37 & lot & 35 & yes & 28 & along & 22 & lots & 19 & toward & 18 & whether & 18 & ought & 14 & oftentimes & 2 \\
 are & 37 & must & 35 & did & 28 & also & 22 & because & 19 & towards & 18 & herself & 17 & beside & 14 & hereinafter & 2 \\
\bottomrule
\end{tabular}}
\caption{Most of the function words that occurred amongst the cleaned tokens, and the number of files they occurred in, ordered from most to least common. The four least common function words are not shown, but they are (``thereon'', 2), (``hereupon'', 1), (``hereabouts'', 1), and (``whereafter'', 1).}
\label{tab:function_words}
\end{table}

\begin{table}
    \centering
    \tiny
    \begin{tabular}{lrlrlrlrlrlrlrlrlr}
\midrule
West & 25 & Fe & 25 & Man & 25 & Per & 25 & Ro & 25 & Sub & 25 & Sol & 25 & Mac & 25 & Plan & 24 \\
 Lou & 25 & New & 25 & Lin & 25 & Win & 25 & At & 25 & For & 25 & Su & 25 & Mo & 25 & Ban & 24 \\
 Tom & 25 & Go & 25 & Ch & 25 & Red & 25 & Il & 25 & World & 25 & My & 25 & Off & 25 & Es & 24 \\
 Val & 25 & Paul & 25 & Bell & 25 & Cal & 25 & We & 25 & Top & 25 & Web & 25 & God & 25 & Cho & 24 \\
 En & 25 & No & 25 & Po & 25 & Gen & 25 & On & 25 & Bill & 25 & High & 25 & Rod & 25 & Gal & 24 \\
 Reg & 25 & War & 25 & Mad & 25 & Tour & 25 & Nor & 25 & Be & 25 & Mal & 25 & William & 25 & Micro & 24 \\
 Dr & 25 & Up & 25 & Rock & 25 & Ph & 25 & Mart & 25 & To & 25 & Ray & 25 & Ke & 25 & Set & 24 \\
 May & 25 & Pre & 25 & Ha & 25 & De & 25 & Master & 25 & Ka & 25 & Mr & 25 & Or & 25 & Wolf & 24 \\
 Open & 25 & Free & 25 & Sc & 25 & Bo & 25 & Em & 25 & He & 25 & Sa & 25 & Cha & 24 & Bang & 24 \\
 Lim & 25 & Phil & 25 & Radio & 25 & Day & 25 & Se & 25 & It & 25 & Land & 25 & Ter & 24 & Bro & 24 \\
 Bus & 25 & Ed & 25 & Min & 25 & Am & 25 & Louis & 25 & Long & 25 & Big & 25 & By & 24 & Dam & 24 \\
 Lu & 25 & Sur & 25 & Mark & 25 & Frank & 25 & Rich & 25 & Pat & 25 & Will & 25 & Monte & 24 & Berg & 24 \\
 Saint & 25 & Angel & 25 & Blu & 25 & Si & 25 & Al & 25 & Do & 25 & Arm & 25 & Job & 24 & Kit & 24 \\
 Ser & 25 & Port & 25 & Can & 25 & Un & 25 & You & 25 & Roy & 25 & Trans & 25 & Well & 24 & Martin & 24 \\
 Fire & 25 & Je & 25 & And & 25 & Mon & 25 & King & 25 & Luc & 25 & Tor & 25 & Del & 24 & Ant & 24 \\
 Mar & 25 & Jack & 25 & One & 25 & Stand & 25 & Air & 25 & Ju & 25 & Euro & 25 & Gil & 24 & Ali & 24 \\
 Ann & 25 & La & 25 & Le & 25 & Christ & 25 & Carl & 25 & San & 25 & Op & 25 & Mc & 24 & Team & 24 \\
 Her & 25 & In & 25 & As & 25 & Pro & 25 & Art & 25 & Band & 25 & So & 25 & Lang & 24 & Sp & 24 \\
 Don & 25 & Bay & 25 & Pan & 25 & Sam & 25 & Di & 25 & Sal & 25 & Con & 25 & Francis & 24 & Pay & 24 \\
 Tu & 25 & Bel & 25 & Power & 25 & Alex & 25 & Ex & 25 & Green & 25 & Pol & 25 & Dar & 24 & How & 24 \\
 Jean & 25 & Mel & 25 & Ben & 25 & An & 25 & Col & 25 & Mi & 25 & Bon & 25 & Road & 24 & Hard & 24 \\
 Gold & 25 & Co & 25 & Ad & 25 & Bar & 25 & Pet & 25 & Hal & 25 & Mid & 25 & Let & 24 & Chi & 24 \\
 Inter & 25 & Na & 25 & South & 25 & Camp & 25 & Out & 25 & Les & 25 & Cat & 25 & Jacob & 24 & Oh & 24 \\
 Run & 25 & Sun & 25 & Pop & 25 & Good & 25 & She & 25 & Hot & 25 & Cap & 25 & Spring & 24 & Vic & 24 \\
 Super & 25 & Jo & 25 & Ba & 25 & Grand & 25 & Pi & 25 & Dan & 25 & Tim & 25 & Res & 24 & Bern & 24 \\
 Te & 25 & North & 25 & St & 25 & Boy & 25 & Vol & 25 & Pa & 25 & Mat & 25 & Che & 24 & Of & 24 \\
 Mont & 25 & Mill & 25 & Net & 25 & All & 25 & Car & 25 & Li & 25 & Fi & 25 & Tam & 24 & Thor & 24 \\
 Lo & 25 & Fr & 25 & Ta & 25 & Ra & 25 & Ma & 25 & The & 25 & Black & 25 & Sky & 24 & Bab & 24 \\
 Da & 25 & Max & 25 & John & 25 & Re & 25 & Jan & 25 & Du & 25 & El & 25 & Mus & 24 & Bi & 24 \\
 Me & 25 & Hall & 25 & Star & 25 & Jon & 25 & Tri & 25 & Hi & 25 & Rob & 25 & Met & 24 & Sar & 24 \\
\bottomrule
\end{tabular}
    \caption{Some of the most common clean tokens that have an initial first letter followed by all lower letters, and the number of files they occur in.}
    \label{tab:common_proper_noun_candidates}
\end{table}

\begin{figure}
    \centering
    \includegraphics[width=0.5\textwidth]{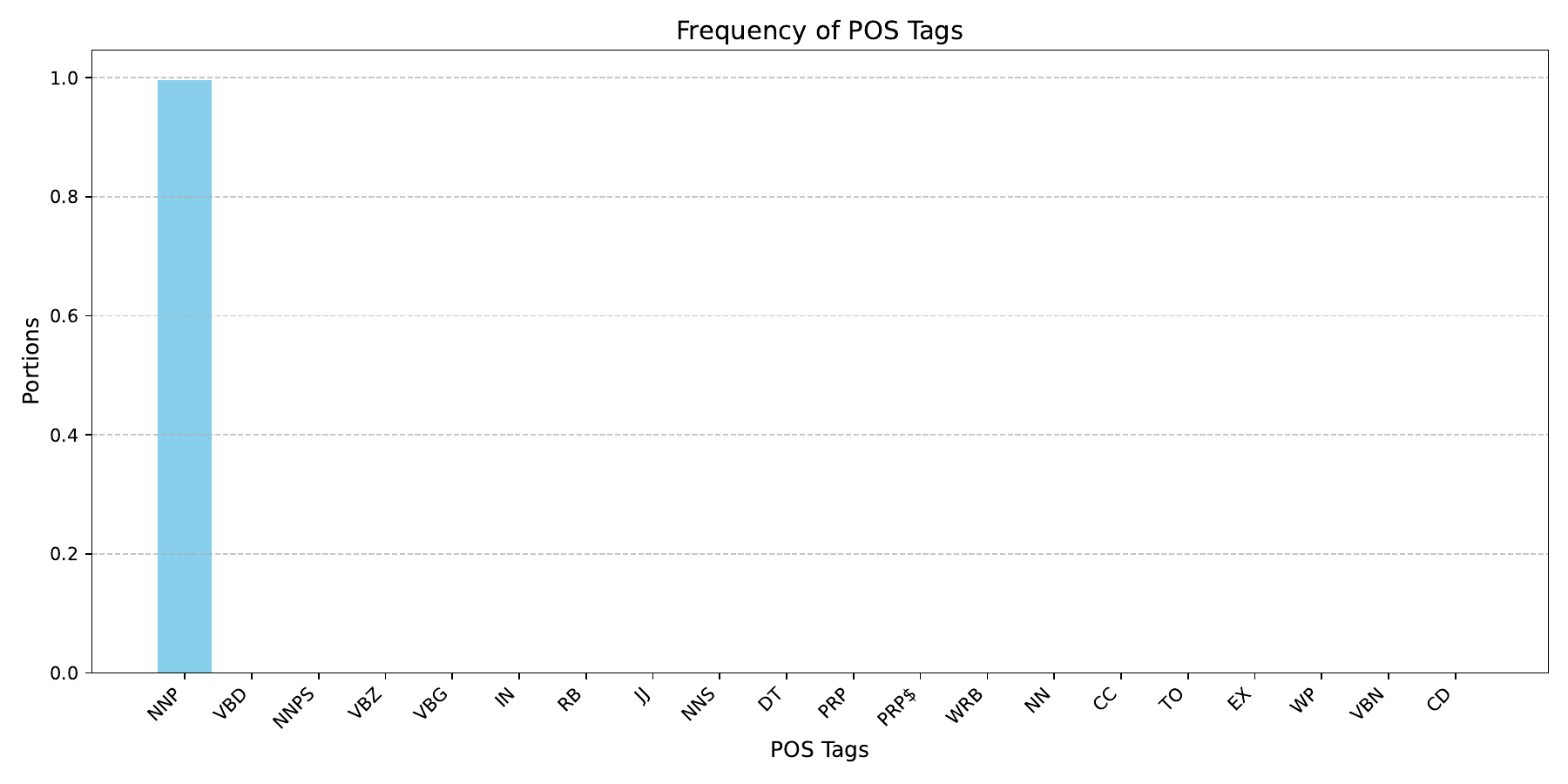}
    \includegraphics[width=0.5\columnwidth]{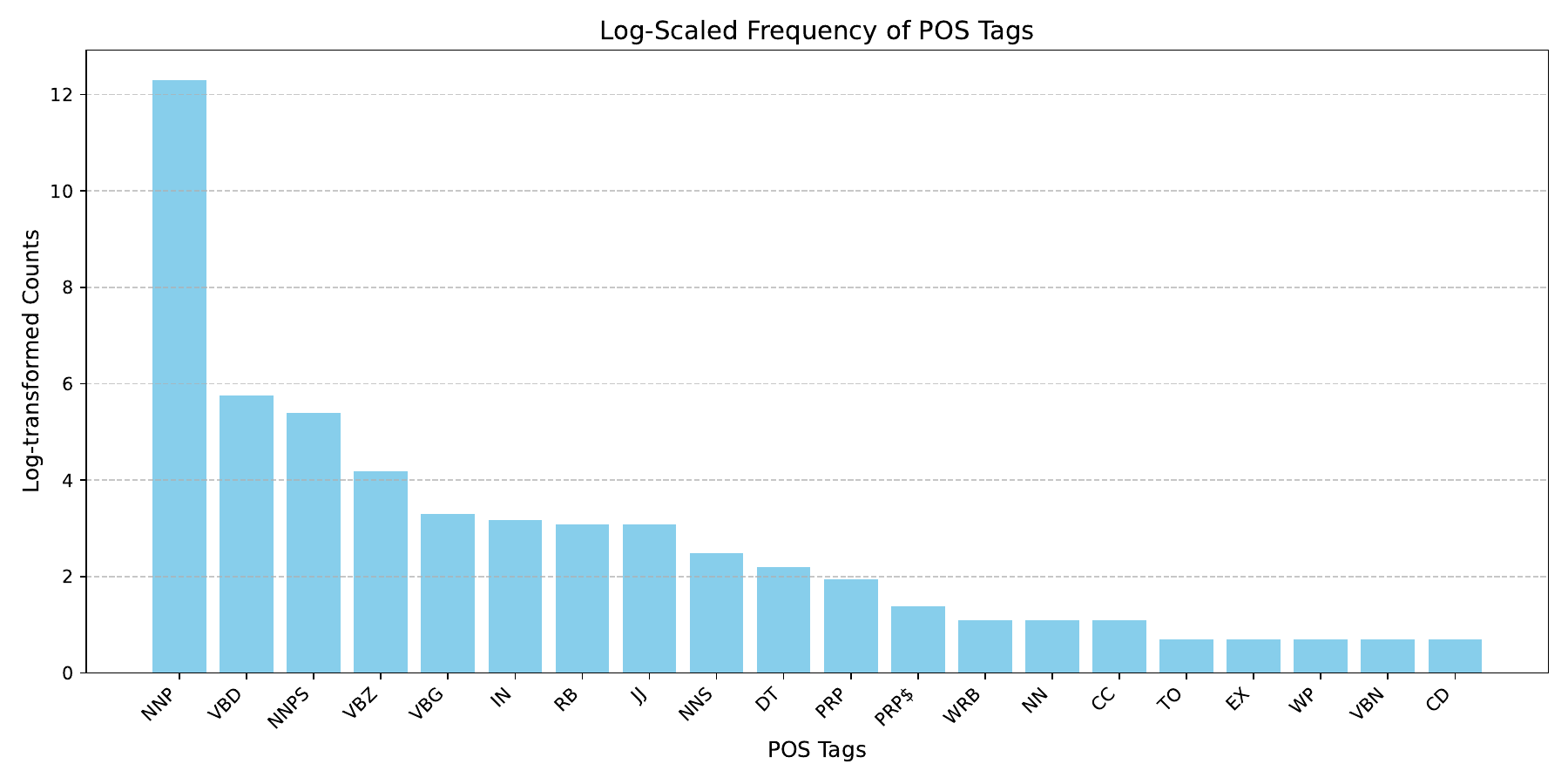}
    \caption{There are 687,085 unique clean cased tokens, and 222,281 of those fall in the proper-noun-candidate category (identified using a very simple and imperfect heuristic, an initial capital letter). Of those 222,281 tokens, \num{99.67203674628061}\% were labeled by nltk as NNP, singular proper nouns. In the upper chart, we show the proportions of each POS category. In the lower chart, we show the log(count + 1) so the smaller categories are more visible.}
    \label{fig:proper_noun_pos}
\end{figure}

Figure~\ref{fig:proper_noun_pos} shows POS tags for tokens in the proper-noun-candidate category (the tokens in Table~\ref{tab:common_proper_noun_candidates}).

\begin{figure*}
    \centering
    \includegraphics[width=\textwidth]{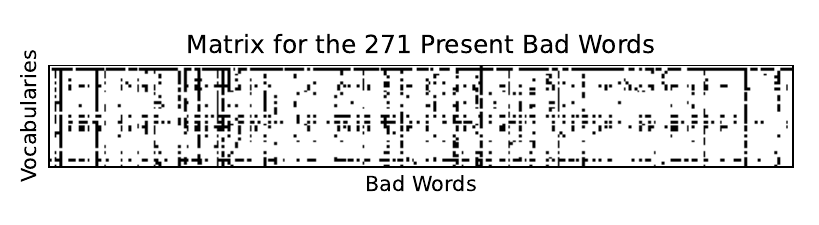}
    \caption{Out of the 965 bad words in the largest bad word list, 271 words were in at least one of the 38 vocabulary files from Hugging Face and 96 words were in at least 5 of the files. This visualization shows presence of a word in a vocabulary file as black and absence as white. From the apparent crosshatching, we can see that some vocabulary files feature more bad words, and some bad words are featured in more vocabulary files; overall, it's not uncommon for some words from the list to show up in a file.}
    \label{fig:heatmap}
\end{figure*}

\begin{figure}
    \centering
    \includegraphics[width=0.5\textwidth]{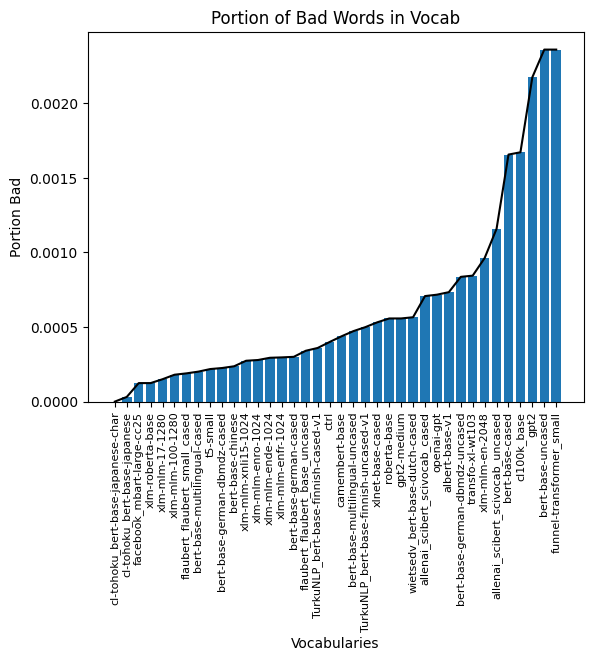}
    \caption{This visualization shows what portion of bad words (from the largest list with 965 words) occur in each of the Hugging Face vocabulary files, based on the approximate size of the vocabulary, if appending whitespace is ignored and subsequent duplicates are thrown out. However note that case is still present. The mean is 6.32e-04. While this is a very simple interpretation of the portion of bad words of the whole, there are other reasonable interpretations, such as discarding any tokens of length 1 from the vocabulary size.}
    \label{fig:portions}
\end{figure}

Table~\ref{tab:total_bad} and Figures~\ref{fig:heatmap},~\ref{fig:portions} show the occurrence of bad words in the vocabulary files.

\begin{table}[t]  % Use the table environment
    \centering
    \small
\begin{tabular}{lrrr}
\toprule
Vocab & Bad Words & Vocab Size & Portion Bad \\
\midrule
bert-base-chinese & 5 & 21127 & 2.37e-04 \\
cl-tohoku\_bert-base-japanese & 1 & 31999 & 3.13e-05 \\
transfo-xl-wt103 & 226 & 267734 & 8.44e-04 \\
bert-base-german-dbmdz-uncased & 26 & 31101 & 8.36e-04 \\
xlm-mlm-17-1280 & 30 & 199999 & 1.50e-04 \\
xlm-mlm-xnli15-1024 & 26 & 94999 & 2.74e-04 \\
bert-base-multilingual-cased & 24 & 119547 & 2.01e-04 \\
xlm-mlm-100-1280 & 36 & 199999 & 1.80e-04 \\
bert-base-uncased & 72 & 30522 & 2.36e-03 \\
bert-base-multilingual-uncased & 50 & 105878 & 4.72e-04 \\
TurkuNLP\_bert-base-finnish-uncased-v1 & 25 & 50101 & 4.99e-04 \\
roberta-base & 28 & 50260 & 5.57e-04 \\
xlm-mlm-enro-1024 & 18 & 64592 & 2.79e-04 \\
flaubert\_flaubert\_small\_cased & 13 & 68728 & 1.89e-04 \\
openai-gpt & 29 & 40477 & 7.16e-04 \\
allenai\_scibert\_scivocab\_uncased & 36 & 31089 & 1.16e-03 \\
xlm-mlm-en-2048 & 29 & 30145 & 9.62e-04 \\
xlm-mlm-ende-1024 & 19 & 64699 & 2.94e-04 \\
t5-small & 7 & 32100 & 2.18e-04 \\
gpt2 & 90 & 41403 & 2.17e-03 \\
camembert-base & 14 & 32004 & 4.37e-04 \\
ctrl & 98 & 245470 & 3.99e-04 \\
TurkuNLP\_bert-base-finnish-cased-v1 & 18 & 50104 & 3.59e-04 \\
funnel-transformer\_small & 72 & 30522 & 2.36e-03 \\
bert-base-cased & 48 & 28996 & 1.66e-03 \\
wietsedv\_bert-base-dutch-cased & 17 & 30073 & 5.65e-04 \\
allenai\_scibert\_scivocab\_cased & 22 & 31115 & 7.07e-04 \\
bert-base-german-dbmdz-cased & 7 & 31101 & 2.25e-04 \\
gpt2-medium & 28 & 50252 & 5.57e-04 \\
xlnet-base-cased & 17 & 31997 & 5.31e-04 \\
albert-base-v1 & 22 & 30000 & 7.33e-04 \\
cl-tohoku\_bert-base-japanese-char & 0 & 3999 & 0.00e+00 \\
facebook\_mbart-large-cc25 & 31 & 250025 & 1.24e-04 \\
bert-base-german-cased & 9 & 29999 & 3.00e-04 \\
xlm-mlm-enfr-1024 & 19 & 64139 & 2.96e-04 \\
cl100k\_base & 127 & 76014 & 1.67e-03 \\
flaubert\_flaubert\_base\_uncased & 23 & 67541 & 3.41e-04 \\
xlm-roberta-base & 31 & 250000 & 1.24e-04 \\
\bottomrule
\end{tabular}
    \caption{This table shows how many bad words (from the largest list of 965 words) occur in each of the Hugging Face vocabulary files. They also show the approximate size of the vocabulary, if appending whitespace is ignored and subsequent duplicates are thrown out. However note that case is still present. While this is a simple interpretation of the portion of bad words of the whole, there are other reasonable interpretations, such as discarding any tokens of length 1 from the vocabulary size.}
    \label{tab:total_bad}
\end{table}

\appendixsection{GPT-4o token length}
\label{sec:gpt4otokenlength}

\begin{figure}
    \centering
    \includegraphics[width=0.7\columnwidth]{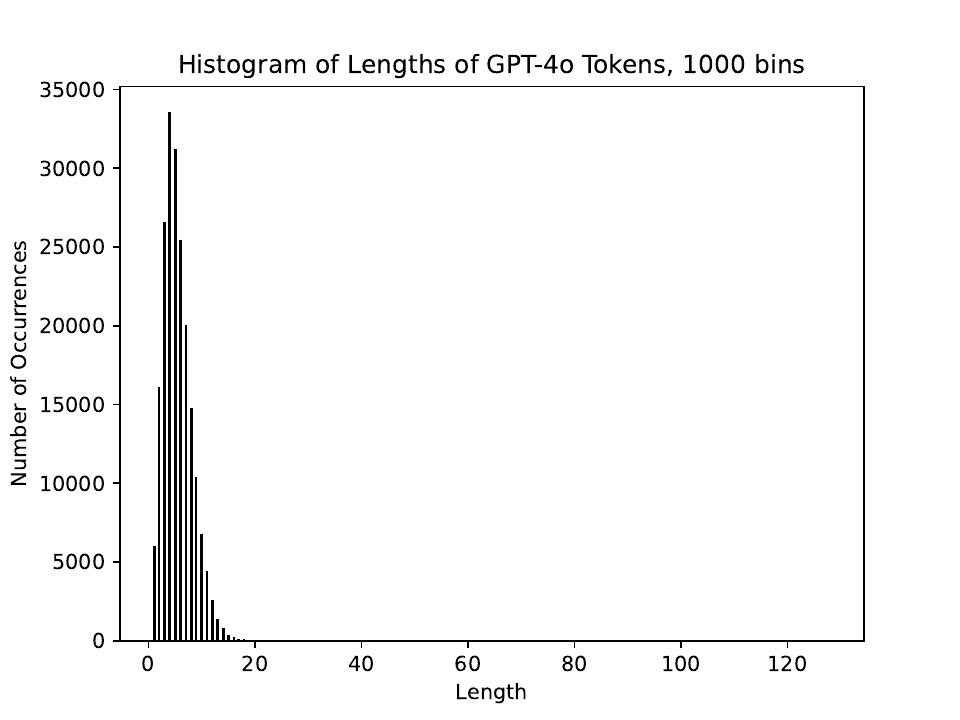}
    \includegraphics[width=0.7\columnwidth]{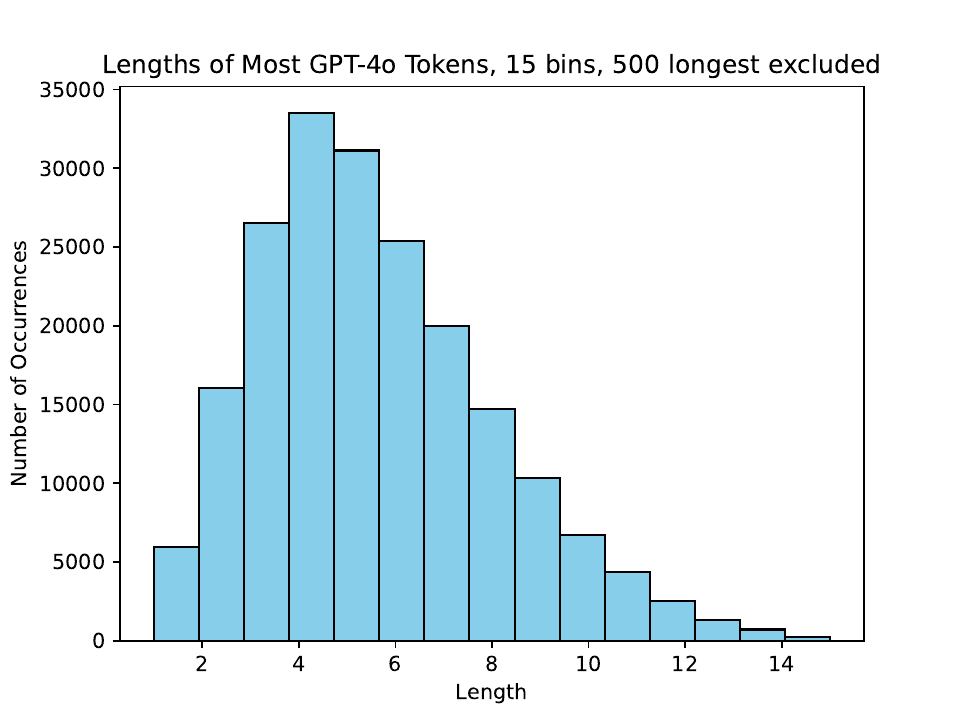}
    \caption{Using tiktoken, we looked at the lengths of the tokens in the file o200k\_base for GPT-4o. In the second chart, we excluded the 500 longest tokens.}
    \label{fig:length_histogram_gpt4o}
\end{figure}

Figure~\ref{fig:length_histogram_gpt4o} shows the lengths of the GPT-4o tokens. We think the shape of the distribution suggests tokenization being applied once across all relevant training data, rather than in processes segregated e.g. by language. Otherwise we think you would see multiple distributions overlaid based on how much information each language typically stores in one unit of length. We note that the long Chinese tokens in Figure~\ref{fig:longest_chinese_tokens_gpt4o} look to be about the length of slightly less common English words. Many of the GPT-4o tokens are English subword or word length, but that escalates suddenly with tokens that are long strings of repeated characters (Figure~\ref{fig:long_tokens_gpt4o}). The longest non-repeating tokens are include abcdefghijklmnopqrstuvwxyz, ABCDEFGHIJKLMNOPQRSTUVWXYZ, verantwoordelijkheid, telecommunications, .onreadystatechange, significativamente, and disproportionately, some with leading spaces. There seem to be many emojis amongst the tokens.

\appendixsection{Extispicy}

Figures~\ref{fig:run},~\ref{fig:the} show the gnogeographic maps for two additional tokens. Figure~\ref{fig:teacher} shows the un-annotated clusters for `` teacher''.

\appendixsection{Limitations}
\label{sec:limitations}

We discuss limitations for this work, broken out into subcategories.

\appendixsection{Vocabulary Limitations}
\label{sec:vocablimitations}

Sometimes identical strings appear multiple times as different tokens. This seems likely to be due to encoding. The collision could be happening at a number of levels, e.g. as the result of our cleaning processes, or it could be lower down, perhaps due to underlying choices in UTF-8 to accommodate relic systems.\footnote{If it is the latter, that could be a place where supplementary approximations of supradiegetic linguistic information could be helpful in facilitating the merger of tokens~\cite{zimmerman2024blind}; tokenizers can have built-in merging functionality, depending on whether the algorithm is working at the byte or character level, so perhaps information about the sounds and orthography of the tokens involved could augment those processes.} Depending on how the UTF-8 encoded character set is being handled, there could be scenarios where two or more different encodings would be be interpreted as the same character. Encoding is confusing and unwieldy; it seems like an area prone to human error (by the authors or others).

Relatedly, although many tokenizers work at the byte level, our analysis was at the character level. As aforementioned, the map from bytes to legible English characters was imperfect. For example, a token could be a series of bytes that cannot be parsed into a legible character without conjunction with another token. The token could still be perfectly useful to the model, but it would not render as a human-legible string without significant additional work. Such tokens seem unlikely to be directly relevant to the specific tokens discussed here, so we ignored any tokens that we couldn't read.

We focused on the types of tokens, not the tokens of tokens, but there is room for alternate interpretation as to what types of tokens means in future work, as we only occasionally and superficially resolved variants (future work could look at overlapping tokens, components of the same words, cased variations, misspellings and typos, word families, or tokens with the same referents). We're particularly interested in the number and distributions (since tokenization is deterministic but context-dependent) of ways a word could be constructed using tokens in the vocabulary and what relationship that has to model output and cognition.

In general, some of the tokenizers, model weights, or other parameters available through Hugging Face may be from different versions of the relevant models (than what is accessible through tiktoken, and than what is deployed in any particular context). From our perspective, this was indiscernible. We do not know whether any of these files are up to date with what is deployed now, and we do not know the details of how encoding was handled.

The vocabulary files across different sources are not necessarily identical. For example, the vocabulary file we got for GPT-2 from Hugging Face and the vocabulary file we got for GPT-2 from tiktoken do not appear to be exactly identical. 

Where exactly these vocabulary files end up in use is not in scope for this project. Some vocabularies are used by many models or families of models. The models themselves can then end up in a variety of contexts. We likewise were not specific about which models probably used which kinds of tokenization. Sometimes this information is available. For more in this line, see ~\cite{bostrom-durrett-2020-byte}. 

There are other models, and other tokens, that could be of concern, beyond what was included in this project.

Spacing, punctuation, and case are part of what gets tokenized, so tokens may have leading or trailing whitespace or punctuation, and each cased version of a letter is encountered as a unique symbol. A leading space is particularly common for tokens that correspond to surface word forms. We sometimes ignored appending whitespace, and sometimes left punctuation and case intact. We almost exclusively considered English words, and treated all vocabulary files as if they only contained English. We did not attempt to translate anything when carrying out tests, although we did look at the translations for some long Chinese tokens in GPT-4o~\cite{incident_729}.

\appendixsection{Extispicy Limitations}
\label{sec:extispicylimitations}
We looked at only a few words, and focused mostly on a couple words. We scrutinized ``bank'' most closely, although we looked at ``run'' and ``the'' as well. In addition, our corpus (of token instances that we passed into the model to create the gnogeographic maps) was relatively small and narrow.

If classical word embeddings work to locate word meanings, then it makes sense that word embeddings for individual words would work to locate polysemy, senses, and phrase architecture for that word. However, a caveat is that because we're applying a clustering algorithm -- UMAP --, we will \textit{always} find clusters, so it is unclear how to interpret the categorical relations involved. 

Obvious main clusters for two different words does not imply that those main clusters represent structured information that exists in an equivalent way within the model. Likewise, that two clusters exist within the same word does not mean the model treats those two categories identically. The clusters for a word with fewer utterances in the training data can have differently-meaningful clusters than a word with many utterances. The caveats are endless.

In the visualizations that we used to label the clusters in the gnogeographic maps, the context visible to the viewer (10 tokens on either side of the exemplar token) was smaller than the context used by the model (100 tokens on either side of the exemplar token). This decision was made for practical (legibility and computational) purposes. We reason that including the additional context would be very unlikely to negate any structure we inferred from the smaller context, although it could refine it, or surface additional structure.

From passing through 1,000 samples of a word versus 10,000 samples of the same word, we know that changing the number of samples -- providing more vectors reflecting contexts that show how the word can be used in the set we apply PCA and UMAP to in order to create the gnogeographic maps -- changes the resulting composition. This stands to reason, but since we witnessed it (e.g. observing the clusters for the word ``death'' with significantly more or fewer samples), it seems worth noting explicitly. For the gnogeographic maps included here, we used all instances in our corpus that had enough context around them (e.g., for ``bank'', 62,463 instances), but there is (1) more information in the vectors beyond what we explored (due to PCA, UMAP, and showing a subset of the context tokens in the visualization we used to interpret the clusters), (2) likely more relevant information in the model's gnogeography already, as it likely saw many more instances during training than we had in our corpus, and (3) in addition, we suppose that the model would learn more -- expand its gnogeography -- when exposed to more (varied) instances during training. Again, the direction of this limitation is that the model does (and could) know more than we captured here.

For each exemplar word we passed through the model, we used a leading space (`` bank'' vs. ``bank''). Although not a 1-1 match, this should more or less align with our vernacular conceptions of the word ``bank''. For convenience, throughout the paper we sometimes referred to this as the word or token ``bank'', but that is not exactly accurate. As a result of our methodology and how tokenization works, there are things we cannot see in our data that could meaningfully exist for the model, for example, any connections between ``bank'' and ``embankment'' (because ``embankment'' would not match the pattern we were looking for of ``bank'' with a leading space).

Since we applied PCA to the vector output by each model layer, there is some ambiguity as to how the resultant vectors relate to each other on a point-by-point basis. Reflection or rotation may have occurred relative to previous layers. These differences would not alter the internal relationships within a layer -- they maintain the absolute distances between all points in a layer. Due to the model's architecture, the residual layers mean that each attention layer takes the output of the former layer, modifies it, and adds it back to the output of the former layer, then normalizes the resulting combination. This acts as a constraint that limits how much each token can diverge from its former value in the previous layer. There is therefore inherently a degree of similarity between adjacent layers. Although this could be used to align the time-series data across the layers, the decomposition we’re doing on a layer by layer basis for visualization did not take this into account. Interestingly though, the fact that this similarity, or constraint on the divergence, exists is reflected in the structure we saw in our 3D visualizations. There seems to be a tendency for layers to align in a human-interpretable way, particularly where the degree of polysemy in the relevant token is high. Further, when we did see misalignments, they were usually reflections, which were obvious on visual inspection.

We don't think any of the aforementioned limitations would lead to structure that didn't exist for the model being created \textit{ex nihilo}, which is our primary concern. However, the level at which that structure exists can be ambiguous. In general, we think that the fact that such fine-grained information show up in our depiction of the model's inner state even given our limitations makes this initial exploration even more compelling.

A final, potentially grave, limitation for our methodology and analysis of the embedding space is that there are many more underlying details we could have engaged with more deeply. There are two particularly glaring regions where this limitation applies. For one, there is an emerging body of research focused on how specific model architecture (components, objectives) relates to the geometry of the embedding spaces and how the embedding space behaves in different layers of the model~\cite{wolfe-caliskan-2022-contrastive, fuster-baggetto2022findings}. For another, intertwined with the first region, is the choice of how to evaluate the quality of an embedding, or, more broadly, any representation made by the model. These judgments ought to sensibly depend on the expected geometry (of the latent space or other representational medium) as well as the intended task for the model and the type of representation being judged (e.g. different judgments rely on different structures; the Rubenstein and Goodenough 1965 dataset (RG65) uses words~\cite{rubenstein10.1145/365628.365657} whereas the SemEval-2017 Semantic Textual Similarity (STS) Benchmark uses sentences~\cite{cer2017semeval}). However, we mostly treated these aspects like black boxes: there is (1) the model architecture, which we generally assume has an overarching objective function of something like generating plausible tokens, except in a few cases such as CLIP~\cite{radford2021learning}, and there is (2) the embedding matrix (or matrices), which is basically a latent space capturing the meaning of the vocabulary for the model, and (3) there are the reported judgments of the model's performance or the quality of its learned representations (usually we boiled this down to whether the source we were looking at seemed to think the model was doing a ``good'' or ``bad'' job), all of which we treated as comparable enough across the models. This is a significant simplification, and one which we will likely peel back (and, it is worth acknowledging, potentially come to regret) in future work.

\appendixsection{General Limitations}

This project also took a long time, and we, imperfect and humble vessels, did not always create the clearest documentation as it progressed. For example, artifact names we thought would surely stand for all eternity (or at least, long enough) may now be somewhat inscrutable. Our focus and plans also shifted as the project progressed, meaning that some things we didn't think were important to write down at the time became more significant. We don't think these shortcomings significantly impair the subsequent information we're relaying, but it is possible there are some mistakes of recollection or interpretation. Our data and pipeline were not documented and organized in such a way as to make them easily reuseable. Importantly, due to unspecified technical malfunction (what we mean is, one of our computers broke) we can no longer directly access a non-trivial portion of our work (data, code). However, we do have quite a few artefacts saved to the cloud, which we are happy to share with any reasonable (generously construed!) interested parties.

This paper is almost exclusively English-focused and thus should not be expected to generalize in its particulars to other languages (or models in other languages). Significantly, the units and categorizations relied on in this paper, notably `words' and the parts of speech, are not implemented in the same way [as in English] cross-linguistically~\cite{rijkhoff2002verbs}.
 However, since English is the \textit{lingua franca} of much of computer science and significant portions of the internet~\cite{atari2023which}, it is likely to end up in LLMs. For that reason, and because it is the language most familiar to most of the authors of this paper, we think it is a reasonable place to start exploring this topic.

Although it stands to reason, it is not clear to us whether the performance of approaches similar in approach to BPE are in fact boosted by language-specific features, such as the relative morphological simplicity of English~\cite{avramidis-koehn-2008-enriching}. For example, imagine the tokenization process applied to a language with complicated rules of harmony that mean word forms undergo significant surface changes in each utterance. This seems like it would result in more, smaller word chunks being necessary in tokenization, and like it would make it more difficult for the model to consolidate information about word families within its embedding matrix. Future work could probe this topic.

Some of the common terms used to discuss LLMs can be ambiguous. For example, the meanings of ``pre-training'' and ``training'' can overlap, especially if fine-tuning also took place. Likewise, ``character-level'' can be defined in opposition to either smaller levels (the byte-level) or larger levels (subwords, words).\footnote{For example, compare usage in Liu \textit{et al.}, 2019~\cite{liu2019robertarobustlyoptimizedbert} vs. Huang \textit{et al.}, 2023~\cite{huang2023lexinvariantlanguagemodels}.} In the former case, a subword vocabulary could also be called a character-level vocabulary. We tried to interpret these terms appropriately when reading our sources, but we could have misunderstood.

We are, in general, not entirely sure how to meaningfully interpret some of the benchmarking results, especially head-to-head with the aspect of LLMs we find most remarkable, that is, that they can have strikingly good linguistic performance at the output level. When judgments of models are reported, they are often in terms of benchmarks of performance, comparison to human judgments, and similarity measures, sometimes at the output level, and sometimes throughout the model's layers. Separately, but relatedly, sometimes when looking at the benchmark dataset being used, it is unclear to us what the comparable human performance ought to be. This limitation regarding our uncertainty in interpreting comparisons between models is probably most relevant to our interpretation of the papers we read that report contextually impressive language representations or attempt something like grounding (such as ~\cite{radford2021learning, zhuang-etal-2024-lexicon, wolfe-caliskan-2022-contrastive,huang2023lexinvariantlanguagemodels}).

Parts or all of some of the relevant models are proprietary and not directly accessible to the general public. This adversely impacts the research around them. For example, Templeton \textit{et al.}, 2024 states, ``Claude 3 Sonnet is a proprietary model for both safety and competitive reasons. Some of the decisions in this publication reflect this''~\cite{templeton2024scaling}. Access even to the output of a model may be pay-walled (as is the case for use of GPT-4o above a certain threshold in ChatGPT).

Related to the previous point, research sometimes come from the proprietors of these models, which makes it subject to additional financial, political, and reputational pressures, which in turn makes it harder to verify. We criticized this with respect to a recent paper involving Replika in Zimmerman and Ruiz, 2024~\cite{zimmerman2024matters}, and in an ideal world,  we would have applied that outlook here. 
Significant issues with scientific publishing, including peer review as currently implemented, aside, there are reputational benefits to be had from looking like you're participating in it, which is what these companies are going for: ``the release-by-blogpost model'' reaps ``the benefits of mimicking scientific communication... without actually doing the work.'' Similarly, some invocations of the \textit{open source} lineage reap ``the benefits of libre culture — including associations of transparency and associated freedoms — without actually contributing to the commons''~\cite{liesenfeld2024rethinking}. However, though scrutiny is warranted, for this paper, we assumed disinterested good faith on behalf of all the sources we relied on, taking available sources as basically equal. 

We note with sadness that participating in the Generative AI ecosystem shores up the power of the organizations creating these models, many of which also profit from them (to the detriment, financial and otherwise, of many others), hoard resources related to them, and, relatedly, keep much about them secret, the latter of which is, of course, antithetical to science. Companies like Google engage in ``predatory practices of rent-seeking and subordination of the companies
and research institutions that participate in their CIS [corporate innovation systems]''~\cite{rikap2022big}.

J.W.Z., at least, believes it is the intention (POSIWID) of many companies -- Google, Meta, Amazon, OpenAI, and the like -- to increase the reliance of businesses, workers, and students (including children) on their AI products in as many contexts as possible, intentionally de-skilling individuals and organizations to secure their dependence, stably capturing them as renters (maybe, with a historical lens, as vassals). This strategic de-skilling, at the very least, when it comes to writing -- a key development in the deeply human ability to communicate via language -- is, in this author's opinion, predatory, unethical, and gross.

As an illustration of the previous points, paraphrasing Rikap and Lundvall, 2022: by 2019, Google had published 6,447 scientific papers, co-authored with almost 2,400 organizations. And yet, none of those organizations were among the top (100) co-owners of Google's patents, and pre-2017, Google had only shared 3 patents with a university (Stanford). Similarly, although Amazon owned over 10,000 patents by 2018, it only shared 13 (0.13\%) with other companies and \textit{none} with universities, even though Amazon frequently touts collaboration (e.g. ``Collaboration between Amazon and UC Berkeley advances AI and machine learning''~\cite{amazon_ucberkeley_ai_ml_collaboration}) with the scientific community and has co-authored papers with more than 750 organizations, academic institutions among them~\cite{rikap2022big}.

The limitations about access and resources are especially relevant as the research strategy hews to scaling, since only elite institutions, often run for profit and power, have the resources to fully pursue that line~\cite{liesenfeld2024rethinking}.

Participating in that ecosystem also means contributing to the environmental impact of Generative AI (``training a single moderately large model produces twice the American lifetime’s worth of CO2''~\cite{shumailov2024collapse}). We tried where possible to use data that already existed. For personal, philosophical, and environmental reasons, we also tried to argue from principals and small proof-of-concept approaches rather than full-scale experimentation where plausible.

In general, although we treat each model as an essentially discrete system, the boundaries between man and machine are porous~\cite{veselovsky2023artificialartificialartificialintelligence, liesenfeld2024rethinking, Davis2024GPTPoetry}. As a starting point, ``[h]uman labour can be involved at multiple points, from reinforcement learning datasets to crowd-sourced ratings''~\cite{liesenfeld2024rethinking}, not to mention the entire historial context that spawned the text these models are trained on. As a corollary, we should be \textit{extremely skeptical} of ``human out of the loop'' claims~\cite{veselovsky2023artificialartificialartificialintelligence, Davis2024GPTPoetry} (for more on this topic, see the Luddite~\cite{theluddite})! The boundaries between the individual and the collective are also poorly-defined. The medium of language only makes all of these boundaries fuzzier.

\begin{acknowledgments}
The authors wish to express their appreciation for conversations with and support from Juniper Lovato, Jonathan St. Onge, Aviral Chawla, Parisa Suchdev, Mohsen Ghasemizade, Desi Alexander, Ashley Dennis-Henderson, Michael Arnold, William Thompson, Gabriel Meyer-Lee, Robert Wolfe, Danbee Kim, Alice Patania, Josh Bongard, Kenneth C. Fan, Haim Dubossarsky, Alexa Woodward, Charlie Brooks, and participants and leaders of CEL (the Computational Ethics Lab) and SCRaPs (Student Complexity Research and Pizza Seminars), and ChatGPT and NotebookLM. Thanks to the people who created the lists, models, training data, and packages we relied on in this project. Finally, thanks to the speakers, participants, and organizers of the excellent June 2024 UQAM summer school, ``Understanding LLM Understanding'', moderated by Stevan Harnad.
The authors acknowledge financial support from 
MassMutual Center of Excellence in Complex Systems and Data Science, 
Google Open Source, 
and
The National Science Foundation award \#2242829 (J.W.Z., C.M.D., P.S.D).
Additionally, M. Z. Trujillo is supported by the Northeastern University Future Faculty Postdoctoral Fellowship Program.
\end{acknowledgments}

%\newpage
%If you would like to use \starttwocolumn below for references: Unfortunately, there is a known bug in the existing \starttwocolumn command when used without \newpage above it. If you find issues such as whole pages disappearing, add \newpage above \starttwocolumn as a temporary fix. We will provide a permanent solution in a future update.
%\starttwocolumn

\bibliographystyle{compling}
\bibliography{COLI_template}

\end{document}